\documentclass{article} 
\usepackage[table]{xcolor}
\usepackage[preprint]{nips2026_conference}

\usepackage{microtype}
\usepackage{hyperref}
\usepackage{url}
\usepackage{booktabs}
\usepackage{wrapfig}
\usepackage{amsmath}
\usepackage{amssymb}
\usepackage{mathtools}
\usepackage{amsthm}
\usepackage{makecell}
\usepackage{graphicx}
\usepackage{subcaption}
\usepackage{multirow}
\usepackage{multicol}
\usepackage{pifont}
\usepackage[misc]{ifsym}
\usepackage{enumitem}
\usepackage{color}
\definecolor{mygreen}{RGB}{0,120,90}
\usepackage{algorithm}          
\usepackage{algpseudocode}      
\usepackage{adjustbox}
\usepackage{titlesec}
\usepackage{caption} 
\usepackage{placeins}

\usepackage[capitalize,noabbrev]{cleveref}

\makeatletter
\providecommand{\theHALG@line}{}
\renewcommand{\theHALG@line}{\thealgorithm.\arabic{ALG@line}}
\makeatother

\definecolor{darkblue}{rgb}{0, 0, 0.5}
\hypersetup{colorlinks=true, citecolor=mid, linkcolor=primary, urlcolor=mid}

\renewcommand{\cite}[1]{\citep{#1}}

\title{When Should LLMs Reason? Adaptive Inference Routing via Entropy Dynamics}

\author{
\mbox{
 Wei Xia$^{2,1}$, Haoqing Wang$^{1}$, Yehui Tang$^{1}$ and Zhi-Hong Deng$^{2{~\textrm{\Letter}}}$}\\
$^1$ Samsung Research, Beijing, China\\
$^2$ State Key Laboratory of General Artificial Intelligence, School of Intelligence Science and Technology, Peking University\\
\texttt{xwisawesome}@stu.pku.edu.cn, \texttt{zhdeng}@pku.edu.cn\\
\texttt{\{haoqing.wang, yehui.tang\}}@samsung.com \\
$^\textrm{\Letter}$~Corresponding Author
}

\fancypagestyle{firstpage}{
  \fancyhead{}
  \fancyhead[L]{\emph{Preprint}}
}
\hypersetup{
  pdftitle={When Should LLMs Reason? Adaptive Inference Routing via Entropy Dynamics},
  pdfauthor={Wei Xia, Haoqing Wang, Yehui Tang, Zhi-Hong Deng}
}

\begin{document}

\maketitle
\begin{abstract}
Chain-of-thought (CoT) reasoning can improve LLM performance, but its benefits vary across tasks, models, and individual queries. On some tasks, CoT provides marginal or negative accuracy gains while substantially increasing token consumption. We investigate whether early decoding dynamics can help identify when explicit reasoning is useful. Our analysis shows that tasks with larger CoT gains tend to exhibit decreasing early-stage entropy, while low-gain tasks more often show oscillatory or increasing trajectories. These patterns motivate an empirical regime interpretation of decoding behavior. We propose \textbf{EDRM} (Entropy Dynamics-based Reasoning Manifold), a lightweight routing framework whose default heuristic requires no router training. EDRM summarizes early entropy trajectories with a compact, interpretable representation and selects among Direct, Standard, and CoT decoding. Across 15 benchmarks and 7 LLMs, its global and instance-level variants offer \textbf{16--55\%} token savings at comparable accuracy. The Full instance-level configuration combines routing with a short Direct fallback branch; our heuristic-router ablation reports up to \textbf{4.9 percentage points} of additional accuracy from oracle answer selection over routing without fallback. These results support selective reasoning based on early decoding signals and demonstrate the potential of entropy-based control for efficient LLM inference.
\end{abstract}

\section{Introduction}
Chain-of-thought (CoT) reasoning is widely used to elicit multi-step problem-solving from large language models (LLMs). Its benefits, however, vary with task type, model capability, and individual queries. Extended reasoning can even reduce performance on tasks requiring direct recall or fluent generation~\cite{Liu2024MindYS,Liu2025TokenSP}. Applying CoT indiscriminately can therefore increase token consumption and latency without improving accuracy. This raises a practical question: \emph{when should a model engage in explicit reasoning, and how can this decision be made at inference time?}

Understanding when reasoning helps requires considering both model capabilities and the process that unfolds during generation~\cite{Kim2024ReasoningCI,Li2024ChainOT,Zhao2026EntropyTS}, as well as the interaction between model capability and problem complexity~\cite{Sprague2024ToCO,Ding2025BESTRouteAL}. Prior work~\cite{Sui2025MetaReasonerDG,Su2025EntropyAwareSD} has explored these dimensions through routing and intervention strategies that adaptively invoke CoT. Such methods can require offline profiling or token-level modifications. We study an alternative source of routing information: the decoding dynamics of a specific model--query pair. These dynamics become observable during generation and may help predict the relative utility of explicit reasoning.

Can early decoding dynamics provide a useful signal for this decision? We investigate token-level entropy, which quantifies uncertainty in the next-token distribution. Tasks with larger CoT gains tend to exhibit decreasing early-stage entropy, while low-gain tasks more often show oscillatory or increasing trajectories. These patterns vary across tasks within a model and across models on the same task. We interpret them as empirical decoding regimes, with decreasing entropy consistent with a transition toward more concentrated predictions. This interpretation motivates adaptive routing without assuming that low entropy alone establishes reasoning quality.

Based on these observations, we propose \textbf{EDRM} (Entropy Dynamics-based Reasoning Manifold), a routing framework that uses early decoding entropy to select among Direct, Standard, and CoT. Its default heuristic uses three interpretable entropy descriptors and requires no router training; an optional MLP provides a learned routing alternative. Across 15 benchmarks and 7 LLMs, global and instance-level configurations offer \textbf{16--55\%} token savings at comparable accuracy. We separately report the additional accuracy obtained by the Full instance-level configuration with oracle answer selection.

\textbf{Contributions.} (1) We identify early entropy dynamics associated with CoT utility and use change-point detection to characterize temporal structure in entropy-based routing groups. (2) We introduce \textbf{EDRM}, a lightweight framework for dataset-level and instance-level routing with a training-free heuristic, an optional learned router, and a Direct fallback branch in the Full instance configuration. We distinguish routing efficiency from the accuracy contribution of oracle answer selection. (3) We evaluate the framework across 15 benchmarks and 7 LLMs spanning 3B--35B parameters, dense, MoE, and reasoning-distilled models, and examine its applicability across model scales and architectures under model-specific adaptation.

\section{Related Work}
\paragraph{Static \textit{vs.} Dynamic Views of LLM Reasoning.}
Research on LLM reasoning examines both model-level mechanisms and generation dynamics. Mechanistic interpretability studies identify neural circuits associated with reasoning~\cite{Kim2024ReasoningCI,Conmy2023TowardsAC}, while prompting studies investigate how CoT elicits reasoning capabilities~\cite{Li2024ChainOT,Wang2024ChainofThoughtRW}. Complementary work studies signals that vary during generation: semantic entropy serves as a hallucination risk indicator~\cite{Farquhar2024DetectingHI}, and entropy trajectories have been associated with reasoning reliability~\cite{Zhao2026EntropyTS,Zhu2026EDISDL}, motivating entropy-aware decoding and adaptive injection~\cite{Su2025EntropyAwareSD,Jin2025WellKT,He2026ThinkTB}. Recent work also quantifies reasoning effort via deep-thinking tokens~\cite{Chen2026ThinkDN} and distinguishes reasoning from recall via activation patterns~\cite{Fartale2025DisentanglingRA}.

\paragraph{Task-Only \textit{vs.} Task-Model Co-Dependency.}
Task-level studies report substantial CoT gains on mathematical reasoning and more limited or negative gains on factual retrieval~\cite{Sprague2024ToCO,Liu2024MindYS}. Other studies emphasize the interaction between problem difficulty and model capability: stronger models can directly solve some instances that require extensive CoT for weaker models. Methods align computation with this interaction through item response theory~\cite{Fernandez2025RADARRA}, budget-aware routing~\cite{Ding2025BESTRouteAL}, task-aware adaptation~\cite{Liu2026TaskAwareLR}, hybrid reasoning~\cite{Jiang2025ThinkOW}, mixture-of-experts~\cite{FeinAshley2025MixtureOT}, and task decomposition~\cite{Shao2025RouteandReasonSL,Shao2025DivisionofThoughtsHH,Qi2025PlanBS}. EDRM complements these approaches with routing signals observed during a short decoding probe, allowing decisions to reflect the current model--query pair.

\paragraph{Adaptive Efficient Reasoning.} Adaptive methods differ in their training requirements, decoding interventions, and use of offline information. Path guides~\cite{Sui2025MetaReasonerDG}, co-evolved routers~\cite{Huang2026EvolveRouterCR}, and synthetic-data approaches~\cite{Liu2026TaskAwareLR} require additional training. Speculative decoding~\cite{Su2025EntropyAwareSD}, position-specific trigger injection~\cite{Jin2025WellKT}, and entropy modulation~\cite{He2026ThinkTB} intervene within decoding, while RADAR~\cite{Fernandez2025RADARRA} and BEST-Route~\cite{Ding2025BESTRouteAL} use offline profiling. Lightweight entropy-based approaches~\cite{Sharma2025ThinkJE,Zhao2026EntropyTS,Zhu2026EDISDL} and hybrid expert systems~\cite{Jiang2025ThinkOW,FeinAshley2025MixtureOT} motivate efficient alternatives. Studies of error accumulation~\cite{Gan2025RethinkingES} and test-time scaling~\cite{Snell2024ScalingLT} further highlight the need to allocate reasoning effort selectively. Token Signature~\cite{Liu2025TokenSP} is a closely related two-way CoT/Direct router; we compare it with EDRM in \S\ref{sec:experiments}. EDRM combines cumulative uncertainty, trend, and local variation in a compact representation for three-way routing. Its default heuristic requires no router training or architectural modifications. Full instance-level results include a Direct fallback branch with oracle answer selection, while w/o-FB and Selected-FB assess routing alone and practical answer selection.

\begin{figure}[htbp]
    \centering
    \begin{minipage}[b]{0.48\textwidth}
        \centering
        \includegraphics[width=\linewidth]{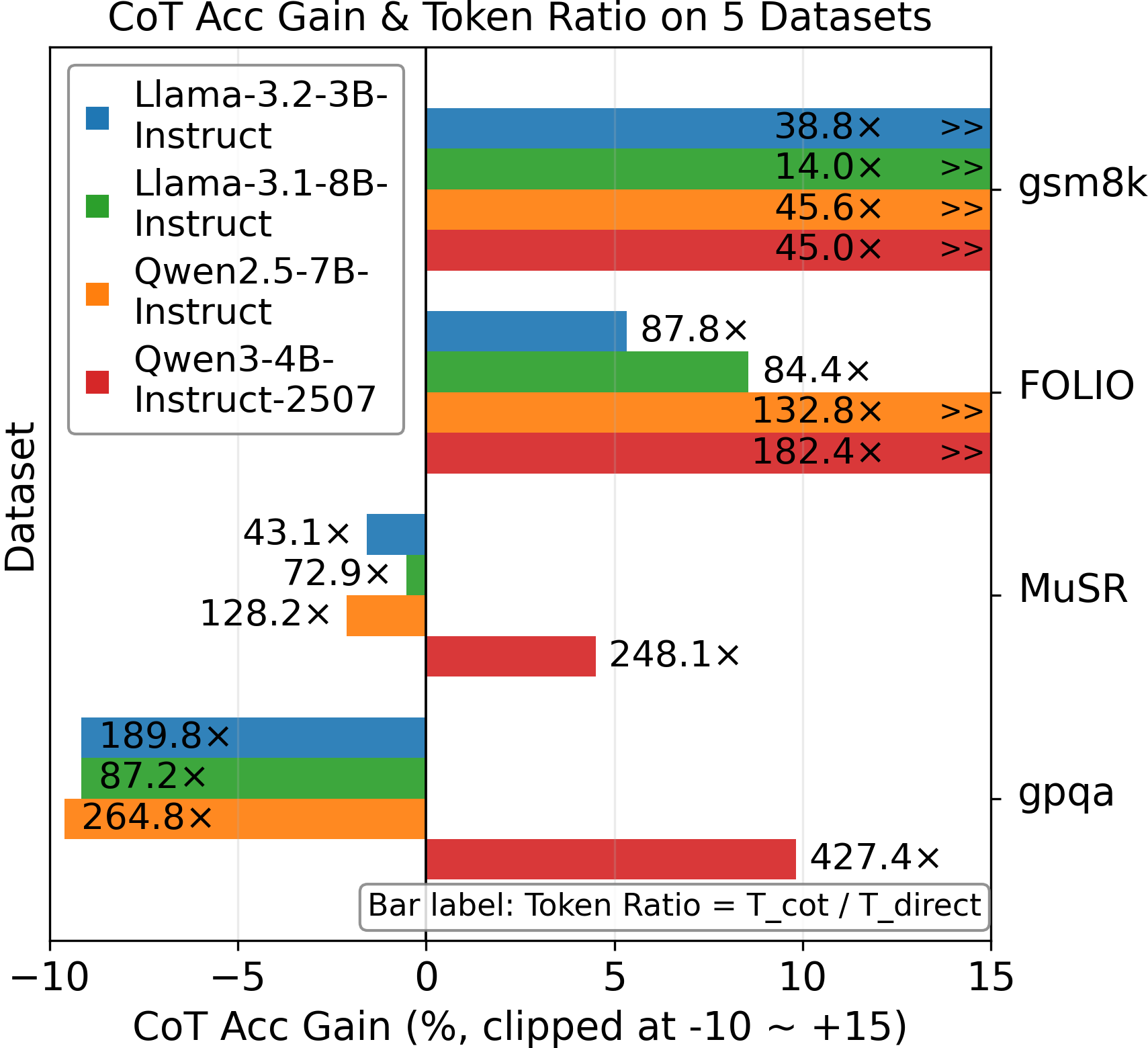}
        \captionof{figure}{CoT accuracy gains over Direct across models and tasks. Complete results are in Appendix~\ref{app:compare}. Legend abbreviations are expanded in Table~\ref{tab:overall_comparison}; ``Qwen3-4B'' denotes Qwen3-4B-Instruct-2507.}
        \label{fig:cot_gain}
    \end{minipage}
    \hfill
    \begin{minipage}[b]{0.49\textwidth}
        \centering
        \includegraphics[width=\linewidth]{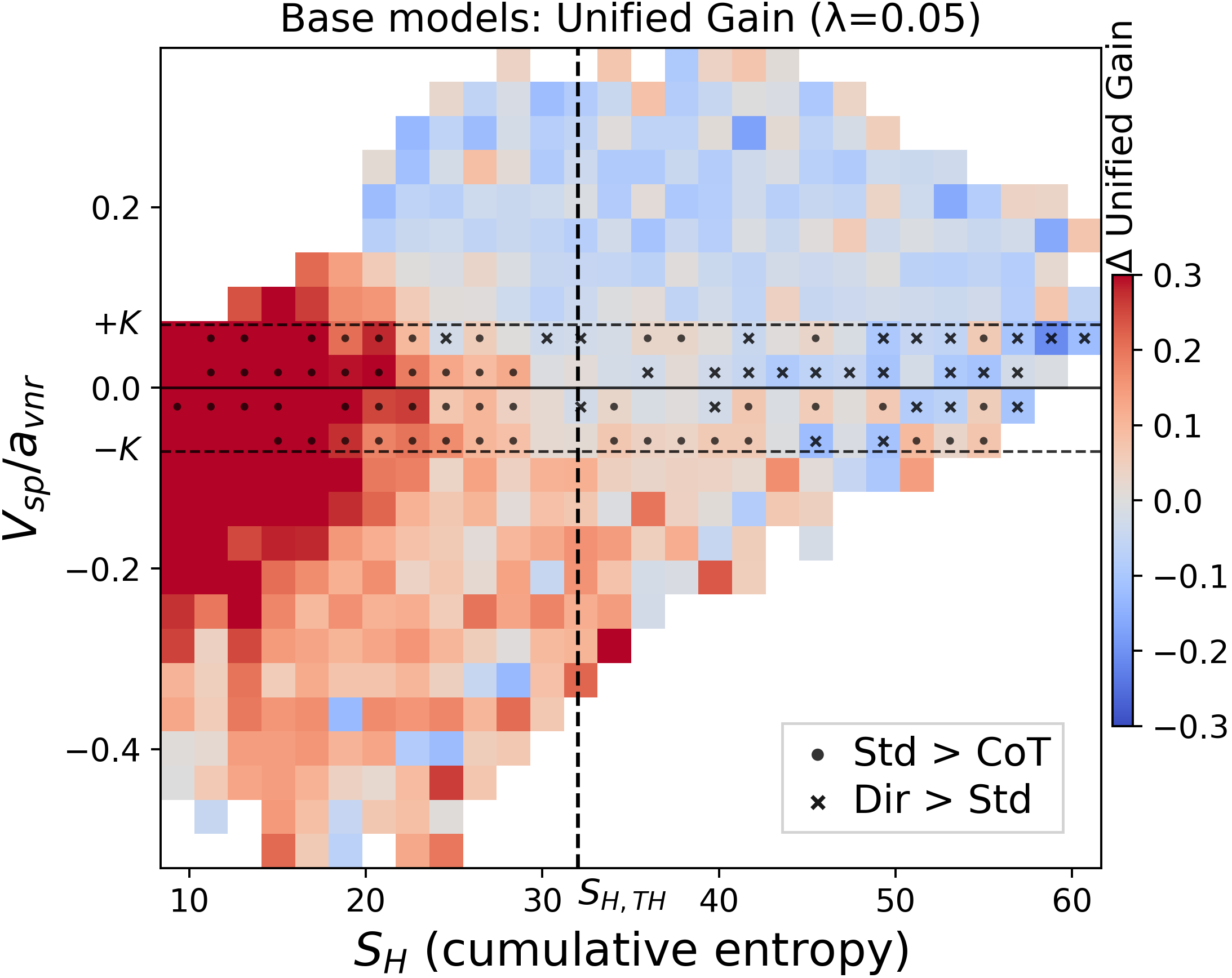}
        \captionof{figure}{Unified Gain Heatmap. Each cell shows the unified gain of \textit{CoT-Direct} for instances in a specific region of the $(V_{\text{sp}}/a_{\text{vnr}}, S_H)$ space.}
        \label{fig:heatmap}
    \end{minipage}
\end{figure}

\begin{figure}[t]
    \centering
    \includegraphics[width=\textwidth]{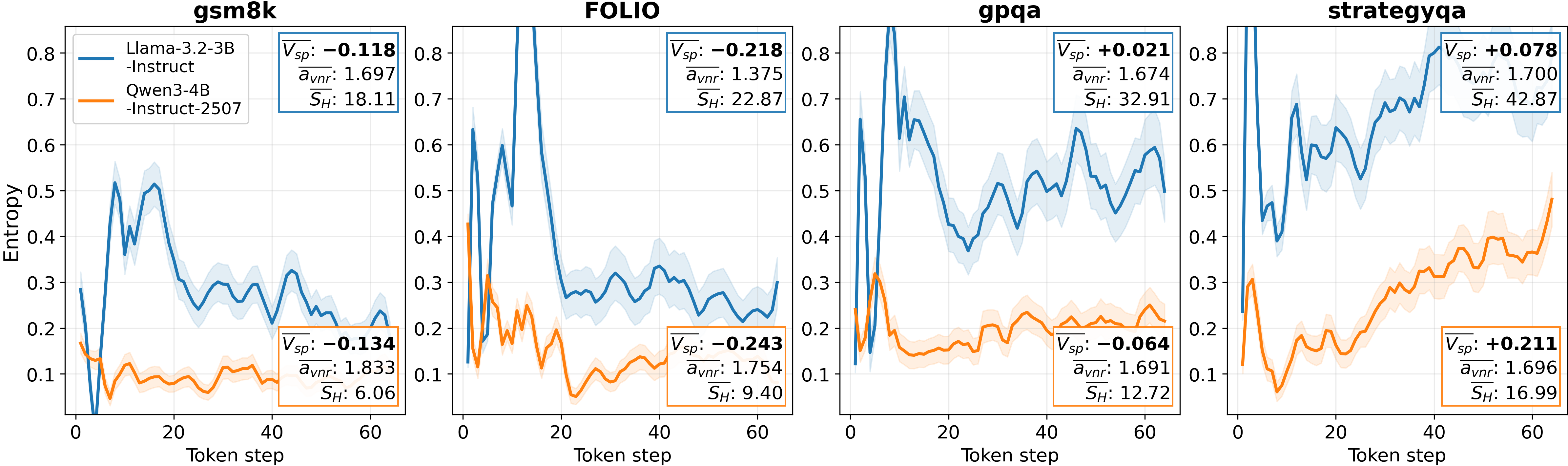}
    \caption{Mean token-level entropy over the first $N$ tokens under Standard probing. Tasks with larger CoT gains tend to show decreasing entropy; lower-gain tasks more often show oscillatory or increasing trajectories. Legend labels identify Llama-3.2-3B-Instruct, Llama-3.1-8B-Instruct, Qwen2.5-7B-Instruct, and Qwen3-4B-Instruct-2507; ``Qwen3-4B'' denotes the last checkpoint.}
    \label{fig:entropy_curve}
\end{figure}

\section{Methodology}
\label{sec:methodology}
We first define the decoding modes and entropy statistics, then analyze their relationship to CoT utility, and finally introduce EDRM's routing policy.
\subsection{Preliminaries}
\paragraph{Decoding paradigms.}
We consider three basic decoding paradigms under identical task descriptions:
\begin{itemize}[leftmargin=*,nosep]
\item \textbf{Direct}: the model is instructed to return the final answer without explanation. This mode minimizes explicit reasoning but can fail on problems requiring multi-step decomposition.
\item \textbf{Standard}: the model receives the task description without an explicit step-by-step reasoning instruction or a restriction against explanations. We use this mode for \textit{probing} and feature extraction because it permits explanatory generation without forcing CoT.
\item \textbf{CoT}: the model is prompted to reason step by step. This mode encourages explicit intermediate steps and can improve performance on complex problems, at a higher generation cost.
\end{itemize}
These paradigms range from minimal explicit reasoning (Direct) to prompted step-by-step reasoning (CoT), with Standard allowing explanations without an explicit CoT instruction. Qwen3-4B-Instruct-2507 uses its standard chat template in all three regimes, with no thinking-related control. Thinking-channel restrictions for Qwen3.5-35B-A3B and Direct-only prefilling for DeepSeek-R1-Distill-Qwen-7B are model-specific; Appendix~\ref{app:prompts} gives the exact settings. All generations use greedy decoding without token-level intervention.

\paragraph{Autoregressive generation and token-level entropy.}
Consider an autoregressive LLM that generates tokens sequentially. At each decoding step $i$, the model produces a probability distribution $p_i$ over the vocabulary $\mathcal{V}$ conditioned on the input context and previously generated tokens. The \emph{token-level entropy} at step $i$ is defined as:
\begin{equation}
H_i = -\sum_{v \in \mathcal{V}} p_i(v) \log p_i(v),
\label{eq:entropy}
\end{equation}
which quantifies uncertainty in the model's next-token prediction. Low entropy indicates a concentrated next-token distribution, while high entropy indicates multiple plausible continuations. The trajectory $\{H_i\}_{i=1}^{N}$ describes how this uncertainty evolves during early decoding; it does not by itself establish answer correctness.

\paragraph{Sequential statistics for entropy trajectory characterization.}
To systematically characterize entropy dynamics, we introduce two generic statistics for a scalar sequence $\{x_i\}_{i=1}^{N}$:

\textbf{Spearman correlation.} Spearman correlation measures monotonic association using ranks rather than raw values. We use it to summarize the direction of an entropy trajectory. Rank transformation reduces sensitivity to the magnitude of outliers and captures nonlinear monotonic trends. For a sequence $\{x_i\}_{i=1}^{N}$, its correlation with the step index is:
\begin{equation}
\text{Spearman}(\{1,\dots,N\},\{x_1,\dots,x_N\})=\mathrm{corr}(\mathrm{rank}(\{1,\dots,N\}),\mathrm{rank}(\{x_1,\dots,x_N\}))
\label{eq:vsp}
\end{equation}
For entropy trajectories, a negative correlation indicates a decreasing trend and a positive correlation an increasing trend. A value near zero indicates little overall monotonic association; the VNR below separately characterizes local variation.

\textbf{von Neumann ratio (VNR).} The VNR measures successive squared differences relative to overall variance. We use it to characterize local variation in the entropy sequence, complementing the monotonic trend captured by Spearman correlation. For a sequence $\{x_i\}_{i=1}^{N}$, it is defined as:
\begin{equation}
 \text{VNR}(x)=\frac{\frac{1}{N-1}\sum_{i=1}^{N-1}(x_{i+1}-x_i)^2}{\frac{1}{N}\sum_{i=1}^{N}(x_i-\bar{x})^2+\epsilon}
\label{eq:avnr}
\end{equation}
where $\bar{x}=\frac{1}{N}\sum_{i=1}^{N}x_i$ and $\epsilon$ is a small constant for numerical stability. A small VNR indicates small successive changes relative to overall variance, while a larger VNR indicates greater local variation. It complements Spearman correlation by characterizing temporal smoothness.

\subsection{Systematic Analysis}
\label{sec:systematic_analysis}

We begin with two empirical observations (Figure~\ref{fig:cot_gain}). First, CoT reasoning is not universally beneficial: on a substantial subset of tasks, it yields marginal or even negative gains while significantly increasing token consumption. Second, even on the same benchmark, different models can exhibit opposite CoT gains, suggesting that reasoning utility is not an intrinsic property of the task itself, but emerges from the interaction between the model and the task.

Motivated by these observations, we study whether early uncertainty reduction is associated with CoT utility. We analyze entropy trajectories over the first $N{=}64$ decoding steps under Standard prompting (Figure~\ref{fig:entropy_curve}). Three patterns emerge: (1) tasks with strong CoT gains typically exhibit a downward entropy trend; (2) tasks with weak or negative CoT gains more often show oscillatory or increasing entropy; and (3) the same model can display different dynamics across tasks, while different models can exhibit different trends on the same task. These observations motivate using early entropy dynamics as a routing signal conditioned on the model--query pair.

We interpret these patterns through an empirical \emph{regime shift} analogy: decreasing entropy describes a transition from relatively diffuse to more concentrated predictions. Oscillatory or increasing trajectories suggest that uncertainty persists or grows. This interpretation motivates three complementary descriptors of the early entropy trajectory:
\begin{equation}
S_H=\sum_{i=1}^{N}H_i,\quad
V_{\text{sp}}=\text{Spearman}(\{1,\dots,N\},\{H_1,\dots,H_N\}),\quad a_{\text{vnr}}=\text{VNR}(H_i)
\label{eq:sva}
\end{equation}

Empirically, we observe several consistent geometric patterns in this low-dimensional entropy space:

\begin{itemize}[leftmargin=*]
\item \textbf{Convergent reasoning regimes.}
Tasks with strong CoT gains typically exhibit negative $V_{\text{sp}}$ and relatively small $a_{\text{vnr}}$, corresponding to stable entropy reduction during decoding.

\item \textbf{Exploratory reasoning regimes.}
Tasks with weak or negative CoT gains often show increasing entropy or substantial local variation. These behaviors are captured separately by positive $V_{\text{sp}}$ and larger $a_{\text{vnr}}$; the heuristic combines them so that greater local variation reduces the influence of the trend.

\item \textbf{Uncertainty-overload regimes.}
Excessively large cumulative entropy $S_H$ often indicates that decoding remains highly uncertain even in early stages, where CoT is unlikely to reliably converge.
\end{itemize}

The heatmap in Figure~\ref{fig:heatmap} associates regions of this feature space with different CoT utilities, supporting its use as an empirical representation for routing.

\paragraph{Quantitative validation via change-point detection.}
To move beyond qualitative trajectory inspection, we apply two complementary change-point detection algorithms---CUSUM (cumulative sum) and BinSeg (binary segmentation)---to the per-sample entropy sequences over the first 64 tokens. For each model, we partition samples into three groups based on their routing decisions (Direct/Standard/CoT) and analyze change-point locations within each group ($n \approx 23$k samples per model). This analysis is label-free: grouping depends only on entropy-derived routing features, and detection reads only the entropy sequences (protocol details in Appendix~\ref{app:changepoint}). Three findings emerge:
(1) \emph{Detected splits explain more variance than random splits}: in the reported permutation comparison, BinSeg-detected change points explain 2.27--3.97$\times$ as much between-group variance as random split points (all $p<0.001$). This characterizes structure in the entropy sequences under the stated comparison.
(2) \emph{Change-point locations differ across routing groups}: CoT-routed samples exhibit the earliest median change points (BinSeg median 9--28 tokens across models), while Direct-routed samples exhibit the latest (median 33--38 tokens). Standard-routed samples lie between them on three of four models, consistent with earlier changes in trajectories assigned to CoT.
(3) \emph{Pre/post change-point entropy differs significantly}: 11 out of 12 model--group pairs show significant mean entropy differences before and after the detected change point (paired $t$-test, $p < 10^{-3}$; the sole exception is Qwen2.5-7B-Instruct under Standard, $p = 0.87$).
These results localize temporal structure within the entropy trajectories and are consistent with the proposed regime interpretation. Because grouping and detection both use entropy-derived information, this analysis does not independently establish a transition in reasoning quality. Full results are provided in Appendix~\ref{app:changepoint}.

\noindent\textbf{Core Insights.}
Early entropy trajectories exhibit patterns associated with CoT utility. Their trend, cumulative uncertainty, and local variation provide a compact empirical signal for adaptive routing, while change-point analysis further characterizes their temporal structure.

\subsection{Entropy Dynamics-based Reasoning Manifold (EDRM)}
\label{sec:edrm}

Motivated by these observations, we formulate adaptive reasoning as a \emph{decoding-state routing} problem. EDRM uses a short Standard decode to estimate whether a query is likely to benefit from explicit reasoning. The routing signal is derived from observable entropy dynamics rather than assigned from task categories alone.

\paragraph{Framework overview.}
Given a model $M$ and query $Q$, EDRM performs a short \textit{Standard} probe and records next-token entropy during the first $N$ decoding steps. We summarize the trajectory as $\mathbf{z}(Q)=(S_H,\;V_{\text{sp}},\;a_{\text{vnr}})$ using Eq.~\eqref{eq:sva} and the statistics in Eqs.~\eqref{eq:vsp} and~\eqref{eq:avnr}. The coordinates describe cumulative uncertainty, trend direction, and local variation, respectively. Negative $V_{\text{sp}}$ indicates decreasing uncertainty; larger $a_{\text{vnr}}$ indicates greater successive variation relative to overall variance; and larger $S_H$ indicates greater cumulative uncertainty. We call this empirical feature representation the \textbf{Entropy Dynamics-based Reasoning Manifold}; the term names the routing space without asserting a formal manifold structure. The default router applies a lightweight heuristic to these features and requires no router training.

\paragraph{Adaptive reasoning routing.}
Based on the manifold representation, EDRM adaptively selects among \textit{Direct}, \textit{Standard}, and \textit{CoT} decoding. The routing policy follows directly from the empirical regimes summarized in the previous subsection:
\begin{equation}
\mathcal{M}(Q)=\begin{cases}\text{Direct}, & V_{\text{sp}} > k\cdot a_{\text{vnr}} \;\lor\; (V_{\text{sp}}>0 \land S_H>S_{H,\text{th}}), \\
\text{CoT}, & V_{\text{sp}} < -k\cdot a_{\text{vnr}}, \\
\text{Standard}, & \text{otherwise}.
\end{cases}
\label{eq:router}
\end{equation}
A sufficiently negative $V_{\text{sp}}$ relative to $a_{\text{vnr}}$ favors CoT, following the observed association between decreasing entropy and CoT gains. Positive trends favor Direct when they exceed the volatility-scaled boundary or coincide with high cumulative entropy. Increasing $a_{\text{vnr}}$ widens the trend-based Standard region, subject to the separate cumulative-entropy condition for Direct. After selecting $\mathcal{M}(Q)$, EDRM generates the answer using that decoding mode. This task-agnostic heuristic uses a common feature representation across models, with calibration settings described below.

\paragraph{Learned manifold router.}
The heuristic above is the default, training-free router. EDRM also supports an \emph{optional} three-layer MLP trained on a 10\% stratified tuning split from each benchmark. We derive supervision from the correctness of the three decoding modes, using either multi-label targets or a single-label target that prioritizes cheaper modes when multiple modes are correct (Appendix~\ref{app:mlp_details}). The MLP selects $\mathcal{M}(Q)=\arg\max_m p_\theta(m\mid\mathbf{z}(Q))$ from the 3D descriptors or uses the full entropy trajectory or a hybrid input. The main-text results use the 64D trajectory and full-set evaluation; the appendix also reports input ablations and held-out results.

\paragraph{Fallback compensation and deployable selector.}
The Full instance-level configuration includes a \emph{Direct fallback} branch: when routing selects \textit{Standard} or \textit{CoT}, we also generate a \textit{Direct} answer. In Full, the final answer is chosen using ground-truth correctness, which we denote \emph{oracle answer selection}. The candidate-generation branches can both be run at inference time; this answer-selection rule is an evaluation reference that requires the true answer. Routing accounts for the substantial token reduction, while the short Direct branch adds a small generation cost and oracle selection provides additional accuracy. We separately evaluate \emph{w/o-FB}, which omits the Direct branch, and \emph{Selected-FB}, which replaces oracle selection with a per-benchmark probability threshold. Selected-FB switches to Direct on answer disagreements when its minimum token probability $d_{\mathrm{mp}}$ meets the per-benchmark threshold $\theta_{\mathrm{bench}}$. This threshold is calibrated by 5-fold cross-validation on 50 calibration samples per benchmark without using test labels, following Appendix~\ref{app:selector}; the entropy-threshold calibration used for routing is described separately below.

\begin{algorithm}[b]
\caption{Entropy Dynamics-based Routing (EDRM-Global and EDRM-Heuristic)}
\label{alg:routing}
\begin{algorithmic}[1]
\small
\Require Input $x$ (sample or dataset), probe length $N$, hyper-parameters $(k, S_{H,\text{th}})$
\Ensure Inference mode $M \in \{\text{Direct}, \text{Standard}, \text{CoT}\}$
\If{dataset-level routing}
    \State Compute statistics $(S_H, V_{\text{sp}}, a_{\text{vnr}}) \leftarrow (\bar{S}_H, \bar{V}_{\text{sp}}, \bar{a}_{\text{vnr}})$
\Else\Comment{instance-level routing}
    \State Run probing on $x$ $\rightarrow$ entropy $E_x$
    \State Compute $S_H(x), V_{\text{sp}}(x), a_{\text{vnr}}(x)$
\EndIf
\If{$V_{\text{sp}} > k\cdot a_{\text{vnr}}$ \textbf{or} $(V_{\text{sp}} > 0 \;\textbf{and}\; S_H > S_{H,\text{th}})$}
    \State $M \leftarrow \text{Direct}$
\ElsIf{$V_{\text{sp}} < -k\cdot a_{\text{vnr}}$}
    \State $M \leftarrow \text{CoT}$
\Else
    \State $M \leftarrow \text{Standard}$
\EndIf
\State \Return $M$
\end{algorithmic}
\end{algorithm}

\paragraph{Deployment scenarios.}
EDRM supports both dataset-level and instance-level deployment. For low-resource or cold-start settings, EDRM estimates global manifold statistics from a small calibration subset to initialize routing hyper-parameters without additional training. This provides a lightweight initialization mechanism for new datasets or models. For instance-level deployment, each query independently undergoes probing, manifold embedding, and adaptive routing, enabling fine-grained control over the trade-off between reasoning quality and token efficiency during inference. The working pipelines are presented in Algorithm~\ref{alg:routing}.

\paragraph{Calibrating $S_{H,\text{th}}$ (practical step).}
In the heuristic router, $S_{H,\text{th}}$ flags instances whose early decoding exhibits excessive cumulative uncertainty. We consider two ways to set $S_{H,\text{th}}$:\\
\textbf{(1)} Empirical thresholds: $S_{H,\text{th}}$ is set per model via binary search over candidate values, yielding $S_{H,\text{th}}{=}32$ for Llama-3.2-3B-Instruct, Llama-3.1-8B-Instruct, and Qwen2.5-7B-Instruct, $39$ for Qwen2.5-32B-Instruct, and $10/30/20$ for the models exhibiting verbose generation: Qwen3-4B-Instruct-2507, DeepSeek-R1-Distill-Qwen-7B, and Qwen3.5-35B-A3B, respectively.\\
\textbf{(2)} Cross-dataset calibration: we use dataset-level entropy trends to set $S_{H,\text{th}}$ without correctness labels. Motivated by the observed association between decreasing entropy and CoT utility, we count datasets with negative mean trends and use this count to index the ordered dataset-mean cumulative entropies: $S_{H,\text{th}} = \lfloor S_{(M)} \rfloor$, where $M = |\{j : V_{\text{sp}}^{(j)} < 0\}|$ and $S_{(M)}$ is the $M$-th order statistic of $\{S_H^{(j)}\}$. Appendix~\ref{app:calibration} gives the complete definition, boundary handling, and pseudocode.

\paragraph{Probe length selection.}
We set $N{=}64$, within the accuracy plateau observed over $N\in\{16,32,48,64,80,96\}$. For $N\geq48$, accuracy varies by at most 1.2 percentage points without fallback and 0.9 points under Full, supporting $N{=}64$ as a stable operating point (Appendix~\ref{app:probe_n}).

\paragraph{Other details about EDRM.}
We provide comprehensive technical details and extended analyses in Appendix~\ref{app:meth}. Appendix~\ref{app:descriptors} formally defines the three entropy descriptors and justifies design choices. Appendix~\ref{app:compare} extends the CoT-vs-Direct comparison to all model--benchmark pairs, showing substantial CoT gains on math tasks but marginal or negative gains on commonsense tasks. Appendix~\ref{app:visual} visualizes entropy trajectories across all settings. Appendix~\ref{app:unified_gain} introduces the Unified Gain metric and heatmaps that link CoT utility to manifold coordinates. Appendix~\ref{app:calibration} details the cross-dataset threshold calibration procedure. Appendix~\ref{app:routing_logic} analyzes the three-branch routing design and decision boundaries. Appendix~\ref{app:mlp_details} describes the learned MLP router variants and training setup.


\section{Experiments}
\label{sec:experiments}

\subsection{Experimental Setup}
\label{sec:experimental_setup}

We evaluate EDRM on 15 benchmarks covering varied reasoning tasks and difficulty levels, using 7 LLMs to assess its applicability across models.

\textbf{Datasets.}
We group the 15 benchmarks into four categories:
(1) mathematical and multi-step reasoning: \texttt{gsm8k}~\cite{Cobbe2021TrainingVT}, \texttt{MultiArith}~\cite{Roy2015SolvingGA}, \texttt{bbh}~\cite{Suzgun2022BBH};
(2) commonsense and knowledge reasoning: \texttt{commonsenseqa}~\cite{Talmor2019CommonsenseQA}, \texttt{strategyqa}~\cite{Geva2021DidAU}, \texttt{piqa}~\cite{Bisk2019PIQARA}, \texttt{siqa}~\cite{sap2019social}, \texttt{MuSR}~\cite{sprague2024musr};
(3) scientific reasoning: \texttt{arc\_challenge}~\cite{clark2018think}, \texttt{arc\_easy}~\cite{clark2018think}, \texttt{gpqa}~\cite{rein2023gpqa};
(4) formal and logical reasoning: \texttt{FOLIO}~\cite{han2024folio}, \texttt{ContextHub\_abductive}~\cite{hua2025disentangling}, \texttt{ContextHub\_deductive}~\cite{hua2025disentangling}, \texttt{lsat}~\cite{Zhong2023AGIEvalAH}.

\textbf{Models.}
We evaluate 7 LLMs in two groups:
\textit{Non-reasoning instruction-tuned models} (4): (1) Llama-3.2-3B-Instruct~\cite{grattafiori2024llama}, (2) Llama-3.1-8B-Instruct~\cite{grattafiori2024llama}, (3) Qwen2.5-7B-Instruct~\cite{Qwen2024Qwen25}, (4) Qwen2.5-32B-Instruct~\cite{Qwen2024Qwen25}. The first three cover the 3B--8B scale and different model families, while the 32B model tests generalization at larger scale.
\textit{Models exhibiting verbose generation} (3): (5) Qwen3-4B-Instruct-2507~\cite{yang2025qwen3}, a non-thinking checkpoint whose three decoding regimes differ only in their prompts; it still produces long responses under Standard (518.4 tokens on average); (6) DeepSeek-R1-Distill-Qwen-7B~\cite{DeepSeek2025R1}, a reasoning-distilled model built on Qwen2.5-Math-7B and producing long CoT traces (1314.2 tokens); (7) Qwen3.5-35B-A3B, a 35B MoE model with a switchable thinking channel, whose Standard responses remain long even with thinking disabled (1895.2 tokens, compared with 1703.4 under CoT). These models cover verbose generation without an explicit thinking channel, reasoning distillation, and persistent verbosity under thinking-channel suppression. Model-specific controls are detailed in Appendix~\ref{app:prompts}. We use full checkpoint names in the text. Abbreviations in tables are expanded in their captions; Qwen3-4B-2507 specifically denotes Qwen3-4B-Instruct-2507, the non-thinking checkpoint evaluated here.

\textbf{Baselines.}
We compare 9 decoding strategies in two categories:
\textbf{Static regimes} apply a fixed decoding mode to all samples: (1) \textit{Direct} (no reasoning), (2) \textit{Standard} (minimal prompting), (3) \textit{CoT} (always-on chain-of-thought).
\textbf{Adaptive routing} dynamically selects the reasoning mode: (4) \textit{Token-Signature}, which uses a token-probability trend and logistic regression to select between Direct and CoT (specification in Appendix~\ref{app:ts_baseline}); and the \textbf{EDRM variants}, at two granularities---(5--6) \textit{EDRM-Global-E/C} for dataset-level routing and (7--8) \textit{EDRM-Inst-E/C} for instance-level routing, where ``E'' uses empirical thresholds ($S_{H,\text{th}}{=}32$ for Llama-3.2-3B-Instruct, Llama-3.1-8B-Instruct, and Qwen2.5-7B-Instruct, $39$ for Qwen2.5-32B-Instruct, $10$ for Qwen3-4B-Instruct-2507, $30$ for DeepSeek-R1-Distill-Qwen-7B, and $20$ for Qwen3.5-35B-A3B) and ``C'' uses cross-dataset calibration; (9) \textit{EDRM-MLP} uses a learned instance-level router (training and data-partitioning protocol in Appendices~\ref{app:mlp_details} and~\ref{app:calibration_protocol}). EDRM variants involving randomness are evaluated over 8 random seeds, and we report means and variances; the fixed-seed TS fitting protocol is detailed in Appendix~\ref{app:ts_baseline}.

\textbf{Evaluation metrics.}
We report accuracy and average token consumption to characterize the performance--efficiency trade-off. For multi-trial settings we report means and variances. Token consumption includes amortized probe overhead (1.46--2.08 tokens/sample, $<1.8\%$ of total; see Appendix~\ref{app:amortization}). Detailed computations are given in Appendix~\ref{app:eval_metrics}.

\textbf{Other details.}
The probe length is $N{=}64$, which sits at the center of the accuracy plateau (Appendix~\ref{app:probe_n}; ablation over $N\in\{16,\dots,96\}$). The heuristic router uses $k{=}0.07$ (the four non-reasoning instruction-tuned models and Qwen3-4B-Instruct-2507) or $k{=}0.15$ (DeepSeek-R1-Distill-Qwen-7B and Qwen3.5-35B-A3B, whose entropy trajectories are more volatile, where $k{=}0.15$ is optimal). All generation uses greedy decoding. For EDRM calibration sampling, we use 50 instances per benchmark (1\%--11\%, depending on benchmark size) and run 8 trials with different random seeds. The main-text MLP results use the full entropy trajectory (64D) input and are evaluated on the full benchmark sets; variants with 3D/67D inputs and original/calibrated labels are in Appendix~\ref{app:mlp_details}.

\subsection{Main Results}

Table~\ref{tab:overall_comparison} presents the overall comparison of 7 models across 9 methods. We analyze the results along three dimensions.

\begin{table*}[htbp]
\centering
\caption{Overall comparison across 15 benchmarks (sample-weighted accuracy in \% / average tokens). Inst-E/C and MLP use Full with oracle answer selection; TS and MLP use full-set evaluation, including their router-fitting samples; TS uses a 64-step probe (Appendix~\ref{app:ts_baseline}). Global-E saves 37--55\% of tokens on the four non-reasoning instruction-tuned models at comparable accuracy. Full instance-level accuracy includes both routing and oracle answer selection. Bold marks highest accuracy per model, including Full results; underlining marks lowest token cost excluding Direct. Token counts include amortized probe overhead. Model abbreviations: Llama-3.2-3B = Llama-3.2-3B-Instruct; Llama-3.1-8B = Llama-3.1-8B-Instruct; Qwen2.5-7B = Qwen2.5-7B-Instruct; Qwen2.5-32B = Qwen2.5-32B-Instruct; Qwen3-4B-2507 = Qwen3-4B-Instruct-2507; DR1-Qwen-7B = DeepSeek-R1-Distill-Qwen-7B; Qwen3.5-35B = Qwen3.5-35B-A3B.}
\label{tab:overall_comparison}
\begin{adjustbox}{max width=\textwidth}
\small
\setlength{\tabcolsep}{3pt}
\renewcommand{\arraystretch}{1.18}
\begin{tabular}{lccccccc}
\toprule
\textbf{Method} & \makecell{\textbf{Llama-3.2}\\\textbf{3B}} & \makecell{\textbf{Llama-3.1}\\\textbf{8B}} & \makecell{\textbf{Qwen2.5}\\\textbf{7B}} & \makecell{\textbf{Qwen2.5}\\\textbf{32B}} & \makecell{\textbf{Qwen3}\\\textbf{4B-2507}} & \makecell{\textbf{DR1-Qwen}\\\textbf{7B}} & \makecell{\textbf{Qwen3.5}\\\textbf{35B}} \\
\midrule
Direct & 50.73 / 4.4 & 57.31 / 5.0 & 62.28 / 3.4 & 71.35 / 4.7 & 63.75 / 6.2 & 53.28 / 3.5 & 73.66 / 6.6 \\
Standard & 59.89 / 194.4 & 66.08 / 127.1 & 71.43 / 240.6 & 78.37 / 238.7 & 79.90 / 518.4 & 70.20 / 1389.0 & 83.52 / 1895.2 \\
CoT & 60.20 / 251.7 & 68.11 / 277.8 & 72.44 / 330.3 & 78.57 / 306.7 & \textbf{80.75} / 642.5 & 70.82 / 1314.2 & 83.34 / 1703.4 \\
Token-Sign. & 60.78 / 251.0 & 68.26 / 264.3 & 72.78 / 356.7 & 79.54 / 266.9 & 80.74 / 682.3 & 70.76 / 1298.4 & 84.50 / 1512.2 \\
\midrule
Global-E & 59.87 / \underline{113.2} & 67.27 / 164.4 & 71.66 / \underline{191.8} & 78.89 / \underline{194.6} & 76.35 / \underline{335.1} & 69.98 / 1342.9 & 83.13 / 1711.3 \\
Global-C & 59.99 / 145.9 & 67.15 / 172.1 & 71.76 / 195.5 & 78.75 / 205.3 & 78.42 / 404.2 & 70.20 / 1389.0 & 83.03 / 1651.0 \\
\midrule
Inst-E (Full) & 63.63 / 179.4 & 70.04 / 153.9 & 76.07 / 242.7 & 81.87 / 248.6 & 78.54 / 401.1 & 74.78 / 1139.4 & 85.25 / 1276.1 \\
Inst-C (Full) & 63.89 / 182.7 & \textbf{70.26} / 155.9 & \textbf{76.20} / 244.8 & \textbf{81.99} / 250.9 & 79.33 / 426.7 & \textbf{75.71} / 1188.7 & 85.18 / \underline{1268.0} \\
MLP (Full) & \textbf{64.91} / 180.8 & 69.14 / \underline{119.9} & 75.35 / 224.3 & 81.31 / 215.8 & 79.33 / 422.0 & 74.26 / \underline{1008.3} & \textbf{86.00} / 1423.1 \\
\bottomrule
\end{tabular}
\end{adjustbox}
\end{table*}

\paragraph{Token efficiency across models.}
EDRM-Global-E saves 37--55\% of tokens relative to CoT on the four non-reasoning instruction-tuned models and 48\% on Qwen3-4B-Instruct-2507 (benchmark-level paired permutation tests $p{<}0.001$ on all five; see \S\ref{sec:stat_analysis}). On the former group, accuracy remains within 0.9 percentage points of CoT, with 113.2--194.6 tokens compared with CoT's 251.7--330.3. On DeepSeek-R1-Distill-Qwen-7B, EDRM-MLP (Full) uses 1008.3 tokens compared with CoT's 1314.2, a 23.3\% reduction. Its 3.4-percentage-point accuracy gain includes oracle answer selection. The token savings arise from routing; the Direct fallback branch adds a small generation cost, while answer selection does not change that generation budget.

\paragraph{Routing, fallback, and answer selection.}
The Full instance-level configuration combines routing with a Direct fallback branch and oracle answer selection. We evaluate it alongside \textit{w/o-FB}, which omits the fallback branch, and \textit{Selected-FB}, which uses a deployable per-benchmark $d_{\mathrm{mp}}$ selector. Selected-FB switches to Direct on answer disagreements when $d_{\mathrm{mp}}\geq\theta_{\mathrm{bench}}$; thresholds follow 5-fold calibration on 50 samples per benchmark without test-label access (Appendix~\ref{app:selector}). Table~\ref{tab:ablation} reports these configurations and the feature ablations. Full gains 1.7--4.9 percentage points over w/o-FB across the four evaluated models, while w/o-FB itself saves 27--46\% of tokens relative to CoT (all $p{<}0.001$). Selected-FB recovers 3.1--15.1\% of the Full--w/o-FB accuracy gap on the first three models at the same generation cost as Full. Recovery is $-$6.3\% on Qwen3-4B-Instruct-2507, whose disagreement subset is highly imbalanced ($D_d{:}D_m\approx1{:}8.6$). Recovery rates use unrounded accuracies. Retrospective per-benchmark threshold selection with test-label access recovers only 0.07--1.18 percentage points under the evaluated selector families; this is an empirical result for those selectors (Appendix~\ref{app:selector}).

\paragraph{Full instance-level results relative to global routing.}
On the four non-reasoning instruction-tuned models, EDRM-Inst-E (Full) exceeds EDRM-Global-E accuracy by 2.8--4.4 percentage points (Llama-3.2-3B-Instruct: 59.87$\to$63.63; Llama-3.1-8B-Instruct: 67.27$\to$70.04; Qwen2.5-7B-Instruct: 71.66$\to$76.07; Qwen2.5-32B-Instruct: 78.89$\to$81.87), while using fewer tokens than CoT. These configurations differ in both routing granularity and answer selection: Global has no Direct fallback, whereas Full uses oracle answer selection. The ablations separately quantify routing without fallback and the additional accuracy from answer selection.

\subsection{Cross-Scale and Cross-Architecture Generalization}

To assess EDRM's applicability across the seven evaluated models, we organize the results in Table~\ref{tab:overall_comparison} along three dimensions: model scale, verbose generation in reasoning-distilled and non-thinking checkpoints, and dense/MoE architectures. Here, generalization refers to the applicability of the same routing framework under model-specific calibration or router training, as detailed in Appendices~\ref{app:calibration_protocol} and~\ref{app:mlp_details}.

\paragraph{Scale generalization.}
\textbf{(1) Qwen2.5-7B-Instruct$\to$Qwen2.5-32B-Instruct.}
CoT accuracy increases by 6.1 percentage points from the 7B to the 32B model (72.4$\to$78.6), while EDRM retains substantial token savings at both scales. The MLP (w/o-FB) saves 32.7\% and 30.6\% of tokens on the 7B and 32B models, respectively, with accuracy within 0.7 percentage points of CoT (Table~\ref{tab:mlp-remaining-ml64d} and Appendix~\ref{app:ablation-mlp}). These results show that the deployable routing configuration maintains comparable accuracy with 30--33\% fewer tokens across the two scales. On the 32B model, CoT still exceeds Direct by 7.2 percentage points, highlighting the value of selectively allocating reasoning effort even at the larger scale.

\smallskip
\noindent\textbf{(2) Llama-3.2-3B-Instruct$\to$Llama-3.1-8B-Instruct.}
CoT accuracy increases from 60.20\% to 68.11\% across these checkpoints. Global-E reduces token use from 251.7 to 113.2 on the 3B model and from 277.8 to 164.4 on the 8B model, saving 55.0\% and 40.8\%, respectively. Its accuracies of 59.87\% and 67.27\% remain within 0.9 percentage points of CoT. Thus, substantial routing savings remain available on the more accurate 8B checkpoint. This pair differs in both release version and parameter count, so it assesses applicability across the two released Llama models.

\paragraph{Over-reasoning suppression.}
\textbf{(1) Qwen2.5-7B-Instruct$\to$DeepSeek-R1-Distill-Qwen-7B.}
Compared with Qwen2.5-7B-Instruct, DeepSeek-R1-Distill-Qwen-7B generates CoT traces that are 4.0$\times$ as long (330$\to$1314 tokens), yet its aggregate CoT accuracy is 1.6 percentage points lower (72.4$\to$70.8). This comparison illustrates long-form generation in the reasoning-distilled model: additional reasoning effort does not consistently translate into higher accuracy on these benchmarks. EDRM-MLP (Full) reduces the distilled model's token consumption to 1008.3 tokens, a 23.3\% saving relative to CoT. Direct accuracy is also lower on the distilled model (53.3\% vs.\ 62.3\%), underscoring the importance of adapting routing to each model's decoding-mode performance. With oracle answer selection, Inst-C (Full) reaches 75.7\% accuracy, 4.9 percentage points above CoT, illustrating the combined accuracy of routing and oracle answer selection in this configuration.

\smallskip
\noindent\textbf{(2) Qwen2.5-7B-Instruct$\to$Qwen3-4B-Instruct-2507.}
Even without a thinking channel, Qwen3-4B-Instruct-2507 generates 518.4 tokens under Standard and 642.5 under CoT, approximately 2.2$\times$ and 1.9$\times$ the corresponding Qwen2.5-7B-Instruct lengths. Its CoT accuracy is also higher (80.75\% vs.\ 72.44\%). Persistent verbosity after instruction post-training leaves room for inference-time compute control in this non-thinking checkpoint. EDRM-MLP (w/o-FB) uses 418.2 tokens at 77.49\% accuracy, a 34.9\% token saving relative to CoT (Table~\ref{tab:mlp-remaining-ml64d}). Full uses 422.0 tokens at 79.33\% accuracy, saving 34.3\% of tokens with a 1.42-percentage-point accuracy reduction from CoT; its accuracy additionally includes oracle answer selection. These results show that EDRM can reduce generation length beyond the non-thinking configuration, complementing model-level post-training with an explicit accuracy--cost trade-off.

\paragraph{Dense vs.\ MoE (Qwen2.5-32B-Instruct$\to$Qwen3.5-35B-A3B).}
The MoE model has 4.8 percentage points higher CoT accuracy than the dense model, while generating 5.6$\times$ as many CoT tokens (307$\to$1703); its Standard output is longer still, at 1895 tokens despite the thinking-channel suppression described in Appendix~\ref{app:prompts}. EDRM-MLP (Full) saves 29.6\% of tokens on the dense model and 16.5\% on the MoE model. Although the relative saving is smaller on the MoE model, the absolute reduction is larger: approximately 280 tokens per sample, compared with 91 on the dense model. With oracle answer selection, MLP (Full) exceeds CoT accuracy by approximately 2.7 percentage points on both models. These results demonstrate the framework's applicability to the evaluated dense and MoE models under their respective routing configurations. On the 35B model, CoT retains a 9.7-percentage-point advantage over Direct, highlighting the accuracy--cost trade-off that adaptive routing addresses.

Together, these comparisons support EDRM's applicability across the evaluated scales, architectures, and reasoning behaviors under model-specific adaptation. The magnitude of token savings varies across models, with verbose models offering substantial absolute reductions even when relative savings are smaller. The evaluated settings include long responses from a non-thinking checkpoint, reasoning-distilled outputs, and verbosity persisting under thinking-channel suppression; this motivates routing beyond mode selection alone. Accuracy results for the Full configurations additionally include the contribution of oracle answer selection, as distinguished from routing alone in the ablations.

\subsection{Global-Level Routing}

Table~\ref{tab:llama32_results} presents the per-dataset results of EDRM-Global on Llama-3.2-3B-Instruct. Per-dataset results for the other models are in Appendix~\ref{app:edrm_global_results}.

\begin{table*}[t]
\centering
\caption{EDRM-Global(E\&C) Performance comparison of Llama-3.2-3B-Instruct across 15 benchmarks. EDRM-Global-E achieves 59.87\% accuracy with 113.2 average tokens, reducing token cost by 55.0\% compared to CoT (251.7 tokens). All per-dataset accuracy variances are below $1.9 \times 10^{-3}$.}
\vspace{-0.5em}
\label{tab:llama32_results}
\begin{adjustbox}{max width=0.90\textwidth}
\Large
\setlength{\tabcolsep}{1.5pt}
\renewcommand{\arraystretch}{1.3}
\begin{tabular}{l|cc|cc|cc|ccc|ccc}
\toprule
\multirow{2}{*}{\textbf{Dataset}} & \multicolumn{2}{c|}{\textbf{Direct}} & \multicolumn{2}{c|}{\textbf{Standard}} & \multicolumn{2}{c|}{\textbf{CoT}} & \multicolumn{3}{c|}{\textbf{EDRM-Global-E}} & \multicolumn{3}{c}{\textbf{EDRM-Global-C}} \\
\cmidrule{2-13}
& Acc(\%) & Tok & Acc(\%) & Tok & Acc(\%) & Tok & $\overline{\text{Acc}}$(\%) & $\overline{\text{Tok}}$ & D:S:C & $\overline{\text{Acc}}$(\%) & $\overline{\text{Tok}}$ & D:S:C \\
\midrule
ARC-C & 71.93 & 4.0 & 76.54 & 177.9 & 75.17 & 243.8 & 71.93 & 4.0 & 8:0:0 & 73.66 & 69.2 & 5:3:0 \\
ARC-E & 87.37 & 4.0 & 88.85 & 169.2 & 89.23 & 237.1 & 87.37 & 4.0 & 8:0:0 & 88.66 & 148.6 & 1:7:0 \\
BBH & 44.65 & 4.2 & 53.33 & 215.5 & 55.06 & 277.8 & 53.55 & 223.3 & 0:7:1 & 53.54 & 223.3 & 0:7:1 \\
CSQA & 65.60 & 5.4 & 67.90 & 88.3 & 65.44 & 186.1 & 65.60 & 5.4 & 8:0:0 & 65.60 & 5.4 & 8:0:0 \\
CH-Abd & 33.46 & 3.6 & 35.08 & 242.2 & 36.42 & 254.6 & 35.25 & 243.7 & 0:7:1 & 35.25 & 243.7 & 0:7:1 \\
CH-Ded & 20.67 & 3.9 & 42.83 & 280.1 & 41.54 & 258.1 & 42.02 & 266.4 & 0:3:5 & 42.03 & 266.4 & 0:3:5 \\
FOLIO & 43.27 & 3.8 & 44.77 & 298.2 & 48.59 & 334.4 & 48.59 & 334.4 & 0:0:8 & 48.59 & 334.4 & 0:0:8 \\
GPQA & 33.93 & 4.0 & 25.67 & 530.0 & 24.78 & 767.0 & 30.83 & 201.3 & 5:3:0 & 25.67 & 530.0 & 0:8:0 \\
GSM8K & 8.42 & 6.0 & 75.66 & 227.2 & 74.07 & 231.5 & 75.06 & 228.8 & 0:5:3 & 75.07 & 228.8 & 0:5:3 \\
LSAT & 41.13 & 4.0 & 41.13 & 303.3 & 39.25 & 420.5 & 41.13 & 4.0 & 8:0:0 & 41.13 & 191.1 & 3:5:0 \\
MultiArith & 24.50 & 5.6 & 96.83 & 129.6 & 97.67 & 115.7 & 97.46 & 119.2 & 0:2:6 & 97.46 & 119.2 & 0:2:6 \\
MuSR & 51.32 & 4.1 & 48.68 & 154.9 & 49.74 & 178.1 & 51.32 & 4.1 & 8:0:0 & 51.32 & 4.1 & 8:0:0 \\
PIQA & 75.52 & 4.8 & 71.49 & 130.4 & 72.52 & 191.6 & 75.52 & 4.8 & 8:0:0 & 75.52 & 4.8 & 8:0:0 \\
SIQA & 63.15 & 4.9 & 64.84 & 76.0 & 63.36 & 191.1 & 63.15 & 4.9 & 8:0:0 & 63.15 & 4.9 & 8:0:0 \\
StratQA & 61.05 & 4.3 & 61.14 & 121.9 & 64.06 & 225.9 & 61.05 & 4.3 & 8:0:0 & 61.05 & 4.3 & 8:0:0 \\
\midrule
\textbf{Overall} & 50.73 & 4.4 & 59.89 & 194.4 & 60.20 & 251.7 & 59.87 & 113.2 & -- & 59.99 & 145.9 & -- \\
\bottomrule
\end{tabular}
\end{adjustbox}
\end{table*}

\paragraph{Routing stability.}
EDRM-Global-E (empirical threshold) and EDRM-Global-C (calibrated threshold) produce highly consistent routing decisions across 8 random seeds, with per-dataset accuracy variance below $1.9\times10^{-3}$. The highly concentrated D:S:C counts (e.g., 8:0:0 on \texttt{ARC-C}, 0:0:8 on \texttt{FOLIO}) confirm that the lightweight global probe yields reliable, low-variance dataset-level routing.

\paragraph{Token efficiency at comparable accuracy.}
On Llama-3.2-3B-Instruct, EDRM-Global-E reaches 59.87\% accuracy compared with CoT's 60.20\%, a difference of 0.33 percentage points, while reducing token use by 55.0\% (251.7$\to$113.2). Relative to Standard, it saves 41.8\% of tokens (194.4$\to$113.2) at nearly identical accuracy (59.87\% vs.\ 59.89\%). The main benefit in this setting is lower generation cost at comparable accuracy.

\paragraph{Routing decisions align semantically with reasoning demand.}
In this evaluation, Global-E often selects \textit{Direct} on benchmarks where CoT offers limited gains relative to its token cost, including \texttt{ARC-C/E}, \texttt{CSQA}, and \texttt{PIQA}. It selects \textit{CoT} on \texttt{FOLIO}, where CoT accuracy exceeds Direct. Mixed decisions across calibration trials on \texttt{GSM8K}, \texttt{BBH}, and \texttt{GPQA} indicate sensitivity near the routing boundaries and motivate instance-level analysis.

\subsection{Instance-Level Routing}

To mitigate the sampling sensitivity on multi-step benchmarks (\texttt{GSM8K}, \texttt{BBH}), we evaluate instance-level routing, which adapts dynamically to per-sample entropy trajectories for fine-grained, on-demand reasoning. Table~\ref{tab:llama32_instance} presents the per-dataset results on Llama-3.2-3B-Instruct.

\begin{table*}[t]
\centering
\caption{EDRM instance-level performance on Llama-3.2-3B-Instruct. All instance-level EDRM columns use Full with oracle answer selection. EDRM-Inst-C achieves 63.89\% accuracy with 182.7 tokens; EDRM-MLP achieves 64.91\% with 180.8 tokens under full-set evaluation. Detailed results for other models are in Appendix~\ref{app:edrm_instance_results}.}
\vspace{-0.5em}
\label{tab:llama32_instance}
\begin{adjustbox}{max width=1.0\textwidth}
\Large
\setlength{\tabcolsep}{1.5pt}
\renewcommand{\arraystretch}{1.3}
\begin{tabular}{l|cc|cc|cc|cc|cc|cc|cc}
\toprule
\multirow{2}{*}{\textbf{Dataset}} & \multicolumn{2}{c|}{\textbf{Direct}} & \multicolumn{2}{c|}{\textbf{Standard}} & \multicolumn{2}{c|}{\textbf{CoT}} & \multicolumn{2}{c|}{\textbf{Token-Sign}} & \multicolumn{2}{c|}{\textbf{EDRM-Inst-E}} & \multicolumn{2}{c|}{\textbf{EDRM-Inst-C}} & \multicolumn{2}{c}{\textbf{EDRM-MLP}} \\
\cmidrule{2-15}
& Acc(\%) & Tok & Acc(\%) & Tok & Acc(\%) & Tok & Acc(\%) & Tok & Acc(\%) & Tok & $\overline{\mathrm{Acc}}$ (\%) & $\overline{\mathrm{Tok}}$ & Acc(\%) & Tok \\
\midrule
ARC-C & 71.93 & 4.0 & 76.54 & 177.9 & 75.17 & 243.8 & 75.17 & 305.0 & 77.56 & 129.8 & 77.56 & 135.8 & 77.65 & 126.4 \\
ARC-E & 87.37 & 4.0 & 88.85 & 169.2 & 89.23 & 237.1 & 89.35 & 289.5 & 89.73 & 132.1 & 89.89 & 138.2 & 89.90 & 117.5 \\
BBH & 44.65 & 4.2 & 53.33 & 215.5 & 55.06 & 277.8 & 55.06 & 340.4 & 58.87 & 242.2 & 59.01 & 242.7 & 60.08 & 228.3 \\
CSQA & 65.60 & 5.4 & 67.90 & 88.3 & 65.44 & 186.1 & 65.60 & 58.4 & 68.22 & 82.4 & 68.50 & 83.4 & 70.60 & 90.5 \\
CH-Abd & 33.46 & 3.6 & 35.08 & 242.2 & 36.42 & 254.6 & 36.42 & 318.5 & 46.88 & 220.5 & 47.51 & 226.0 & 50.79 & 242.1 \\
CH-Ded & 20.67 & 3.9 & 42.83 & 280.1 & 41.54 & 258.1 & 41.54 & 322.0 & 41.63 & 258.6 & 42.31 & 264.0 & 41.92 & 253.2 \\
FOLIO & 43.27 & 3.8 & 44.77 & 298.2 & 48.59 & 334.4 & 47.51 & 349.4 & 57.56 & 324.1 & 57.70 & 326.9 & 54.15 & 243.7 \\
GPQA & 33.93 & 4.0 & 25.67 & 530.0 & 24.78 & 767.0 & 33.93 & 58.5 & 38.84 & 358.8 & 39.23 & 375.0 & 41.83 & 453.6 \\
GSM8K & 8.42 & 6.0 & 75.66 & 227.2 & 74.07 & 231.5 & 74.07 & 295.2 & 72.40 & 251.9 & 72.55 & 252.8 & 72.63 & 248.6 \\
LSAT & 41.13 & 4.0 & 41.13 & 303.3 & 39.25 & 420.5 & 42.02 & 279.2 & 46.98 & 287.4 & 47.08 & 290.3 & 48.46 & 314.4 \\
MultiArith & 24.50 & 5.6 & 96.83 & 129.6 & 97.67 & 115.7 & 97.67 & 179.6 & 94.50 & 159.4 & 94.50 & 159.4 & 96.17 & 145.2 \\
MuSR & 51.32 & 4.1 & 48.68 & 154.9 & 49.74 & 178.1 & 52.65 & 137.8 & 54.63 & 92.3 & 54.73 & 94.7 & 58.33 & 123.1 \\
PIQA & 75.52 & 4.8 & 71.49 & 130.4 & 72.52 & 191.6 & 75.52 & 172.7 & 77.04 & 97.4 & 77.07 & 98.4 & 77.69 & 99.4 \\
SIQA & 63.15 & 4.9 & 64.84 & 76.0 & 63.36 & 191.1 & 63.15 & 65.2 & 63.77 & 66.8 & 63.97 & 67.1 & 67.55 & 98.0 \\
StratQA & 61.05 & 4.3 & 61.14 & 121.9 & 64.06 & 225.9 & 64.06 & 285.4 & 66.20 & 118.4 & 66.51 & 120.8 & 67.12 & 119.4 \\
\midrule
\textbf{Overall} & 50.73 & 4.4 & 59.89 & 194.4 & 60.20 & 251.7 & 60.78 & 251.0 & 63.63 & \underline{179.4} & 63.89 & 182.7 & \textbf{64.91} & 180.8 \\
\bottomrule
\end{tabular}
\end{adjustbox}
\end{table*}

\paragraph{Full instance-level accuracy relative to global routing.}
EDRM-Inst-E (Full) has higher accuracy than Global-E on the four non-reasoning instruction-tuned models. On Llama-3.2-3B-Instruct, accuracy rises from 59.87\% to 63.63\% (+3.76 percentage points), with 28.7\% fewer tokens than CoT (179.4 vs.\ 251.7). The corresponding accuracy differences are +2.77 points for Llama-3.1-8B-Instruct, +4.41 for Qwen2.5-7B-Instruct, and +2.98 for Qwen2.5-32B-Instruct. These gains combine instance-level routing and oracle answer selection; Table~\ref{tab:ablation} separates the latter contribution.

\paragraph{An efficiency spectrum across the three variants.}
The three Full instance-level variants provide different accuracy--efficiency trade-offs: Inst-E reaches 63.63\% accuracy at 179.4 tokens, Inst-C reaches 63.89\% at 182.7 tokens, and MLP reaches 64.91\% at 180.8 tokens. Their accuracies span 1.28 percentage points at similar token cost. All use oracle answer selection. To assess the role of the representation separately, Appendix~\ref{app:ablation-mlp} compares 3D, 64D, and 67D MLP inputs under matched label and fallback settings. The competitive 3D results support the compact descriptors as an effective routing representation, although the preferred input varies by model.

\paragraph{Task-dependent compute allocation.}
Instance-level EDRM allocates different decoding modes across queries. On \texttt{BBH}, Inst-E (Full) reaches 58.87\% accuracy compared with Global-E's 53.55\%; this difference includes the contribution of oracle answer selection. On \texttt{StratQA}, MLP (Full) reduces token use from CoT's 225.9 to 119.4. These examples illustrate the allocation patterns, while the w/o-FB and Selected-FB results assess their practical accuracy trade-offs.

\subsection{Statistical Analysis}
\label{sec:stat_analysis}

We complement the main results with two layers of statistical analysis to assess significance: instance-level per-sample bootstrap and benchmark-level paired permutation tests.

\begin{table}[t]
\centering
\caption{EDRM-Inst-E on four models. Top: benchmark-level paired permutation tests ($n{=}15$, $B{=}10000$), with unweighted benchmark-mean accuracy differences $\Delta_{\mathrm{bench}}$ in percentage points. Bottom: pooled accuracy (\%) / average tokens; accuracy differences correspond to $\Delta_{\mathrm{inst}}$. For Llama-3.2-3B-Instruct w/o-FB vs.\ CoT, $\Delta_{\mathrm{bench}}=-0.88$, whereas the unrounded pooled difference is $\Delta_{\mathrm{inst}}=-1.32$ (subtracting the displayed accuracies gives $58.87-60.20=-1.33$ because of rounding). Token savings compare w/o-FB with CoT. Full uses oracle answer selection; Selected-FB shares its candidate-generation budget. Token counts include amortized probe overhead. Definitions and confidence intervals appear in Appendix~\ref{app:bootstrap}. Model abbreviations: Llama-3.2-3B = Llama-3.2-3B-Instruct; Llama-3.1-8B = Llama-3.1-8B-Instruct; Qwen2.5-7B = Qwen2.5-7B-Instruct; Qwen3-4B-2507 = Qwen3-4B-Instruct-2507.}
\label{tab:ablation}
\begin{adjustbox}{max width=\textwidth}
\small
\setlength{\tabcolsep}{5pt}
\renewcommand{\arraystretch}{1.15}
\begin{tabular}{lcccccc}
\toprule
& \multicolumn{2}{c}{\textbf{w/o-FB vs. CoT}} & \multicolumn{2}{c}{\textbf{Full vs. CoT}} & \multicolumn{2}{c}{\textbf{Token savings}} \\
\cmidrule(lr){2-3}\cmidrule(lr){4-5}\cmidrule(lr){6-7}
\textbf{Model} & $\Delta$Acc & $p$ & $\Delta$Acc & $p$ & \% & $p$ \\
\midrule
Llama-3.2-3B & $-$0.88 & 0.13 (ns) & $+$3.86 & 0.005 ($^{**}$) & 29.7\% & $<0.001^{***}$ \\
Llama-3.1-8B & $-$2.42 & 0.014 ($^{*}$) & $+$2.63 & 0.11 (ns) & 46.1\% & $<0.001^{***}$ \\
Qwen2.5-7B & $-$0.38 & 0.54 (ns) & $+$4.04 & 0.014 ($^{*}$) & 27.1\% & $<0.001^{***}$ \\
Qwen3-4B-2507 & $-$4.01 & $<0.001$ ($^{***}$) & $-$2.12 & 0.025 ($^{*}$) & 38.1\% & $<0.001^{***}$ \\
\bottomrule
\end{tabular}
\end{adjustbox}
\par\smallskip
\begin{adjustbox}{max width=\textwidth}
\small
\setlength{\tabcolsep}{5pt}
\renewcommand{\arraystretch}{1.15}
\begin{tabular}{lccccc}
\toprule
\textbf{Model} & \textbf{Full} & \textbf{Selected-FB} & \textbf{w/o-FB} & \textbf{w/o $S_H$} & \textbf{w/o $a_{\text{vnr}}$} \\
\midrule
Llama-3.2-3B & 63.63 / 179.4 & 59.59 / 179.4 & 58.87 / 177.1 & 59.18 / 188.1 & 58.08 / 190.5 \\
Llama-3.1-8B & 70.04 / 153.9 & 65.44 / 153.9 & 65.11 / 149.8 & 65.36 / 155.4 & 65.08 / 248.2 \\
Qwen2.5-7B & 76.07 / 242.7 & 71.77 / 242.7 & 71.63 / 240.7 & 71.47 / 250.1 & 70.89 / 249.5 \\
Qwen3-4B-2507 & 78.54 / 401.1 & 76.71 / 401.1 & 76.81 / 397.9 & 77.71 / 441.0 & 76.09 / 401.1 \\
\bottomrule
\end{tabular}
\end{adjustbox}
\end{table}

\paragraph{Statistical analysis (top block of Table~\ref{tab:ablation}).}
We apply two complementary tests with different aggregation units. (1)~Instance-level bootstrap ($n{=}23061$/model, 1000 resamples) uses the pooled per-sample accuracy difference $\Delta_{\mathrm{inst}}$, consistent with the main-table Overall rows. It finds a 1.7--4.9-percentage-point Full--w/o-FB accuracy gap across the four models ($p{<}0.01$); w/o-FB accuracy is 0.8--3.9 points below CoT ($p\leq0.01$). (2)~Benchmark-level paired permutation tests ($n{=}15$, $B{=}10000$) use the unweighted mean of per-benchmark differences, $\Delta_{\mathrm{bench}}$. These differences can vary from the pooled ones because benchmark sizes differ: for Llama-3.2-3B-Instruct w/o-FB vs.\ CoT, $\Delta_{\mathrm{inst}}=-1.32$ but $\Delta_{\mathrm{bench}}=-0.88$. Statistical significance can also differ: the corresponding bootstrap $p$-value is 0.001, whereas the benchmark-level permutation $p$-value is 0.13. The benchmark-level tests find significant token savings for w/o-FB on all four models (27--46\%, all $p{<}0.001$). Global-E savings are significant on the four non-reasoning instruction-tuned models and Qwen3-4B-Instruct-2507 (37--55\%, all $p{<}0.001$); Global offers little or no savings on DeepSeek-R1-Distill-Qwen-7B and Qwen3.5-35B-A3B (Table~\ref{tab:overall_comparison}). Benchmark-mean accuracy is comparable to CoT on Llama-3.2-3B-Instruct and Qwen2.5-7B-Instruct, with differences of 0.88 and 0.38 percentage points, respectively; the corresponding permutation tests do not detect significant differences ($p{>}0.1$). Full exceeds CoT in the benchmark-level analysis on Llama-3.2-3B-Instruct (+3.86 points, $p{=}0.005$) and Qwen2.5-7B-Instruct (+4.04 points, $p{=}0.014$), with these gains including oracle answer selection. Appendix~\ref{app:bootstrap} defines both aggregation levels and reports their results separately.

\subsection{Ablation and Analysis}

We ablate the core variant EDRM-Inst-E. It relies on two key design elements: A. synergistic routing through the 3D entropy dynamics $(S_H, V_{\text{sp}}, a_{\text{vnr}})$, and B. fallback compensation for decision robustness. We construct 3 ablation variants: (1) \textbf{w/o Fallback}: removing the Direct fallback branch; (2) \textbf{w/o $S_H$}: removing the cumulative-entropy threshold, using only the coupling of $V_{\text{sp}}$ and $a_{\text{vnr}}$; (3) \textbf{w/o $a_{\text{vnr}}$}: further removing volatility, routing with the univariate entropy trend $V_{\text{sp}}$ alone.

\paragraph{Accuracy contribution of fallback answer selection.}
Full exceeds w/o-FB accuracy by 1.7--4.9 percentage points on the four evaluated models (Llama-3.2-3B-Instruct: 4.76; Llama-3.1-8B-Instruct: 4.93; Qwen2.5-7B-Instruct: 4.44; Qwen3-4B-Instruct-2507: 1.73), with a small increase in token cost. This gap measures how often the additional Direct answer can improve correctness under oracle selection. Selected-FB uses the same candidate-generation budget; its results measure the recovery achievable with the evaluated deployable selector, which improves accuracy on three models but slightly reduces it on Qwen3-4B-Instruct-2507.

\paragraph{Complementary roles of the three descriptors.}
The feature ablations support complementary roles for cumulative uncertainty, trend, and local variation. Removing $S_H$ increases token consumption on all four evaluated models relative to Full (e.g., 441.0 vs.\ 401.1 tokens on Qwen3-4B-Instruct-2507), consistent with its role in identifying cases assigned to Direct. Further removing $a_{\text{vnr}}$ reduces accuracy in the reported comparisons (Llama-3.2-3B-Instruct: 1.10 percentage points; Qwen3-4B-Instruct-2507: 1.62 points). This supports using local variation alongside the trend, with the joint descriptors providing an effective accuracy--efficiency trade-off in these experiments.

\paragraph{Hyperparameter and probe-length sensitivity.}
Across $k\in\{0.03,0.05,0.07,0.10\}$, Global-E accuracy varies by at most 0.4 percentage points; $k{=}0.07$ provides a balanced accuracy--cost operating point in the reported comparison (Appendix~\ref{app:k_sensitivity}). The probe-length ablation over $N\in\{16,32,48,64,80,96\}$ shows an accuracy plateau for $N\geq48$, with variation of at most 1.2 percentage points (Appendix~\ref{app:probe_n}). We use $N{=}64$ within this stable range.

\section{Discussion}
\label{sec:discussion}

\paragraph{The MLP router as a non-linear variant.}
EDRM-MLP learns nonlinear decision boundaries, while the heuristic uses a piecewise-linear rule over three descriptors. Appendix~\ref{app:ablation-mlp} compares MLPs with 3D, 64D, and 67D inputs under matched label strategies and fallback settings. The 3D descriptors remain competitive, and richer inputs do not consistently improve performance, supporting the compact representation for effective routing in the evaluated settings. Comparing the same 3D inputs also shows modest accuracy differences between learned and heuristic routing under Full: for Llama-3.2-3B-Instruct, ML-3D reaches 63.72\% versus Inst-E's 63.63\%; for Llama-3.1-8B-Instruct, the figures are 70.36\% and 70.04\%. These comparisons suggest limited additional accuracy benefit from nonlinear routing in these cases, while token costs and the preferred input remain model-dependent.

\paragraph{The potential of higher-order dynamic features.}
The current framework uses a fixed-length $N{=}64$ probe sequence and three hand-crafted descriptors. Future work could explore adaptive trajectory length (dynamically deciding when to stop probing based on early entropy dynamics) and learnable feature fusion (replacing hand-crafted statistics with a lightweight temporal model) to capture finer-grained dynamic patterns.

\paragraph{Agentic and multi-step scenarios.}
EDRM's single-turn probe signal is effective in single-turn QA, but has limitations in multi-step agent scenarios, where the entropy dynamics of each reasoning step may shift due to the outputs of preceding steps. Integrating entropy-dynamic routing into multi-step agent frameworks---deciding at each step whether to trigger deep reasoning---is an important future direction.

\paragraph{API deployment path.}
APIs that expose top-$K$ log probabilities may support approximate entropy trajectories, but the effect of truncation on routing decisions requires evaluation. Probing also retains a generation cost. For APIs without log-probability access, alternative uncertainty proxies, including verbalized confidence, remain a future research direction; their suitability for EDRM has not been established here.

\section{Conclusion}
This work studies the utility of explicit reasoning through early decoding dynamics. Across the evaluated model--task pairs, decreasing early entropy tends to accompany larger CoT gains, while oscillatory or increasing trajectories more often accompany limited gains. Change-point analysis characterizes temporal structure in these trajectories, supporting an empirical regime interpretation. EDRM uses cumulative entropy, rank trend, and local variation to select among Direct, Standard, and CoT through a training-free heuristic or an optional learned router. Across 15 benchmarks and 7 LLMs spanning 3B--35B parameters, its global and instance-level configurations offer 16--55\% token savings at comparable accuracy. The heuristic-router ablation additionally quantifies the accuracy contribution of oracle answer selection, reaching up to 4.9 percentage points over routing without fallback. These findings support early entropy dynamics as a practical signal for selective reasoning and motivate further study of adaptive inference control.

\paragraph{Limitations and Future Work.}
EDRM still requires a short probing stage, introducing moderate overhead compared to pure direct decoding. Our current experiments evaluate open-weight models in the 3B--35B range on text-only tasks, including dense, MoE, and reasoning-distilled architectures; extending the framework to larger-scale, API-only, and multimodal systems remains important future work. As discussed in \S\ref{sec:discussion}, integrating entropy-dynamic routing into multi-step agentic frameworks---where each step's reasoning decision adapts to the evolving context---is a promising direction for improving efficiency in long-horizon multi-turn reasoning.

\bibliography{nips2026_conference}
\bibliographystyle{nips2026_conference}

\newpage
\appendix

\section{Appendix: Methodological Details}
\label{app:meth}
\subsection{Entropy Dynamics Descriptors}
\label{app:descriptors}
This section defines the three entropy descriptors used by EDRM: cumulative entropy $S_H$, rank trend $V_{\text{sp}}$, and local variation $a_{\text{vnr}}$. We describe their computation and explain how they provide complementary summaries of an early decoding trajectory.
The notation follows an informal position--velocity--acceleration analogy; the descriptors themselves are summary statistics of the entropy sequence.

\paragraph{Input to the descriptor extractor.}
For each instance $x$, we run an $N$-step \textit{Standard} probing decode and record the token-level entropy sequence
$E_x = \{H_1(x),\dots,H_T(x)\}$, where $T\le N$ is the number of generated probing tokens before EOS.
In instance-level heuristic routing, if the probe terminates early ($T<N$), we retain the completed \textit{Standard} answer and skip descriptor-based routing. Early termination is an operational stopping condition, not a correctness test.
Otherwise, we compute the three descriptors on $E_x$ with $T=N$.
The full procedure is summarized in Algorithm~\ref{alg:extract_descriptors}.

\begin{algorithm}[htbp]
\caption{Extracting entropy dynamics descriptors $(S_H, V_{\text{sp}}, a_{\text{vnr}})$}
\label{alg:extract_descriptors}
\begin{algorithmic}[1]
\small
\Require Probing entropy sequence $E=\{H_1,\dots,H_T\}$, max length $N$, small constant $\epsilon$
\Ensure Descriptors $(S_H, V_{\text{sp}}, a_{\text{vnr}})$
\If{$T < N$}
    \State \Return \textsc{EarlyStop} \Comment{route to Standard; skip descriptor-based routing}
\EndIf
\State $S_H \leftarrow \sum_{i=1}^{N} H_i$
\State $V_{\text{sp}} \leftarrow \mathrm{Spearman}(\{1,\dots,N\}, \{H_1,\dots,H_N\})$
\State $\bar{H} \leftarrow \frac{1}{N}\sum_{i=1}^{N} H_i$
\State $\mathrm{Var} \leftarrow \frac{1}{N}\sum_{i=1}^{N}(H_i-\bar{H})^2$
\State $\mathrm{MSD} \leftarrow \frac{1}{N-1}\sum_{i=1}^{N-1}(H_{i+1}-H_i)^2$
\State $a_{\text{vnr}} \leftarrow \frac{\mathrm{MSD}}{\mathrm{Var}+\epsilon}$
\State \Return $(S_H, V_{\text{sp}}, a_{\text{vnr}})$
\end{algorithmic}
\end{algorithm}

\subsubsection{Cumulative Entropy \texorpdfstring{$S_H$}{S H}}
The cumulative entropy $S_H$ sums next-token uncertainty over the $N$-step probe. For an input $x$, let $p_i(\cdot)$ be the next-token distribution and $H_i(x)$ its entropy at step $i$. We compute
\begin{equation}
    S_H(x) = \sum_{i=1}^{N} H_i(x),\qquad
    H_i(x) = -\sum_{v \in \mathcal{V}} p_i(v)\log p_i(v).
    \label{eq:sh_def}
\end{equation}
Here $\mathcal{V}$ is the vocabulary and $H_i(x)$ is the Shannon entropy of the next-token distribution at step $i$.

\paragraph{Interpretation.}
$S_H$ measures cumulative next-token uncertainty during the probe. Larger values indicate greater uncertainty over the observed trajectory. The router uses this descriptor together with a positive entropy trend to identify cases assigned to Direct; cumulative entropy alone is not a correctness criterion.

\paragraph{Why $S_H$ instead of other cumulative summaries?}
For fixed $N$, cumulative entropy and mean entropy contain the same information: $S_H=N\bar{H}$. We use the cumulative form as a summary of the uncertainty accumulated during the probe; a mean-entropy implementation can use an equivalently rescaled threshold. When $N$ changes, the threshold must be recalibrated (Appendix~\ref{app:probe_n}). The maximum entropy emphasizes a single step, while the endpoint difference ignores intermediate behavior and depends strongly on the first and last values. We therefore combine cumulative entropy with separate trend and local-variation descriptors.

\subsubsection{Univariate Trend \texorpdfstring{$V_{\text{sp}}$}{V sp}}
\paragraph{Definition.}
The trend descriptor $V_{\text{sp}}$ quantifies whether entropy tends to decrease or increase over time.
We define it as the Spearman rank correlation between the step index and entropy values:
\begin{equation}
V_{\text{sp}}(x) = \mathrm{Spearman}(\{1,\dots,N\}, \{H_1(x),\dots,H_N(x)\}).
\label{eq:vsp_def}
\end{equation}
Operationally, let $r_i$ be the rank of $H_i$ among $\{H_1,\dots,H_N\}$ (ties are assigned average ranks). Then
\begin{equation}
V_{\text{sp}} = \mathrm{Corr}\left(\{1,\dots,N\}, \{r_1,\dots,r_N\}\right),
\end{equation}
where $\mathrm{Corr}(\cdot,\cdot)$ is the Pearson correlation computed on the rank-transformed sequence.
By construction, $V_{\text{sp}}\in[-1,1]$ for a nonconstant trajectory: negative values indicate a decreasing entropy trend and positive values an increasing trend.

\paragraph{Interpretation.}
$V_{\text{sp}}$ serves as the ``velocity'' signal in our kinematic analogy: it captures \emph{directional evolution} rather than magnitude.
In the evaluated tasks, larger CoT gains tend to accompany decreasing early entropy ($V_{\text{sp}}<0$). Weak or negative gains more often accompany increasing or oscillatory trajectories. Near-zero correlation alone does not identify oscillation, so we use $a_{\text{vnr}}$ to describe local variation separately.

\paragraph{Why Spearman instead of common trend metrics?}
We choose Spearman correlation over several alternatives:
\begin{itemize}[leftmargin=*]
    \item \textbf{Linear regression slope} $\hat{\beta}$ of $H_i$ on $i$: while it provides a signed trend, it is \emph{scale-dependent} and sensitive to outliers (single entropy spikes can dominate $\hat{\beta}$). Moreover, entropy trajectories are frequently non-linear (piecewise or curved), making a slope a fragile summary.
    \item \textbf{Pearson correlation} $\mathrm{Corr}(i, H_i)$: it summarizes linear association and is sensitive to magnitude outliers; it can overreact to a few extreme points.
    \item \textbf{Average increment} $\frac{1}{N-1}\sum_{i=2}^N(H_i-H_{i-1})$: the sum telescopes to $(H_N-H_1)/(N-1)$, so this statistic depends only on the endpoints and ignores intermediate behavior.
\end{itemize}
In contrast, $V_{\text{sp}}$ is \emph{non-parametric} (depends only on ordering), robust to non-linear monotone patterns, and less sensitive to occasional spikes/dips. This makes it well-suited for noisy entropy trajectories produced by early decoding.

\subsubsection{Volatility \texorpdfstring{$a_{\text{vnr}}$}{a vnr} (Von Neumann Ratio)}
\paragraph{Definition.}
We instantiate the volatility descriptor using the \emph{Von Neumann ratio} (VNR), defined as the ratio between the mean square successive difference (MSD) and the (population) variance of the sequence:
\begin{equation}
a_{\text{vnr}}(x)
=\frac{\frac{1}{N-1}\sum_{i=1}^{N-1}\big(H_{i+1}(x)-H_i(x)\big)^2}
{\frac{1}{N}\sum_{i=1}^{N}\big(H_i(x)-\bar{H}(x)\big)^2+\epsilon},
\label{eq:avnr_def}
\end{equation}
where $\bar{H}(x)=\frac{1}{N}\sum_{i=1}^{N}H_i(x)$ and $\epsilon$ is a small constant (e.g., $10^{-8}$). If the variance is zero, we set $a_{\text{vnr}}(x)=0$.

\paragraph{Interpretation.}
$a_{\text{vnr}}$ measures successive squared changes relative to the overall variance of the sequence. Larger values indicate greater local variation relative to global dispersion; smaller values indicate smoother temporal evolution.
Unlike variance alone, the numerator depends on token order and therefore distinguishes trajectories with similar global dispersion but different successive changes.
In routing, $a_{\text{vnr}}$ plays a critical role as a stability prior: trend estimates (e.g., $V_{\text{sp}}$) are less reliable under high volatility, so we couple them via the ratio $V_{\text{sp}}/a_{\text{vnr}}$ in the heuristic rule.

\paragraph{Why this volatility form instead of standard volatility metrics?}
We prefer $a_{\text{vnr}}$ over several common choices:
\begin{itemize}[leftmargin=*]
    \item \textbf{Standard deviation} $\sigma$: captures global dispersion but is scale-dependent and can be inflated by a few extreme points; used alone, it cannot distinguish ``smoothly high'' entropy from ``highly oscillatory'' entropy.
    \item \textbf{Mean absolute increment} $\frac{1}{N-1}\sum_{i=2}^N |H_i-H_{i-1}|$: focuses on local changes, but it is also sensitive to deterministic trends (a steady monotone decrease can still yield large increments), thus conflating trend with volatility.
    \item \textbf{MAD alone} $\frac{1}{N}\sum_i |H_i-\bar{H}|$: more robust than $\sigma$ but still scale-dependent and does not explicitly emphasize bursty oscillations.
\end{itemize}
The VNR normalizes mean squared successive differences by variance, making it approximately invariant to rescaling when $\epsilon$ is negligible. Coupling $V_{\text{sp}}$ with $a_{\text{vnr}}$ reduces the weight assigned to a trend when successive variation is large.

\subsection{Comparison of CoT and Direct Decoding across All Benchmarks}
\label{app:compare}
\begin{figure}
    \centering
    \includegraphics[width=\textwidth]{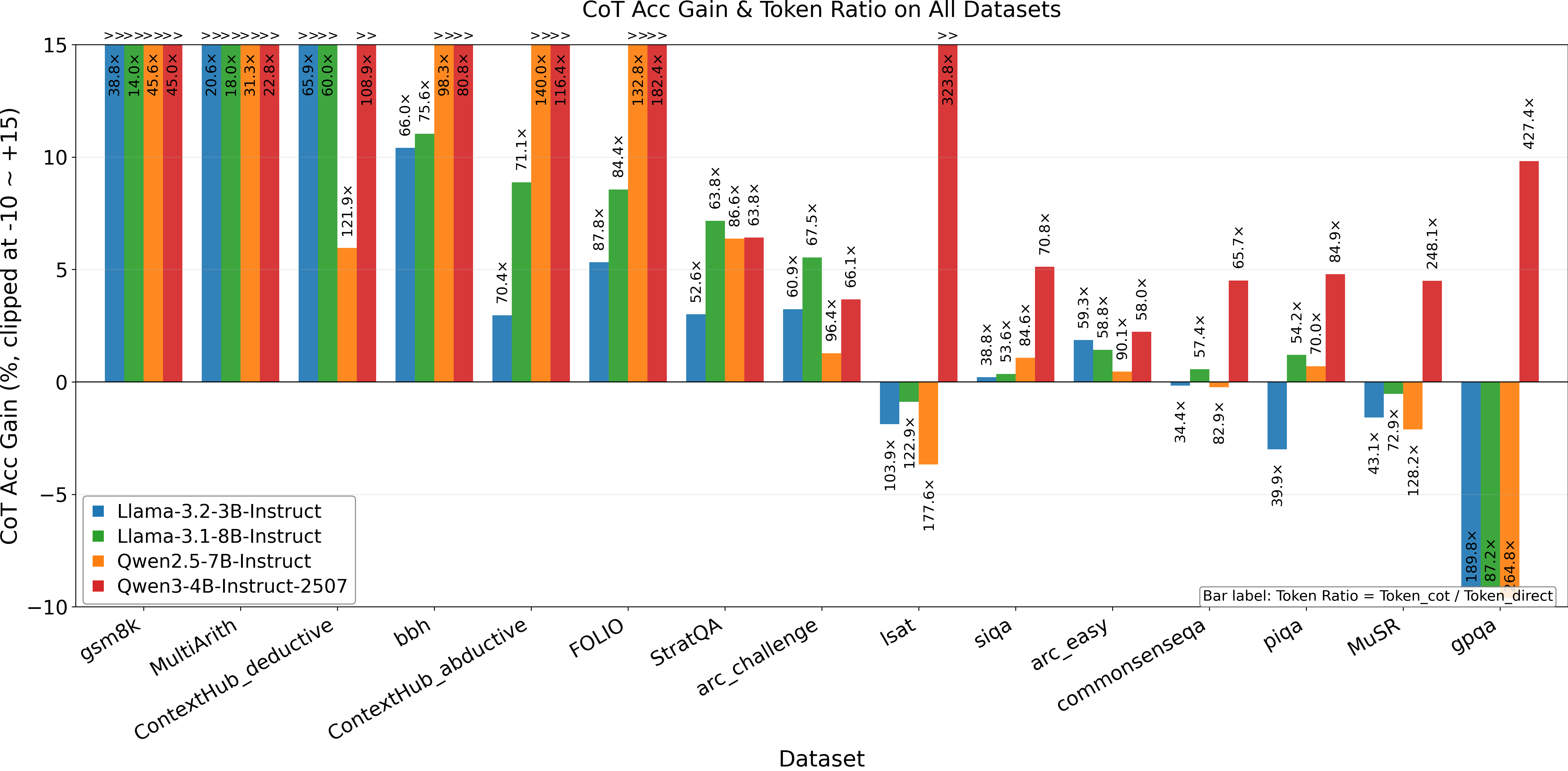}
    \caption{CoT accuracy gains over Direct across 15 benchmarks for four models. Bar colors identify models; labels give the ratio of CoT to Direct output tokens. Accuracy gains are clipped to the displayed range of $[-10,15]$ percentage points. The checkpoints are Llama-3.2-3B-Instruct, Llama-3.1-8B-Instruct, Qwen2.5-7B-Instruct, and Qwen3-4B-Instruct-2507. The plot label ``Qwen3-4B'' denotes Qwen3-4B-Instruct-2507.}
    \label{fig:cot_direct_comparison}
\end{figure}

Figure~\ref{fig:cot_direct_comparison} compares CoT accuracy gains over Direct on all 15 benchmarks for Llama-3.2-3B-Instruct, Llama-3.1-8B-Instruct, Qwen2.5-7B-Instruct, and Qwen3-4B-Instruct-2507. The numerical comparison extends to all seven evaluated models in Table~\ref{tab:overall_comparison} and the per-benchmark result tables.

\paragraph{Mathematical Reasoning.}
On mathematical benchmarks such as \texttt{GSM8K} and \texttt{MultiArith}, CoT substantially improves accuracy over Direct in the evaluated models. Some gains reach the upper limit of the visualization scale, indicating that the plot clips their magnitude. These results support the usefulness of explicit reasoning on these benchmarks.

\paragraph{Logical Deduction and Complex Tasks.}
In logical reasoning tasks like \texttt{CH-ded}, \texttt{bbh}, and \texttt{CH-abd}, \textit{CoT} generally provides a positive boost, though the magnitude varies by model. Notably, the Qwen3-4B-Instruct-2507 model (red bars) demonstrates exceptional robustness, achieving the highest CoT gains in several of these categories, often surpassing larger models like Llama-3.1-8B-Instruct and Qwen2.5-7B-Instruct. This suggests that Qwen3-4B-Instruct-2507 is particularly effective at leveraging reasoning paths for deduction.

\paragraph{General Knowledge and Common Sense.}
For benchmarks involving common sense or general knowledge (e.g., \texttt{siqa}, \texttt{arc-e}, \texttt{csqa}), the advantages of \textit{CoT} are much more marginal. Most models show only slight improvements or negligible differences compared to direct decoding. In some cases, such as \texttt{lsat}, \texttt{gpqa}, and \texttt{StrategyQA}, several models (particularly Llama-3.2-3B-Instruct, Llama-3.1-8B-Instruct, and Qwen2.5-7B-Instruct) exhibit negative gains, suggesting that for these specific tasks, generating a reasoning trace may introduce noise or errors, making direct decoding the superior strategy.

\paragraph{Model Comparison.}
A key observation is the consistent performance of Qwen3-4B-Instruct-2507. While other models fluctuate between positive and negative gains depending on the benchmark, Qwen3-4B-Instruct-2507 maintains positive CoT gains across almost all tasks, including those where other models struggle (e.g., \texttt{gpqa}). This indicates a strong generalization capability in utilizing Chain-of-Thought reasoning across diverse domains.

\subsection{Entropy Trajectories across Benchmarks}
\label{app:visual}
\begin{figure}
    \centering
    \includegraphics[width=\textwidth]{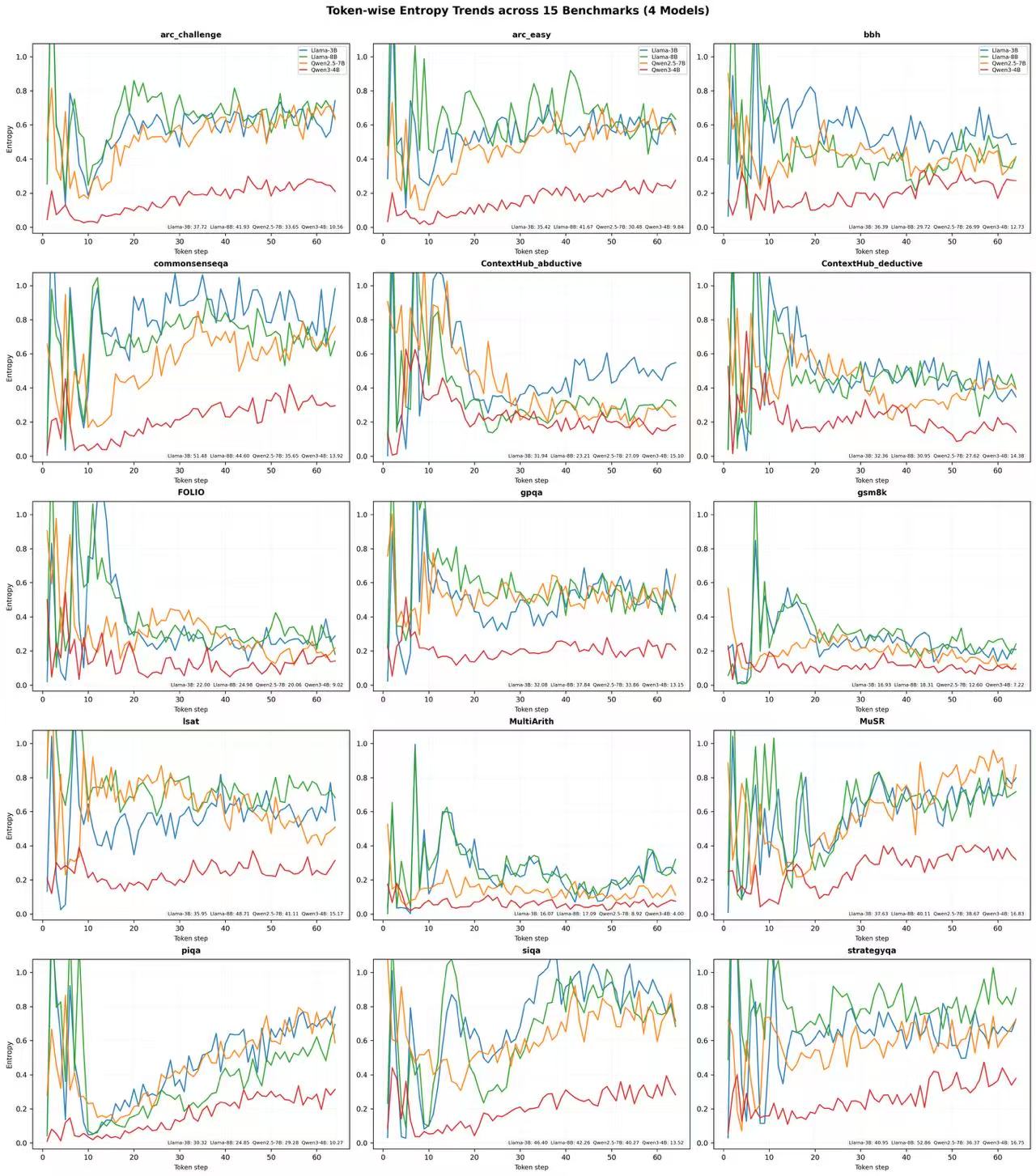}
    \caption{Mean entropy trajectories across 15 benchmarks for Llama-3.2-3B-Instruct, Llama-3.1-8B-Instruct, Qwen2.5-7B-Instruct, and Qwen3-4B-Instruct-2507. Each subplot corresponds to one benchmark; colors identify models. The plot labels ``Llama-3B'', ``Llama-8B'', ``Qwen2.5-7B'', and ``Qwen3-4B'' denote those four checkpoints, respectively.}
    \label{fig:entropy_trajectories}
\end{figure}

Figure~\ref{fig:entropy_trajectories} visualizes the entropy trajectories of Llama-3.2-3B-Instruct, Llama-3.1-8B-Instruct, Qwen2.5-7B-Instruct, and Qwen3-4B-Instruct-2507, revealing different uncertainty dynamics across these models. Qwen3-4B-Instruct-2507 (red line) consistently maintains the lowest entropy levels across almost all benchmarks, exhibiting a stable trajectory with minimal fluctuation. This indicates high confidence in token generation. Conversely, Llama-3.2-3B-Instruct (blue) and Llama-3.1-8B-Instruct (green) generally display higher entropy with significant volatility, particularly in the early decoding steps. Qwen2.5-7B-Instruct (orange) typically falls between these extremes.

\begin{figure}
    \centering
    \includegraphics[width=\textwidth]{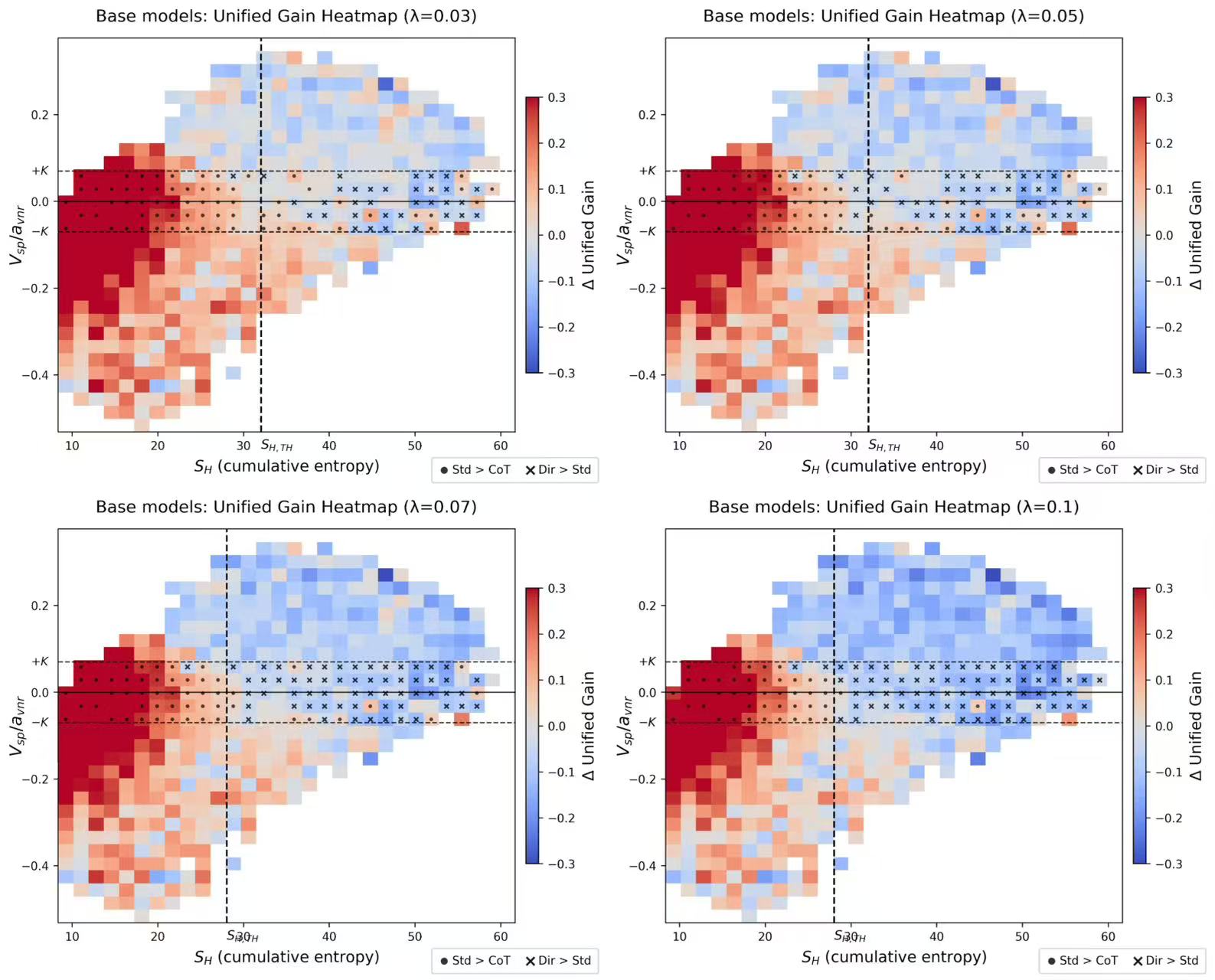}
    \caption{Unified Gain Heatmap with various $\lambda$ (0.03,0.05,0.07,0.10) on \textbf{non-reasoning instruction-tuned models}.}
    \label{fig:heatbase4}
\end{figure}

\begin{figure}
    \centering
    \includegraphics[width=\textwidth]{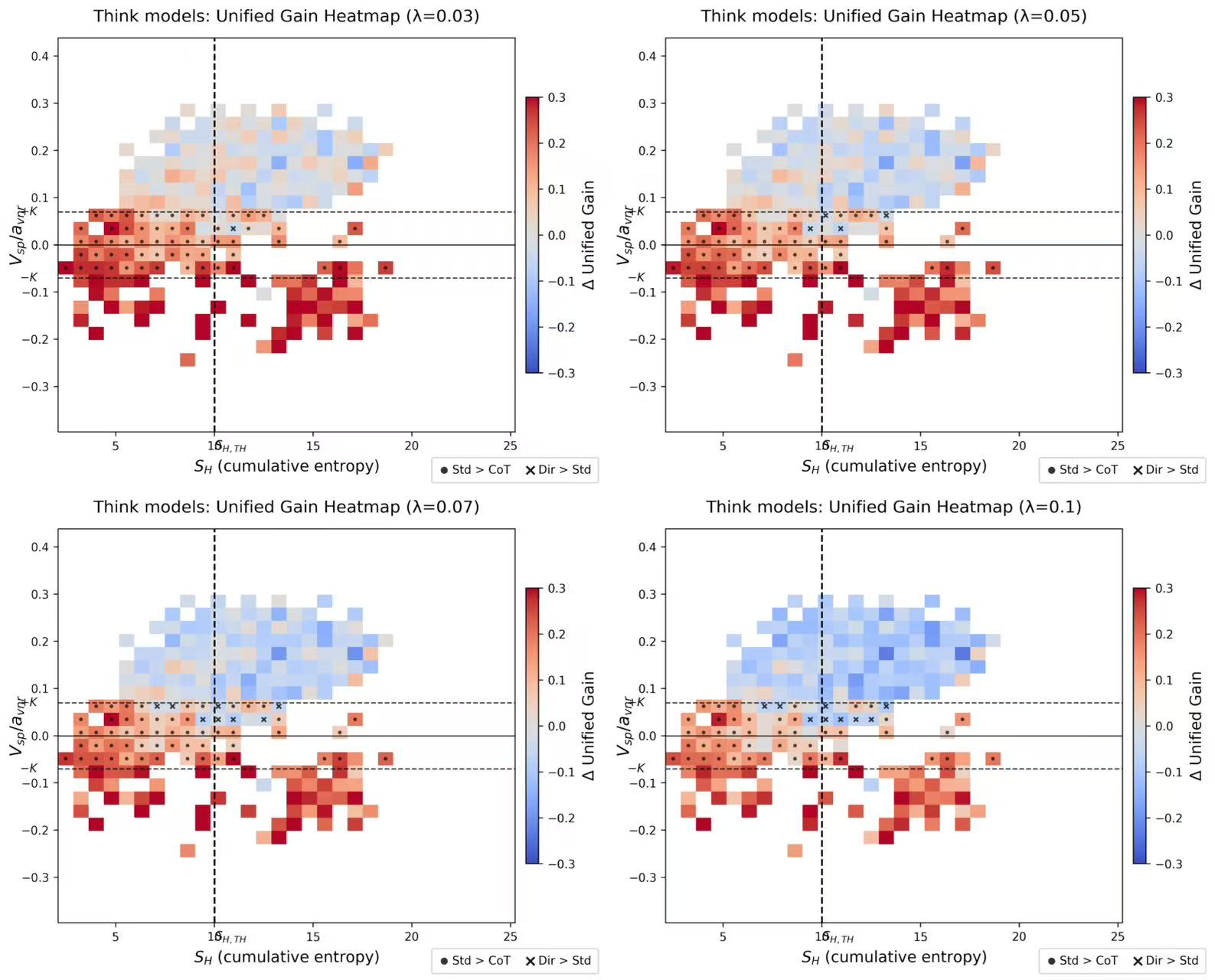}
    \caption{Unified Gain Heatmap with various $\lambda$ (0.03,0.05,0.07,0.10) on \textbf{models with verbose generation}. The plot label ``Think models'' refers to this evaluation group, which includes the non-thinking Qwen3-4B-Instruct-2507 checkpoint.}
    \label{fig:heatthink4}
\end{figure}

\subsection{Unified Gain Heatmap}
\label{app:unified_gain}
Figures~\ref{fig:heatbase4} and~\ref{fig:heatthink4} show how the relative utility of \textit{CoT} and \textit{Direct} varies across regions of the entropy feature space. To account for the generation cost of explicit reasoning, we define a \emph{Unified Gain} that combines correctness and token usage.

\paragraph{Motivation: why unified gain?}
Standard comparisons such as $\mathrm{Acc}(\text{CoT})-\mathrm{Acc}(\text{Direct})$ ignore token usage, yet \textit{CoT} often increases output length by orders of magnitude (Tables~\ref{tab:overall_comparison}, \ref{tab:llama32_results}). For deployment, the relevant question is not only ``which is more accurate?'' but ``which is more cost-effective under a chosen accuracy--cost trade-off?'' Unified Gain provides a single scalar objective aligned with this requirement, enabling a fine-grained visualization over the manifold coordinates.

\paragraph{Relation between instance-level and aggregate utility.}
An aggregate utility can be written as $U(m)=\mathrm{Acc}(m)-\lambda\cdot \mathrm{Cost}(m)$ for readability, where $\mathrm{Acc}(m)$ and $\mathrm{Cost}(m)$ can be understood as \emph{already-aggregated} statistics (e.g., dataset-level averages).
In this appendix, we restate the same objective at the \emph{instance level} to remove ambiguity and to support heatmap construction:
we first define a per-instance utility $U_x(m)=\mathbb{I}[\mathrm{correct}_m(x)]-\lambda\cdot \tilde{T}_m(x)$, and then aggregate by expectation/average over a set of instances.
This formulation makes explicit (i) that accuracy corresponds to the mean of correctness indicators, and (ii) what token cost means and how it is normalized.
When $\mathrm{Acc}(m)=\mathbb{E}_x[\mathbb{I}[\mathrm{correct}_m(x)]]$ and $\mathrm{Cost}(m)=\mathbb{E}_x[\tilde{T}_m(x)]$, averaging the per-instance utility gives the aggregate utility above.
We normalize tokens as $\tilde{T}_m(x)=T_m(x)/1000$ so that $\lambda$ is comparable across datasets and models and remains numerically interpretable.

\paragraph{Definition and computation.}
For an instance $x$ and decoding mode $m\in\{\text{Direct},\text{CoT}\}$, let $\mathbb{I}[\text{correct}_m(x)]\in\{0,1\}$ be the correctness indicator and $T_m(x)$ be the number of generated output tokens.\footnote{In our experiments, $T_m(x)$ corresponds to the recorded output token length in the evaluation logs.}
We define the per-instance utility as
\begin{equation}
U_x(m)=\mathbb{I}[\text{correct}_m(x)]-\lambda\cdot \frac{T_m(x)}{1000},
\qquad
\Delta U_x(\text{CoT},\text{Direct})=U_x(\text{CoT})-U_x(\text{Direct}),
\label{eq:unified_gain}
\end{equation}
where $\lambda\ge 0$ is a cost-penalty coefficient and the token cost is normalized by $1000$ to keep $\lambda$ in a human-interpretable range.\footnote{The main paper figure uses $\lambda=0.05$ for non-reasoning instruction-tuned models as a representative operating point. Larger $\lambda$ increasingly favors shorter outputs; smaller $\lambda$ approaches accuracy-only comparison.}
A positive $\Delta U_x$ means \textit{CoT} is more cost-effective than \textit{Direct} on $x$ under the chosen $\lambda$, and a negative value means the opposite.

\paragraph{From per-instance gain to a heatmap over the manifold.}
To construct Figure~\ref{fig:heatmap}, we compute each instance's manifold coordinates using \textit{Standard} probing:
\[
\left(\frac{V_{\text{sp}}(x)}{a_{\text{vnr}}(x)},\, S_H(x)\right),
\]
and then discretize the 2D plane into bins. Each cell reports the average unified gain among instances that fall into the corresponding bin:
\[
\Delta U_{\text{cell}}=\mathbb{E}\left[\Delta U_x(\text{CoT},\text{Direct}) \mid x \in \text{cell}\right].
\]
This visualization directly links the observed efficiency frontier to entropy dynamics, without requiring any learned router.

\paragraph{How to read the heatmap (interpretation of the main-paper figure).}
The heatmap in Figure~\ref{fig:heatmap} shows a structured separation aligned with our routing logic:
\begin{itemize}[leftmargin=*]
    \item \textbf{Convergent region (large negative $V_{\text{sp}}/a_{\text{vnr}}$):} cells tend to have higher (often positive) $\Delta U$, indicating that when entropy decreases reliably relative to volatility, \textit{CoT} is more likely to justify its extra tokens.
    \item \textbf{Divergent/unstable region (positive $V_{\text{sp}}/a_{\text{vnr}}$):} cells are dominated by negative $\Delta U$, consistent with the observation that \textit{CoT} often overthinks and wastes tokens when uncertainty drifts upward.
    \item \textbf{Uncertainty-load effect ($S_H$):} even when trend is not strongly positive, very large $S_H$ corresponds to early-stage uncertainty overload, where the additional \textit{CoT} budget is less likely to convert into correctness, pushing $\Delta U$ downward.
\end{itemize}
The heatmap links regions of the two-axis representation $(V_{\text{sp}}/a_{\text{vnr}},S_H)$ to differences in cost-adjusted CoT utility, motivating its use for routing.


\subsubsection{Cross-dataset Calibration of \texorpdfstring{$S_{H,\text{th}}$}{S H threshold}}
\label{app:calibration}

\paragraph{Why calibration is needed?}
The uncertainty-overload threshold $S_{H,\text{th}}$ is \emph{model-dependent}. While per-model empirical thresholds (e.g., $S_{H,\text{th}}{=}32$ for the non-reasoning Llama-3.2-3B-Instruct, Llama-3.1-8B-Instruct, and Qwen2.5-7B-Instruct models, $39$ for Qwen2.5-32B-Instruct, and $10/30/20$ for the models exhibiting verbose generation in our experiments) can work well for the models studied in this paper, they are obtained by analyzing a closed set of model--dataset results and therefore may not transfer to a new model with different calibration, decoding behavior, or built-in ``thinking'' biases. Repeating a full benchmark sweep to re-tune $S_{H,\text{th}}$ for every new model is costly and often impractical in cold-start deployment. To address this, we introduce a lightweight \emph{calibration} procedure used by EDRM-Global/Inst-C: for any unseen model (and optionally a new collection of datasets), it estimates a suitable $S_{H,\text{th}}$ from a small number of sampled instances via a simple heuristic, enabling fast and flexible threshold selection without additional training or exhaustive evaluation.

This section describes cross-dataset heuristic calibration for EDRM-Global/Inst-C. For a fixed model and $J$ datasets $\{\mathcal{D}_j\}_{j=1}^{J}$, we sample $n$ instances per dataset, run $N$-step probes, and compute the dataset-level feature means
\begin{equation}
\mu_S^{(j)}=\frac{1}{n}\sum_{x\in \mathcal{D}^{(j)}_{\text{sample}}} S_H(x),\qquad
\mu_V^{(j)}=\frac{1}{n}\sum_{x\in \mathcal{D}^{(j)}_{\text{sample}}} V_{\text{sp}}(x).
\end{equation}
Let
\begin{equation}
M=\left|\left\{j\in\{1,\dots,J\}:\mu_V^{(j)}<0\right\}\right|
\end{equation}
be the number of datasets whose average probing dynamics are convergent ($\mu_V^{(j)}<0$). Sort $\{\mu_S^{(j)}\}_{j=1}^{J}$ in ascending order to obtain $s_{(1)}\le \dots \le s_{(J)}$, and define
\begin{equation}
S_{H,\text{th}}=\left\lfloor s_{(M)}\right\rfloor,
\end{equation}
where we clamp the index as $M\leftarrow \max(1,\min(M,J))$ to handle boundary cases.
The full calibration procedure is presented in Algorithm~\ref{alg:calib_sh}.

\begin{algorithm}[htbp]
\caption{Heuristic Calibration of $S_{H,\text{th}}$ Across Datasets}
\label{alg:calib_sh}
\begin{algorithmic}[1]
\Require Datasets $\{\mathcal{D}_j\}_{j=1}^J$, per-dataset sample size $n$, probe length $N$
\Ensure Calibrated threshold $S_{H,\text{th}}$
\For{$j=1$ \textbf{to} $J$}
    \State Sample $\mathcal{D}^{(j)}_{\text{sample}} \subset \mathcal{D}_j$ with $|\mathcal{D}^{(j)}_{\text{sample}}|=n$
    \For{\textbf{each} $x\in \mathcal{D}^{(j)}_{\text{sample}}$}
        \State Run $N$-step probing on $x$ to obtain entropy sequence $E_x$
        \State Compute $S_H(x)$ and $V_{\text{sp}}(x)$
    \EndFor 
    \State $\mu_S^{(j)}\leftarrow \frac{1}{n}\sum_{x\in \mathcal{D}^{(j)}_{\text{sample}}} S_H(x)$
    \State $\mu_V^{(j)}\leftarrow \frac{1}{n}\sum_{x\in \mathcal{D}^{(j)}_{\text{sample}}} V_{\text{sp}}(x)$
\EndFor
\State $M \leftarrow \left|\left\{j:\mu_V^{(j)}<0\right\}\right|$
\State Sort $\{\mu_S^{(j)}\}_{j=1}^J$ ascending to obtain $s_{(1)}\le \cdots \le s_{(J)}$
\State $M \leftarrow \max(1, \min(M, J))$
\State $S_{H,\text{th}} \leftarrow \lfloor s_{(M)} \rfloor$
\State \Return $S_{H,\text{th}}$
\end{algorithmic}
\end{algorithm}

\subsection{Routing Decision Logic Analysis}
\label{app:routing_logic}

This section analyzes the design rationale of EDRM's three-branch routing (\textit{Direct}, \textit{Standard}, \textit{CoT}), the decision boundaries in Algorithm~\ref{alg:routing}, and how these boundaries relate to the entropy-dynamics descriptors $(S_H, V_{\text{sp}}, a_{\text{vnr}})$.

\paragraph{Why three branches (Direct / Standard / CoT)?}
We adopt three regimes because they span the practical efficiency--reliability spectrum while remaining deployable across both instruction-tuned models and models with verbose reasoning behavior:
\begin{itemize}[leftmargin=*]
    \item \textbf{Direct (low-cost answer generation).}
    Direct is the lowest-cost mode and is empirically strong on retrieval-heavy or low-reasoning tasks, where step-by-step deliberation often yields marginal or negative gains while substantially increasing token cost. Direct is also a conservative choice under decoding drift risk.
    \item \textbf{Standard (neutral default and probing anchor).}
    Standard uses a minimally biased prompt (no explicit CoT instruction), providing a stable ``native'' decoding state for probing and serving as the middle option when the evidence for either extreme is insufficient.
    \item \textbf{CoT (explicit step-by-step reasoning).}
    CoT is reserved for cases where early decoding shows stable uncertainty reduction (a reliable convergence signature). In such cases, additional structured reasoning is most likely to translate into accuracy gains.
\end{itemize}
This tri-partition is also operationally important: it avoids forcing a binary decision (CoT vs.\ Direct) when the model is neither clearly convergent nor clearly divergent, reducing misrouting on boundary cases.

\paragraph{Decision boundaries and their meaning in the entropy feature embedding.}
EDRM uses two complementary signals: a \emph{direction--stability} coupling via $(V_{\text{sp}}, a_{\text{vnr}})$ and an \emph{uncertainty-load} guardrail via $S_H$.
Concretely, Algorithm~\ref{alg:routing} implements:
\begin{itemize}[leftmargin=*]
    \item \textbf{Divergence / drift region $\Rightarrow$ Direct.}
    If $V_{\text{sp}} > k\cdot a_{\text{vnr}}$, the entropy trajectory has an overall \emph{increasing} tendency, and the confidence does not improve with steps. This is a high drift-risk regime: extra reasoning tokens are likely to amplify exploration or produce verbose but ungrounded continuations. Therefore, we route to \textit{Direct}.
    \item \textbf{Convergence region $\Rightarrow$ CoT.}
    If $V_{\text{sp}} < -k\cdot a_{\text{vnr}}$, the trajectory exhibits a \emph{decreasing trend} relative to its local variation. This indicates progressive uncertainty reduction, where structured reasoning is most likely to be beneficial; we route to \textit{CoT}.
    \item \textbf{Uncertainty overload guardrail $\Rightarrow$ Direct.}
    Even when $V_{\text{sp}}$ is not strongly positive, if $V_{\text{sp}} > 0$ and $S_H > S_{H,\text{th}}$, the model accumulates large total uncertainty during early decoding. Empirically, this region often corresponds to poor calibration or inability to settle on a coherent reasoning path; CoT tends to be long and unreliable. We thus route to \textit{Direct} to cap cost and reduce drift.
    \item \textbf{Otherwise $\Rightarrow$ Standard.}
    When neither strong convergence nor strong divergence is detected, we select \textit{Standard} as the default middle ground.
\end{itemize}

\paragraph{How $k$ and $S_{H,\text{th}}$ interact with the descriptors.}
Parameter $k$ controls how much negative/positive trend is required \emph{relative to volatility} to trigger CoT/Direct.
Intuitively, $a_{\text{vnr}}$ serves as a noise-aware normalization: under high volatility, $|V_{\text{sp}}|$ must be larger to be trusted as a genuine monotone trend.
Threshold $S_{H,\text{th}}$ acts as a capacity-dependent ceiling on early uncertainty load (base vs.\ reasoning models differ), preventing expensive CoT in regimes where the model is already ``lost'' in early decoding.

\paragraph{Why use the dataset mean of $V_{\text{sp}}$ (i.e., $\bar{V}_{\text{sp}}=\mathbb{E}[V_{\text{sp}}(x)]$) instead of computing $V_{\text{sp}}$ on the averaged entropy curve?}
For dataset-level routing, we aggregate descriptors as
$\bar{S}_H=\frac{1}{|\mathcal{D}_s|}\sum_x S_H(x)$,
$\bar{V}_{\text{sp}}=\frac{1}{|\mathcal{D}_s|}\sum_x V_{\text{sp}}(x)$,
$\bar{a}_{\text{vnr}}=\frac{1}{|\mathcal{D}_s|}\sum_x a_{\text{vnr}}(x)$.
A tempting alternative is to first average the token-level entropies across samples,
$\tilde{H}_i=\frac{1}{|\mathcal{D}_s|}\sum_x H_i(x)$,
and then compute $V_{\text{sp}}(\tilde{H}_{1:N})$.
We intentionally avoid this for three reasons:
\begin{itemize}[leftmargin=*]
    \item \textbf{Non-commutativity (nonlinearity) of Spearman.}
    Spearman correlation is computed on \emph{ranks} and is not a linear operator, hence in general
    \[
    \mathrm{Spearman}\big(i, \mathbb{E}[H_i(x)]\big)\;\neq\;\mathbb{E}\big[\mathrm{Spearman}(i, H_i(x))\big].
    \]
    Averaging before ranking can change the ordering structure and distort the monotonicity signal.
    \item \textbf{Heterogeneity cancellation.}
    Many datasets contain mixed instance types (some clearly convergent, others divergent). Averaging token-level entropies across such a mixture often yields an artificially smooth curve that can appear weakly decreasing even when a large fraction of instances are increasing (or vice versa). In contrast, averaging per-instance $V_{\text{sp}}(x)$ summarizes the direction and strength of instance-level trends, which is exactly what dataset-level routing aims to summarize.
    \item \textbf{Robustness to irregularities and early termination.}
    Some instances terminate before $N$ tokens and are assigned to \textit{Standard} without descriptor-based routing. Averaging their trajectories would require alignment or padding choices. Computing $V_{\text{sp}}(x)$ on valid $N$-token probes and then averaging follows the feature computation used by the router.
\end{itemize}
Empirically, we find $\bar{V}_{\text{sp}}$ computed as the mean of per-instance trends aligns better with dataset-level CoT gains and yields more stable routing under sampling (cf.\ low variances reported in Tables~\ref{tab:llama32_results}).



\subsection{MLP Variants Design and Training Details}
\label{app:mlp_details}
This subsection details the data construction, variant design, and training setup for \textbf{EDRM-MLP}. The router is trained as an instance-level classifier over three regimes $\mathcal{M}\in\{\textit{Direct},\textit{Standard},\textit{CoT}\}$ using entropy-dynamics signals extracted from an $N{=}64$-step \textit{Standard} probing decode.

\paragraph{Training targets (multi-label vs. priority single-label).}
For each instance, we evaluate all three decoding regimes and derive correctness indicators
$\mathbf{y}=[y_D,y_S,y_C]\in\{0,1\}^3$, where $y_m{=}1$ iff mode $m$ answers correctly.
This yields the \textbf{original multi-label target} (possibly multi-hot, e.g., $[1,1,0]$; or $[0,0,0]$ when all fail).
To incorporate a token-efficiency preference, we also construct a \textbf{priority-constrained single-label target} $\mathbf{y}'$ by mapping multi-hot labels to a one-hot vector using the rule \textit{Direct}$>$\textit{Standard}$>$\textit{CoT}; samples with $[0,0,0]$ are filtered out in this setting.

\paragraph{Input variants (3D / 64D / 67D).}
All variants share the same probing source: we take the token-level entropy sequence $E=\{H_1,\dots,H_T\}$ recorded in \textit{Standard} decoding logs and align it to a fixed length $N{=}64$ by truncation or zero-padding:
if $T\ge N$, keep the first $N$ entropies; otherwise pad $E$ with zeros to length $N$.
We then form three input representations:
(i) \textbf{3D descriptors}: $(S_H, V_{\text{sp}}, a_{\text{vnr}})$, where $S_H=\sum_{i=1}^{N}H_i$, $V_{\text{sp}}$ is the Spearman trend, and $a_{\text{vnr}}$ is the Von Neumann ratio;
(ii) \textbf{64D trajectory}: the aligned entropy vector $(H_1,\dots,H_N)$;
(iii) \textbf{67D hybrid}: concatenation of the 64D trajectory and the 3D descriptors.
Combining the two label strategies with the three input forms yields the six ablation variants used in our experiments.

\paragraph{Data construction and split.}
We build per-instance features and labels by merging: (a) the entropy trajectory from \textit{Standard} probing logs, and (b) per-instance descriptors and correctness labels computed by the EDRM evaluation pipeline.
To keep training lightweight while preserving coverage, we perform \textbf{10\% stratified sampling} for training: for each dataset, we group instances by the 8 possible label patterns (e.g., $000,001,\dots,111$) and sample 10\% from each group (at least one sample when available). The remaining instances are used for testing.

\paragraph{Model and optimization.}
We use a compact 3-layer MLP that outputs 3 logits. Inputs are standardized with a \texttt{StandardScaler} fitted on the training split.
For \textbf{multi-label} training, we use \texttt{BCEWithLogitsLoss} with per-class \texttt{pos\_weight} (computed as neg/pos on the training set) to mitigate label imbalance.
For \textbf{single-label} training, we use \texttt{CrossEntropyLoss} with inverse-frequency class weights.
Unless otherwise specified, we train with Adam, learning rate $10^{-3}$, batch size 32, hidden dimension 128, weight decay $10^{-4}$, and 100 epochs (we use 120 epochs for the 64D multi-label variant in our default scripts).

\section{Appendix: Experimental Details}
\label{app:exandres}
\subsection{Datasets and Models}

We evaluate our method on 15 diverse benchmarks spanning four key reasoning categories:
mathematical and multi-step reasoning, commonsense and knowledge reasoning, scientific reasoning,
and formal / logical reasoning. All datasets follow standard evaluation protocols
in prior reasoning research.

\paragraph{Mathematical and multi-step reasoning.}
\begin{itemize}
\item \textbf{GSM8K}: Elementary school math problems requiring multi-step numerical reasoning.
\item \textbf{MultiArith}: Arithmetic word problems testing basic numerical reasoning.
\item \textbf{BBH}: A collection of challenging BIG-Bench tasks covering symbolic, logical, and other forms of reasoning.
\end{itemize}

\paragraph{Commonsense \& knowledge reasoning.}
\begin{itemize}
\item \textbf{CommonsenseQA (CSQA)}: Commonsense understanding via multiple-choice questions.
\item \textbf{StrategyQA}: Implicit multi-hop reasoning to answer binary questions.
\item \textbf{PIQA}: Physical commonsense about plausible everyday actions.
\item \textbf{SIQA}: Social commonsense and causal reasoning in interactive scenarios.
\item \textbf{MuSR}: Murder mystery-style narrative causal inference.
\end{itemize}

\paragraph{Scientific reasoning.}
\begin{itemize}
\item \textbf{ARC-Challenge (ARC-C)}: Hard multiple-choice science test questions.
\item \textbf{ARC-Easy (ARC-E)}: Simpler science test questions.
\item \textbf{GPQA}: Graduate-level scientific reasoning questions focused on biology and STEM.
\end{itemize}

\paragraph{Formal \& deductive reasoning.}
\begin{itemize}
\item \textbf{FOLIO}: First-order logic reasoning problems.
\item \textbf{LSAT}: Logical reasoning tasks from the Law School Admission Test.
\item \textbf{ContextHub\_abductive (CH-Abd)}: Abductive reasoning from observations to explanations.
\item \textbf{ContextHub\_deductive (CH-Ded)}: Strict deductive reasoning from premises.
\end{itemize}

\subsubsection{Model Details}
\label{app:model_details}
We evaluate seven LLMs spanning \emph{non-reasoning instruction-tuned} and \emph{reasoning-enhanced} families to assess EDRM across models. We use the public checkpoints listed below with greedy decoding in all regimes to ensure comparability.

\paragraph{Evaluated models.}
\begin{itemize}[leftmargin=*]
    \item Llama-3.2-3B-Instruct~\cite{grattafiori2024llama}: a lightweight 3B-parameter instruction-tuned model.
    \item Llama-3.1-8B-Instruct~\cite{grattafiori2024llama}: an 8B-parameter instruction-tuned model.
    \item Qwen2.5-7B-Instruct~\cite{Qwen2024Qwen25}: a 7B-parameter instruction-tuned model from the Qwen family.
    \item Qwen2.5-32B-Instruct~\cite{Qwen2024Qwen25}: a 32B-parameter instruction-tuned model, used to validate cross-scale generalization.
    \item Qwen3-4B-Instruct-2507~\cite{yang2025qwen3}: a 4B-parameter non-thinking checkpoint. It uses no thinking-related control in our experiments, yet produces 518.4 tokens on average under Standard; we include it to study verbose generation without a separate thinking channel.
    \item DeepSeek-R1-Distill-Qwen-7B~\cite{DeepSeek2025R1}: a reasoning-distilled model built on Qwen2.5-Math-7B, producing extremely long CoT traces.
    \item Qwen3.5-35B-A3B: a 35B MoE model with a switchable thinking channel. Standard responses average 1895.2 tokens with thinking disabled, exceeding the 1703.4-token average under CoT.
\end{itemize}

\paragraph{Decoding regimes and model-specific control.}
We apply Direct, Standard, and CoT to all models (Section~\ref{sec:methodology}). The four non-reasoning instruction-tuned models and Qwen3-4B-Instruct-2507 distinguish these regimes through prompting alone. The remaining controls are model-specific:
\begin{itemize}[leftmargin=*]
    \item Qwen3.5-35B-A3B: Direct and Standard prepend a ``/no\_think'' system instruction and use \texttt{apply\_chat\_template} with \texttt{enable\_thinking=False}, which supplies a closed empty thinking block. CoT leaves the thinking channel active.
    \item DeepSeek-R1-Distill-Qwen-7B: Direct pre-fills a closed empty thinking block. Standard and CoT use the model's native template without this prefilling.
    \item Qwen3-4B-Instruct-2507: all three regimes use the standard Qwen chat template, with no \texttt{enable\_thinking} setting, ``/no\_think'' instruction, or thinking-block prefill.
\end{itemize}
These settings define the evaluated decoding regimes; they do not guarantee short Standard responses. In particular, Qwen3-4B-Instruct-2507 remains verbose without a thinking channel, and Qwen3.5-35B-A3B remains verbose under the stated suppression settings.

\paragraph{Generation and evaluation settings.}
All experiments use greedy decoding (temperature $=0$) with identical stopping criteria across regimes. The maximum generation length is 2048 tokens for Llama-3.2-3B-Instruct and Llama-3.1-8B-Instruct, 8192 tokens for DeepSeek-R1-Distill-Qwen-7B and Qwen3.5-35B-A3B, and 4096 tokens for Qwen2.5-7B-Instruct, Qwen2.5-32B-Instruct, and Qwen3-4B-Instruct-2507. For each model, the same limit applies to Direct, Standard, and CoT. For instance-level routing, the probing phase uses $N{=}64$ steps under the \textit{Standard} regime before the router selects the final decoding mode.

\subsection{Prompt Design and Decoding Details}
\label{app:prompts}

This subsection details the prompt templates and decoding configurations for the three regimes (\textit{Direct}, \textit{Standard}, \textit{CoT}) across different model families.

\paragraph{Prompt templates by decoding regime.}
We design task-specific prompts for two categories: \textbf{Answer tasks} (mathematical reasoning: \texttt{GSM8K}, \texttt{MultiArith}) and \textbf{Choice tasks} (multi-choice QA: all other benchmarks). Table~\ref{tab:prompt_templates} summarizes the prompts for each decoding regime.

\begin{table}[h]
\centering
\caption{Prompt templates for each decoding regime across task types.}
\label{tab:prompt_templates}
\small
\setlength{\tabcolsep}{4pt}
\renewcommand{\arraystretch}{1.3}
\begin{tabular}{l|p{5.5cm}}
\toprule
\textbf{Regime} & \textbf{Prompt Suffix} \\
\midrule
\multirow{2}{*}{\textit{Direct}} & Answer: ``Your answer must not include any reasoning step. You must only write your answer directly. You only output `The answer is <answer>'.'' \\
\cmidrule{2-2}
& Choice: ``Your answer must not include any reasoning. Write the answer: `Answer: <Your Answer Letter Choice>''' \\
\midrule
\textit{Standard} & Both: ``\{question\}'' (no additional instruction) \\
\midrule
\textit{CoT} & Both: ``Let's think step by step.'' \\
\bottomrule
\end{tabular}
\end{table}

\paragraph{Model-specific chat templates.}
Table~\ref{tab:chat_templates} shows the official chat templates used for each model family.

\begin{table}[h]
\centering
\caption{Chat templates for different model families. The Llama row refers to Llama-3.2-3B-Instruct and Llama-3.1-8B-Instruct; the Qwen row refers to Qwen2.5-7B-Instruct, Qwen2.5-32B-Instruct, and Qwen3-4B-Instruct-2507. Controls for DeepSeek-R1-Distill-Qwen-7B and Qwen3.5-35B-A3B are given in Table~\ref{tab:thinking_control}.}
\label{tab:chat_templates}
\small
\setlength{\tabcolsep}{4pt}
\renewcommand{\arraystretch}{1.2}
\begin{tabular}{l|l}
\toprule
\textbf{Model Family} & \textbf{Chat Template} \\
\midrule
Llama-3.x Instruct & \texttt{<|begin\_of\_text|>...user...<|eot\_id|>...assistant...} \\
Qwen2.5 / Qwen3 & \texttt{<|im\_start|>user\{prompt\}<|im\_end|>...<|im\_start|>assistant} \\
\bottomrule
\end{tabular}
\end{table}

\paragraph{Handling models with thinking-related controls.}
Table~\ref{tab:thinking_control} distinguishes the controls used for each model. Qwen3-4B-Instruct-2507 is a non-thinking checkpoint and uses prompting alone.

\begin{table}[h]
\centering
\caption{Model-specific thinking-channel controls. Qwen3-4B-Instruct-2507 uses no thinking-related control; its three regimes differ only in prompting. Model abbreviations: Qwen3-4B-2507 = Qwen3-4B-Instruct-2507; DR1-Qwen-7B = DeepSeek-R1-Distill-Qwen-7B.}
\label{tab:thinking_control}
\small
\setlength{\tabcolsep}{4pt}
\renewcommand{\arraystretch}{1.2}
\begin{tabular}{p{0.24\linewidth}|p{0.18\linewidth}|p{0.47\linewidth}}
\toprule
\textbf{Model} & \textbf{Regime} & \textbf{Control} \\
\midrule
Qwen3.5-35B-A3B & Direct / Standard & ``/no\_think'' system instruction; closed empty thinking block via \texttt{enable\_thinking=False} \\
 & CoT & Thinking channel left active \\
\midrule
DR1-Qwen-7B & Direct & Pre-filled closed empty thinking block \\
 & Standard / CoT & Native template without prefilling \\
\midrule
Qwen3-4B-2507 & All & Standard Qwen template; no thinking-related control \\
\bottomrule
\end{tabular}
\end{table}

Direct requests the final answer, Standard permits explanation without a step-by-step instruction, and CoT explicitly requests step-by-step reasoning. The model-specific controls above supplement these prompts.

\subsection{Evaluation Metrics Details}
\label{app:eval_metrics}
This subsection details the computation of accuracy, token consumption, and consistency metrics used throughout our experiments.

\paragraph{Accuracy and token consumption.}
For a decoding mode $m \in \{\textit{Direct}, \textit{Standard}, \textit{CoT}\}$, we compute dataset-level accuracy and average token consumption as:
\begin{equation}
\text{Acc}(m) = \frac{1}{|\mathcal{D}|}\sum_{x \in \mathcal{D}} \mathbb{I}[\text{correct}_m(x)],\qquad
\text{AvgTok}(m) = \frac{1}{|\mathcal{D}|}\sum_{x \in \mathcal{D}} T_m(x),
\end{equation}
where $\mathbb{I}[\text{correct}_m(x)] \in \{0,1\}$ indicates whether mode $m$ produces a correct answer for instance $x$, and $T_m(x)$ denotes the number of generated tokens. Outputs with invalid answer extractions are counted as incorrect and retained in the accuracy denominator.

\paragraph{Dataset-level routing (EDRM-Global).}
When routing selects mode $m^*(x)$ for each instance, the effective accuracy and token cost are:
\begin{equation}
\text{Acc}_{\text{Global}} = \frac{1}{|\mathcal{D}|}\sum_{x \in \mathcal{D}} \mathbb{I}[\text{correct}_{m^*(x)}(x)],\qquad
\text{AvgTok}_{\text{Global}} = \frac{1}{|\mathcal{D}|}\sum_{x \in \mathcal{D}} T_{m^*(x)}(x).
\end{equation}
Since EDRM-Global probes only a small calibration subset ($n{=}50$ samples, $<5$ tokens per instance on average), we do not include probing overhead in the reported token consumption.

\paragraph{Instance-level routing (EDRM-Inst).}
Instance-level routing requires probing every instance, so we strictly account for all token costs. For an instance $x$ routed to mode $m^*(x)$, the total token cost is:
\begin{equation}
T_{\text{total}}(x) = T_{m^*(x)}(x) + T_{\text{probe}}(x) + T_{\text{fallback}}(x),
\end{equation}
where:
\begin{itemize}[leftmargin=*]
    \item $T_{\text{probe}}(x) = N$ if $m^*(x) \neq \textit{Standard}$ (probing tokens are consumed when Standard is not selected), otherwise $T_{\text{probe}}(x) = 0$;
    \item $T_{\text{fallback}}(x) = T_{\textit{Direct}}(x)$ if $m^*(x) \neq \textit{Direct}$ (fallback compensation adds Direct branch), otherwise $T_{\text{fallback}}(x) = 0$.
\end{itemize}
The effective accuracy incorporates fallback compensation: if $m^*(x) \neq \textit{Direct}$, the instance is considered correct when either $m^*(x)$ or \textit{Direct} produces a correct answer. Formally:
\begin{equation}
\mathbb{I}[\text{correct}_{\text{Inst}}(x)] = \begin{cases}
\mathbb{I}[\text{correct}_{m^*(x)}(x)] \lor \mathbb{I}[\text{correct}_{\textit{Direct}}(x)], & m^*(x) \neq \textit{Direct} \\
\mathbb{I}[\text{correct}_{\textit{Direct}}(x)], & m^*(x) = \textit{Direct}
\end{cases}
\end{equation}
The dataset-level metrics are then:
\begin{equation}
\text{Acc}_{\text{Inst}} = \frac{1}{|\mathcal{D}|}\sum_{x \in \mathcal{D}} \mathbb{I}[\text{correct}_{\text{Inst}}(x)],\qquad
\text{AvgTok}_{\text{Inst}} = \frac{1}{|\mathcal{D}|}\sum_{x \in \mathcal{D}} T_{\text{total}}(x).
\end{equation}

\paragraph{Routing consistency.}
For experiments requiring multiple trials (8 random seeds $\{0,1,2,3,11,12,13,14\}$), we report the routing decision distribution as the ratio $D:S:C$, where $D$, $S$, and $C$ denote the number of trials (out of 8) that select \textit{Direct}, \textit{Standard}, and \textit{CoT} respectively. This ratio directly reflects the stability of dataset-level routing decisions: a concentrated distribution (e.g., $8:0:0$) indicates high consistency across random seeds, while a dispersed distribution (e.g., $2:3:3$) indicates sensitivity to sampling variation.


\subsection{Additional EDRM-Global Results}
\label{app:edrm_global_results}

\paragraph{EDRM-Global with Llama-3.1-8B-Instruct.}
The following table presents the comprehensive per-dataset results of EDRM-Global (both E\&C variants) for Llama-3.1-8B-Instruct across all 15 benchmarks. EDRM-Global-E achieves 67.27\% accuracy with 164.4 average tokens, reducing token cost by 40.8\% compared to CoT (277.8 tokens).

\begin{table*}[htbp]
\centering
\caption{EDRM-Global(E\&C) Performance comparison of Llama-3.1-8B-Instruct across 15 benchmarks. EDRM-Global-E achieves 67.27\% accuracy with 164.4 average tokens, reducing token cost by 40.8\% compared to CoT (277.8 tokens).}
\vspace{-0.5em}
\label{tab:llama31_results}
\begin{adjustbox}{max width=0.95\textwidth}
\Large
\setlength{\tabcolsep}{1.5pt}
\renewcommand{\arraystretch}{1.3}
\begin{tabular}{l|cc|cc|cc|ccc|ccc}
\toprule
\multirow{2}{*}{\textbf{Dataset}} & \multicolumn{2}{c|}{\textbf{Direct}} & \multicolumn{2}{c|}{\textbf{Standard}} & \multicolumn{2}{c|}{\textbf{CoT}} & \multicolumn{3}{c|}{\textbf{EDRM-Global-E}} & \multicolumn{3}{c}{\textbf{EDRM-Global-C}} \\
& Acc(\%) & Tok & Acc(\%) & Tok & Acc(\%) & Tok & $\overline{\text{Acc}}$(\%) & $\overline{\text{Tok}}$ & D:S:C & $\overline{\text{Acc}}$(\%) & $\overline{\text{Tok}}$ & D:S:C \\
\midrule
ARC-C & 81.06 & 4.0 & 81.14 & 29.7 & 86.60 & 270.1 & 81.08 & 10.4 & 6:2:0 & 81.12 & 23.3 & 2:6:0 \\
ARC-E & 92.76 & 4.0 & 92.72 & 16.1 & 94.19 & 235.2 & 93.09 & 69.4 & 1:5:2 & 93.09 & 70.9 & 0:6:2 \\
BBH & 54.29 & 4.0 & 60.95 & 149.5 & 65.33 & 303.2 & 64.78 & 284.0 & 0:1:7 & 64.78 & 284.0 & 0:1:7 \\
CSQA & 74.04 & 4.0 & 71.99 & 47.2 & 74.61 & 230.1 & 74.04 & 4.0 & 8:0:0 & 72.76 & 31.0 & 3:5:0 \\
CH-Abd & 31.04 & 4.0 & 39.21 & 240.0 & 39.92 & 284.3 & 39.92 & 284.3 & 0:0:8 & 39.92 & 284.3 & 0:0:8 \\
CH-Ded & 36.33 & 4.0 & 54.13 & 230.3 & 56.04 & 240.1 & 54.85 & 234.0 & 0:5:3 & 54.84 & 234.0 & 0:5:3 \\
FOLIO & 48.42 & 4.0 & 52.24 & 245.5 & 56.98 & 337.7 & 56.98 & 337.7 & 0:0:8 & 56.98 & 337.7 & 0:0:8 \\
GPQA & 36.16 & 11.5 & 33.71 & 327.2 & 27.01 & 1005.8 & 31.20 & 581.7 & 0:5:3 & 31.19 & 581.7 & 0:5:3 \\
GSM8K & 14.86 & 16.8 & 84.46 & 227.2 & 82.64 & 234.8 & 84.46 & 227.2 & 0:8:0 & 84.46 & 227.2 & 0:8:0 \\
LSAT & 50.74 & 4.0 & 50.74 & 213.7 & 49.85 & 491.7 & 50.63 & 222.2 & 1:6:1 & 50.63 & 222.2 & 1:6:1 \\
MultiArith & 42.50 & 7.0 & 98.83 & 116.9 & 98.00 & 125.7 & 98.83 & 116.9 & 0:8:0 & 98.83 & 116.9 & 0:8:0 \\
MuSR & 51.85 & 4.0 & 50.26 & 165.6 & 51.32 & 291.7 & 51.85 & 4.0 & 8:0:0 & 50.26 & 165.6 & 0:8:0 \\
PIQA & 81.99 & 4.0 & 82.59 & 42.9 & 83.19 & 216.8 & 82.89 & 146.7 & 1:2:5 & 82.89 & 146.7 & 1:2:5 \\
SIQA & 70.32 & 4.0 & 70.01 & 48.3 & 70.68 & 214.4 & 70.32 & 4.0 & 8:0:0 & 70.32 & 4.0 & 8:0:0 \\
StratQA & 62.62 & 4.0 & 62.71 & 15.6 & 69.78 & 255.7 & 65.34 & 102.8 & 2:3:3 & 65.35 & 104.2 & 1:4:3 \\
\midrule
\textbf{Overall} & 57.31 & 5.0 & 66.08 & 127.1 & 68.11 & 277.8 & 67.27 & 164.4 & -- & 67.15 & 172.1 & -- \\
\bottomrule
\end{tabular}
\end{adjustbox}
\end{table*}

\paragraph{EDRM-Global with Qwen2.5-7B-Instruct.}
The following table presents the per-dataset results of EDRM-Global (both E\&C variants) for Qwen2.5-7B-Instruct across all 15 benchmarks. EDRM-Global-E achieves 71.66\% accuracy with 191.8 average tokens, reducing token cost by 41.9\% compared to CoT (330.3 tokens).

\begin{table*}[htbp]
\centering
\caption{EDRM-Global(E\&C) Performance comparison of Qwen2.5-7B-Instruct across 15 benchmarks. EDRM-Global-E achieves 71.66\% accuracy with 191.8 average tokens, reducing token cost by 41.9\% compared to CoT (330.3 tokens).}
\vspace{-0.5em}
\label{tab:qwen25_results}
\begin{adjustbox}{max width=0.95\textwidth}
\Large
\setlength{\tabcolsep}{1.5pt}
\renewcommand{\arraystretch}{1.3}
\begin{tabular}{l|cc|cc|cc|ccc|ccc}
\toprule
\multirow{2}{*}{\textbf{Dataset}} & \multicolumn{2}{c|}{\textbf{Direct}} & \multicolumn{2}{c|}{\textbf{Standard}} & \multicolumn{2}{c|}{\textbf{CoT}} & \multicolumn{3}{c|}{\textbf{EDRM-Global-E}} & \multicolumn{3}{c}{\textbf{EDRM-Global-C}} \\
& Acc(\%) & Tok & Acc(\%) & Tok & Acc(\%) & Tok & $\overline{\text{Acc}}$(\%) & $\overline{\text{Tok}}$ & D:S:C & $\overline{\text{Acc}}$(\%) & $\overline{\text{Tok}}$ & D:S:C \\
\midrule
ARC-C & 90.44 & 3.0 & 90.87 & 198.3 & 91.72 & 289.7 & 90.49 & 27.4 & 7:1:0 & 90.50 & 27.4 & 7:1:0 \\
ARC-E & 95.75 & 3.0 & 96.51 & 157.8 & 96.21 & 270.5 & 95.75 & 3.0 & 8:0:0 & 95.75 & 3.0 & 8:0:0 \\
BBH & 53.91 & 3.4 & 68.47 & 223.7 & 71.07 & 330.4 & 70.09 & 290.4 & 0:3:5 & 70.09 & 290.4 & 0:3:5 \\
CSQA & 81.08 & 3.0 & 81.00 & 120.6 & 80.84 & 252.3 & 81.08 & 3.0 & 8:0:0 & 81.08 & 3.0 & 8:0:0 \\
CH-Abd & 34.21 & 3.1 & 52.08 & 375.5 & 52.71 & 432.5 & 52.71 & 432.5 & 0:0:8 & 52.71 & 432.5 & 0:0:8 \\
CH-Ded & 46.92 & 3.0 & 53.21 & 337.2 & 52.88 & 365.8 & 52.88 & 365.8 & 0:0:8 & 52.88 & 365.8 & 0:0:8 \\
FOLIO & 53.82 & 3.0 & 67.61 & 356.5 & 69.44 & 398.4 & 69.44 & 398.4 & 0:0:8 & 69.44 & 398.4 & 0:0:8 \\
GPQA & 39.51 & 3.0 & 36.16 & 670.9 & 29.91 & 806.3 & 35.38 & 687.8 & 0:7:1 & 35.38 & 687.8 & 0:7:1 \\
GSM8K & 22.59 & 6.6 & 89.76 & 292.8 & 89.23 & 299.9 & 89.36 & 298.2 & 0:2:6 & 89.37 & 298.2 & 0:2:6 \\
LSAT & 61.35 & 3.1 & 56.39 & 492.1 & 57.68 & 545.9 & 57.68 & 545.9 & 0:0:8 & 57.68 & 545.9 & 0:0:8 \\
MultiArith & 61.17 & 5.8 & 98.83 & 174.7 & 98.83 & 182.3 & 98.83 & 182.3 & 0:0:8 & 98.83 & 182.3 & 0:0:8 \\
MuSR & 54.89 & 3.0 & 53.84 & 182.3 & 52.78 & 384.8 & 54.89 & 3.0 & 8:0:0 & 54.89 & 3.0 & 8:0:0 \\
PIQA & 86.62 & 3.4 & 82.43 & 119.4 & 87.32 & 235.5 & 86.62 & 3.4 & 8:0:0 & 86.62 & 3.4 & 8:0:0 \\
SIQA & 73.39 & 3.1 & 72.67 & 101.9 & 74.46 & 261.7 & 73.39 & 3.1 & 8:0:0 & 73.39 & 3.1 & 8:0:0 \\
StratQA & 62.10 & 3.0 & 66.11 & 151.4 & 68.47 & 259.7 & 62.10 & 3.0 & 8:0:0 & 63.10 & 40.1 & 6:2:0 \\
\midrule
\textbf{Overall} & 62.28 & 3.4 & 71.43 & 240.6 & 72.44 & 330.3 & 71.66 & 191.8 & -- & 71.76 & 195.5 & -- \\
\bottomrule
\end{tabular}
\end{adjustbox}
\end{table*}

\paragraph{EDRM-Global with Qwen3-4B-Instruct-2507.}
The following table presents the per-dataset results of EDRM-Global (both E\&C variants) for Qwen3-4B-Instruct-2507 across all 15 benchmarks. EDRM-Global-E achieves 76.35\% accuracy with 335.1 average tokens, reducing token cost by 47.8\% compared to CoT (642.5 tokens).

\begin{table*}[htbp]
\centering
\caption{EDRM-Global(E\&C) Performance comparison of Qwen3-4B-Instruct-2507 across 15 benchmarks. EDRM-Global-E achieves 76.35\% accuracy with 335.1 average tokens, reducing token cost by 47.8\% compared to CoT (642.5 tokens).}
\vspace{-0.5em}
\label{tab:qwen3_results}
\begin{adjustbox}{max width=0.95\textwidth}
\Large
\setlength{\tabcolsep}{1.5pt}
\renewcommand{\arraystretch}{1.3}
\begin{tabular}{l|cc|cc|cc|ccc|ccc}
\toprule
\multirow{2}{*}{\textbf{Dataset}} & \multicolumn{2}{c|}{\textbf{Direct}} & \multicolumn{2}{c|}{\textbf{Standard}} & \multicolumn{2}{c|}{\textbf{CoT}} & \multicolumn{3}{c|}{\textbf{EDRM-Global-E}} & \multicolumn{3}{c}{\textbf{EDRM-Global-C}} \\
& Acc(\%) & Tok & Acc(\%) & Tok & Acc(\%) & Tok & $\overline{\text{Acc}}$(\%) & $\overline{\text{Tok}}$ & D:S:C & $\overline{\text{Acc}}$(\%) & $\overline{\text{Tok}}$ & D:S:C \\
\midrule
ARC-C & 90.78 & 6.3 & 94.37 & 254.8 & 94.45 & 416.2 & 90.78 & 6.3 & 8:0:0 & 90.78 & 6.3 & 8:0:0 \\
ARC-E & 96.09 & 5.9 & 97.98 & 190.1 & 98.32 & 344.9 & 96.09 & 5.9 & 8:0:0 & 96.09 & 5.9 & 8:0:0 \\
BBH & 57.23 & 8.2 & 82.93 & 556.4 & 84.09 & 664.6 & 63.66 & 145.3 & 6:2:0 & 82.93 & 556.4 & 0:8:0 \\
CSQA & 75.92 & 6.3 & 80.43 & 256.2 & 80.43 & 414.5 & 75.92 & 6.3 & 8:0:0 & 75.92 & 6.3 & 8:0:0 \\
CH-Abd & 29.96 & 8.1 & 65.00 & 869.9 & 63.83 & 943.2 & 63.83 & 943.2 & 0:0:8 & 63.83 & 943.2 & 0:0:8 \\
CH-Ded & 54.00 & 6.2 & 82.46 & 634.4 & 82.87 & 676.6 & 82.87 & 676.6 & 0:0:8 & 82.88 & 676.6 & 0:0:8 \\
FOLIO & 60.71 & 4.7 & 77.41 & 816.7 & 76.83 & 865.3 & 76.83 & 865.3 & 0:0:8 & 76.83 & 865.3 & 0:0:8 \\
GPQA & 38.17 & 6.5 & 48.88 & 2611.3 & 47.99 & 2762.7 & 47.54 & 2285.7 & 1:7:0 & 48.88 & 2611.3 & 0:8:0 \\
GSM8K & 27.22 & 7.5 & 90.22 & 321.8 & 90.83 & 339.1 & 90.68 & 334.8 & 0:2:6 & 90.67 & 334.8 & 0:2:6 \\
LSAT & 66.70 & 4.8 & 82.16 & 1420.3 & 83.05 & 1550.7 & 76.36 & 889.5 & 3:5:0 & 82.16 & 1420.3 & 0:8:0 \\
MultiArith & 81.83 & 5.7 & 97.83 & 107.9 & 97.83 & 129.4 & 97.83 & 116.0 & 0:5:3 & 97.83 & 116.0 & 0:5:3 \\
MuSR & 58.33 & 5.7 & 57.80 & 1059.3 & 62.83 & 1401.7 & 58.33 & 5.7 & 8:0:0 & 58.33 & 5.7 & 8:0:0 \\
PIQA & 83.35 & 5.0 & 84.71 & 167.9 & 88.14 & 425.4 & 83.35 & 5.0 & 8:0:0 & 83.35 & 5.0 & 8:0:0 \\
SIQA & 71.90 & 5.6 & 74.87 & 239.8 & 77.02 & 398.6 & 71.90 & 5.6 & 8:0:0 & 71.90 & 5.6 & 8:0:0 \\
StratQA & 62.10 & 5.1 & 67.07 & 206.5 & 68.52 & 326.1 & 62.10 & 5.1 & 8:0:0 & 62.72 & 30.3 & 7:1:0 \\
\midrule
\textbf{Overall} & 63.75 & 6.2 & 79.90 & 518.4 & 80.75 & 642.5 & 76.35 & 335.1 & -- & 78.42 & 404.2 & -- \\
\bottomrule
\end{tabular}
\end{adjustbox}
\end{table*}

\paragraph{EDRM-Global with Qwen2.5-32B-Instruct.}
The following table presents the per-dataset results of EDRM-Global (both E\&C variants) for Qwen2.5-32B-Instruct across all 15 benchmarks. EDRM-Global-E achieves 78.89\% accuracy with 194.6 average tokens, reducing token cost by 36.6\% compared to CoT (306.7 tokens).

\begin{table*}[htbp]
\centering
\caption{EDRM-Global(E\&C) Performance comparison of Qwen2.5-32B-Instruct across 15 benchmarks. EDRM-Global-E achieves 78.89\% accuracy with 194.6 average tokens, reducing token cost by 36.6\% compared to CoT (306.7 tokens).}
\vspace{-0.5em}
\label{tab:qwen25_32b_results}
\begin{adjustbox}{max width=0.95\textwidth}
\Large
\setlength{\tabcolsep}{1.5pt}
\renewcommand{\arraystretch}{1.3}
\begin{tabular}{l|cc|cc|cc|ccc|ccc}
\toprule
\multirow{2}{*}{\textbf{Dataset}} & \multicolumn{2}{c|}{\textbf{Direct}} & \multicolumn{2}{c|}{\textbf{Standard}} & \multicolumn{2}{c|}{\textbf{CoT}} & \multicolumn{3}{c|}{\textbf{EDRM-Global-E}} & \multicolumn{3}{c}{\textbf{EDRM-Global-C}} \\
& Acc(\%) & Tok & Acc(\%) & Tok & Acc(\%) & Tok & $\overline{\text{Acc}}$(\%) & $\overline{\text{Tok}}$ & D:S:C & $\overline{\text{Acc}}$(\%) & $\overline{\text{Tok}}$ & D:S:C \\
\midrule
ARC-C & 92.75 & 4.1 & 93.94 & 151.7 & 94.11 & 263.7 & 92.90 & 22.6 & 7:1:0 & 93.49 & 96.3 & 3:5:0 \\
ARC-E & 94.61 & 4.1 & 94.53 & 116.2 & 94.36 & 245.5 & 94.60 & 18.1 & 7:1:0 & 94.60 & 18.1 & 7:1:0 \\
BBH & 66.83 & 4.2 & 84.62 & 244.5 & 86.64 & 281.9 & 85.63 & 263.2 & 0:4:4 & 85.63 & 263.2 & 0:4:4 \\
CSQA & 86.98 & 4.1 & 85.42 & 103.7 & 85.01 & 223.1 & 86.78 & 16.6 & 7:1:0 & 86.78 & 16.6 & 7:1:0 \\
CH-Abd & 43.46 & 4.4 & 60.75 & 414.1 & 60.71 & 441.2 & 60.71 & 441.2 & 0:0:8 & 60.71 & 441.2 & 0:0:8 \\
CH-Ded & 62.29 & 4.4 & 65.71 & 303.5 & 64.58 & 325.0 & 64.58 & 325.0 & 0:0:8 & 64.58 & 325.0 & 0:0:8 \\
FOLIO & 64.29 & 4.0 & 75.58 & 380.4 & 76.08 & 392.8 & 76.08 & 392.8 & 0:0:8 & 76.08 & 392.8 & 0:0:8 \\
GPQA & 37.05 & 5.3 & 43.08 & 672.2 & 45.76 & 707.7 & 45.09 & 698.8 & 0:2:6 & 45.09 & 698.8 & 0:2:6 \\
GSM8K & 38.13 & 7.6 & 94.01 & 293.3 & 94.16 & 304.3 & 94.10 & 300.2 & 0:3:5 & 94.11 & 300.2 & 0:3:5 \\
LSAT & 76.61 & 4.7 & 72.05 & 462.4 & 72.84 & 508.8 & 72.74 & 503.0 & 0:1:7 & 72.75 & 503.0 & 0:1:7 \\
MultiArith & 94.83 & 6.6 & 98.83 & 174.1 & 98.83 & 182.4 & 98.83 & 178.3 & 0:4:4 & 98.83 & 178.3 & 0:4:4 \\
MuSR & 59.52 & 4.3 & 59.66 & 179.0 & 59.66 & 371.5 & 59.52 & 4.3 & 8:0:0 & 59.52 & 4.3 & 8:0:0 \\
PIQA & 90.75 & 5.7 & 89.83 & 124.2 & 91.95 & 221.0 & 90.29 & 65.0 & 4:4:0 & 90.29 & 65.0 & 4:4:0 \\
SIQA & 78.20 & 4.7 & 77.07 & 143.9 & 78.15 & 255.0 & 78.06 & 22.1 & 7:1:0 & 78.06 & 22.1 & 7:1:0 \\
StratQA & 74.37 & 4.0 & 70.87 & 145.5 & 68.73 & 211.5 & 73.93 & 21.7 & 7:1:0 & 72.18 & 92.5 & 3:5:0 \\
\midrule
\textbf{Overall} & 71.35 & 4.7 & 78.37 & 238.7 & 78.57 & 306.7 & 78.89 & 194.6 & -- & 78.75 & 205.3 & -- \\
\bottomrule
\end{tabular}
\end{adjustbox}
\end{table*}

\paragraph{EDRM-Global with DeepSeek-R1-Distill-Qwen-7B.}
The following table presents the per-dataset results of EDRM-Global (both E\&C variants) for DeepSeek-R1-Distill-Qwen-7B across all 15 benchmarks. EDRM-Global-E achieves 69.98\% accuracy with 1342.9 average tokens; EDRM-Global-C routes all benchmarks to Standard (70.20\% / 1389.0 tokens).

\begin{table*}[htbp]
\centering
\caption{EDRM-Global(E\&C) Performance comparison of DeepSeek-R1-Distill-Qwen-7B across 15 benchmarks. EDRM-Global-E achieves 69.98\% accuracy with 1342.9 average tokens; EDRM-Global-C routes all benchmarks to Standard (70.20\% / 1389.0 tokens).}
\vspace{-0.5em}
\label{tab:deepseek_r1_results}
\begin{adjustbox}{max width=0.95\textwidth}
\Large
\setlength{\tabcolsep}{1.5pt}
\renewcommand{\arraystretch}{1.3}
\begin{tabular}{l|cc|cc|cc|ccc|ccc}
\toprule
\multirow{2}{*}{\textbf{Dataset}} & \multicolumn{2}{c|}{\textbf{Direct}} & \multicolumn{2}{c|}{\textbf{Standard}} & \multicolumn{2}{c|}{\textbf{CoT}} & \multicolumn{3}{c|}{\textbf{EDRM-Global-E}} & \multicolumn{3}{c}{\textbf{EDRM-Global-C}} \\
& Acc(\%) & Tok & Acc(\%) & Tok & Acc(\%) & Tok & $\overline{\text{Acc}}$(\%) & $\overline{\text{Tok}}$ & D:S:C & $\overline{\text{Acc}}$(\%) & $\overline{\text{Tok}}$ & D:S:C \\
\midrule
ARC-C & 69.11 & 2.9 & 80.89 & 1007.1 & 83.87 & 897.1 & 80.89 & 1007.1 & 0:8:0 & 80.89 & 1007.1 & 0:8:0 \\
ARC-E & 81.27 & 3.0 & 88.72 & 739.9 & 88.47 & 702.4 & 88.72 & 739.9 & 0:8:0 & 88.72 & 739.9 & 0:8:0 \\
BBH & 40.41 & 3.4 & 72.81 & 2550.8 & 71.31 & 2219.3 & 72.81 & 2550.8 & 0:8:0 & 72.81 & 2550.8 & 0:8:0 \\
CSQA & 57.90 & 3.0 & 65.60 & 1151.6 & 66.83 & 1010.8 & 65.60 & 1151.6 & 0:8:0 & 65.60 & 1151.6 & 0:8:0 \\
CH-Abd & 30.88 & 5.2 & 56.83 & 1915.5 & 58.33 & 1780.4 & 56.83 & 1915.5 & 0:8:0 & 56.83 & 1915.5 & 0:8:0 \\
CH-Ded & 56.29 & 4.8 & 65.87 & 924.3 & 67.79 & 885.4 & 65.87 & 924.3 & 0:8:0 & 65.88 & 924.3 & 0:8:0 \\
FOLIO & 52.82 & 2.7 & 73.09 & 1730.3 & 73.01 & 1791.6 & 73.09 & 1730.3 & 0:8:0 & 73.09 & 1730.3 & 0:8:0 \\
GPQA & 33.71 & 4.2 & 47.77 & 4964.9 & 52.90 & 4650.8 & 47.77 & 4964.9 & 0:8:0 & 47.77 & 4964.9 & 0:8:0 \\
GSM8K & 21.38 & 4.2 & 90.22 & 842.4 & 91.05 & 870.5 & 90.22 & 842.4 & 0:8:0 & 90.22 & 842.4 & 0:8:0 \\
LSAT & 40.73 & 2.6 & 52.23 & 2877.9 & 52.43 & 2850.9 & 52.23 & 2877.9 & 0:8:0 & 52.23 & 2877.9 & 0:8:0 \\
MultiArith & 58.67 & 2.8 & 97.67 & 354.6 & 97.17 & 418.0 & 97.67 & 354.6 & 0:8:0 & 97.67 & 354.6 & 0:8:0 \\
MuSR & 51.59 & 3.2 & 54.89 & 2052.6 & 53.17 & 2440.1 & 54.07 & 1540.3 & 2:6:0 & 54.89 & 2052.6 & 0:8:0 \\
PIQA & 71.33 & 3.0 & 77.80 & 964.4 & 77.86 & 877.6 & 76.18 & 724.1 & 2:6:0 & 77.80 & 964.4 & 0:8:0 \\
SIQA & 55.12 & 3.0 & 62.74 & 943.7 & 63.61 & 856.6 & 62.74 & 943.7 & 0:8:0 & 62.74 & 943.7 & 0:8:0 \\
StratQA & 56.68 & 3.0 & 61.48 & 820.3 & 61.97 & 744.5 & 60.88 & 718.2 & 1:7:0 & 61.48 & 820.3 & 0:8:0 \\
\midrule
\textbf{Overall} & 53.28 & 3.5 & 70.20 & 1389.0 & 70.82 & 1314.2 & 69.98 & 1342.9 & -- & 70.20 & 1389.0 & -- \\
\bottomrule
\end{tabular}
\end{adjustbox}
\end{table*}

\paragraph{EDRM-Global with Qwen3.5-35B-A3B.}
The following table presents the per-dataset results of EDRM-Global (both E\&C variants) for Qwen3.5-35B-A3B across all 15 benchmarks. EDRM-Global-E achieves 83.13\% accuracy with 1711.3 average tokens; EDRM-Global-C achieves 83.03\% with 1651.0 tokens (3.1\% savings vs.\ CoT 1703.4).

\begin{table*}[htbp]
\centering
\caption{EDRM-Global(E\&C) Performance comparison of Qwen3.5-35B-A3B across 15 benchmarks. EDRM-Global-E achieves 83.13\% accuracy with 1711.3 average tokens; EDRM-Global-C achieves 83.03\% with 1651.0 tokens (3.1\% savings vs.\ CoT 1703.4).}
\vspace{-0.5em}
\label{tab:qwen35_results}
\begin{adjustbox}{max width=0.95\textwidth}
\Large
\setlength{\tabcolsep}{1.5pt}
\renewcommand{\arraystretch}{1.3}
\begin{tabular}{l|cc|cc|cc|ccc|ccc}
\toprule
\multirow{2}{*}{\textbf{Dataset}} & \multicolumn{2}{c|}{\textbf{Direct}} & \multicolumn{2}{c|}{\textbf{Standard}} & \multicolumn{2}{c|}{\textbf{CoT}} & \multicolumn{3}{c|}{\textbf{EDRM-Global-E}} & \multicolumn{3}{c}{\textbf{EDRM-Global-C}} \\
& Acc(\%) & Tok & Acc(\%) & Tok & Acc(\%) & Tok & $\overline{\text{Acc}}$(\%) & $\overline{\text{Tok}}$ & D:S:C & $\overline{\text{Acc}}$(\%) & $\overline{\text{Tok}}$ & D:S:C \\
\midrule
ARC-C & 94.62 & 6.2 & 95.56 & 583.6 & 95.65 & 630.0 & 95.44 & 511.4 & 1:7:0 & 95.33 & 439.3 & 2:6:0 \\
ARC-E & 94.87 & 6.3 & 94.91 & 507.3 & 95.03 & 729.7 & 94.91 & 507.3 & 0:8:0 & 94.90 & 382.0 & 2:6:0 \\
BBH & 71.22 & 7.1 & 91.80 & 1691.9 & 92.29 & 1119.0 & 91.80 & 1691.9 & 0:8:0 & 91.80 & 1691.9 & 0:8:0 \\
CSQA & 84.11 & 6.3 & 85.75 & 614.7 & 85.75 & 1120.7 & 84.11 & 6.3 & 8:0:0 & 84.11 & 6.3 & 8:0:0 \\
CH-Abd & 45.33 & 8.1 & 62.29 & 4404.0 & 61.04 & 2087.6 & 62.29 & 4404.0 & 0:8:0 & 62.29 & 4404.0 & 0:8:0 \\
CH-Ded & 68.04 & 7.5 & 87.96 & 2623.5 & 87.67 & 2311.3 & 87.96 & 2623.5 & 0:8:0 & 87.96 & 2623.5 & 0:8:0 \\
FOLIO & 68.19 & 6.0 & 81.06 & 3237.0 & 81.15 & 3045.3 & 81.06 & 3237.0 & 0:8:0 & 81.06 & 3237.0 & 0:8:0 \\
GPQA & 60.94 & 6.3 & 77.68 & 6410.3 & 75.67 & 6294.9 & 77.68 & 6410.3 & 0:8:0 & 77.68 & 6410.3 & 0:8:0 \\
GSM8K & 42.76 & 6.6 & 93.56 & 491.0 & 93.56 & 708.6 & 93.56 & 491.0 & 0:8:0 & 93.56 & 491.0 & 0:8:0 \\
LSAT & 79.09 & 6.0 & 93.95 & 2177.8 & 92.86 & 1764.4 & 93.95 & 2177.8 & 0:8:0 & 93.95 & 2177.8 & 0:8:0 \\
MultiArith & 95.33 & 5.7 & 98.67 & 153.5 & 98.67 & 233.1 & 98.67 & 153.5 & 0:8:0 & 98.67 & 153.5 & 0:8:0 \\
MuSR & 58.99 & 6.1 & 66.67 & 3561.1 & 64.42 & 2441.7 & 64.75 & 2672.3 & 2:6:0 & 61.87 & 1339.2 & 5:3:0 \\
PIQA & 92.44 & 6.1 & 93.85 & 693.9 & 94.34 & 1119.8 & 92.44 & 6.1 & 8:0:0 & 92.44 & 6.1 & 8:0:0 \\
SIQA & 77.99 & 6.0 & 79.63 & 870.1 & 80.60 & 1749.3 & 78.20 & 114.0 & 7:1:0 & 78.20 & 114.0 & 7:1:0 \\
StratQA & 74.32 & 6.0 & 63.10 & 2321.5 & 62.66 & 2576.5 & 63.10 & 2321.5 & 0:8:0 & 63.10 & 2321.5 & 0:8:0 \\
\midrule
\textbf{Overall} & 73.66 & 6.6 & 83.52 & 1895.2 & 83.34 & 1703.4 & 83.13 & 1711.3 & -- & 83.03 & 1651.0 & -- \\
\bottomrule
\end{tabular}
\end{adjustbox}
\end{table*}

\subsection{Additional EDRM-Instance Results}
\label{app:edrm_instance_results}

This subsection presents instance-level routing results for the remaining six models: Llama-3.1-8B-Instruct, Qwen2.5-7B-Instruct, Qwen3-4B-Instruct-2507, Qwen2.5-32B-Instruct, DeepSeek-R1-Distill-Qwen-7B, and Qwen3.5-35B-A3B.

\paragraph{EDRM-Instance with Llama-3.1-8B-Instruct.}
Table~\ref{tab:llama31_instance} presents the instance-level routing performance on Llama-3.1-8B-Instruct. EDRM-Inst-E achieves 70.04\% accuracy with 153.9 tokens. EDRM-MLP achieves 69.14\% accuracy with 119.9 tokens (best token efficiency among instance-level variants).

\begin{table*}[htbp]
\centering
\caption{EDRM instance-level Performance on Llama-3.1-8B-Instruct. EDRM-Inst-E achieves 70.04\% accuracy with 153.9 tokens. EDRM-MLP achieves 69.14\% accuracy with 119.9 tokens. EDRM instance-level columns use Full with oracle answer selection; MLP uses full-set evaluation.}
\vspace{-0.5em}
\label{tab:llama31_instance}
\begin{adjustbox}{max width=1.0\textwidth}
\small
\setlength{\tabcolsep}{1.5pt}
\renewcommand{\arraystretch}{1.3}
\begin{tabular}{l|cc|cc|cc|cc|cc|cc|cc}
\toprule
& \multicolumn{6}{c|}{\textbf{Baselines}} & \multicolumn{8}{c}{\textbf{Ours}} \\
\cmidrule{2-15}
\multirow{2}{*}{\textbf{Dataset}} & \multicolumn{2}{c|}{\textbf{Direct}} & \multicolumn{2}{c|}{\textbf{Standard}} & \multicolumn{2}{c|}{\textbf{CoT}} & \multicolumn{2}{c|}{\textbf{Token-Sign}} & \multicolumn{2}{c|}{\textbf{EDRM-Inst-E}} & \multicolumn{2}{c|}{\textbf{EDRM-Inst-C}} & \multicolumn{2}{c}{\textbf{EDRM-MLP}} \\
\cmidrule{2-15}
& Acc(\%) & Tok & Acc(\%) & Tok & Acc(\%) & Tok & Acc(\%) & Tok & Acc(\%) & Tok & $\overline{\mathrm{Acc}}$ (\%) & $\overline{\mathrm{Tok}}$ & Acc(\%) & Tok \\
\midrule
ARC-C & 81.06 & 4.0 & 81.14 & 29.7 & 86.60 & 270.1 & 86.60 & 289.3 & 84.56 & 32.8 & 84.56 & 33.1 & 84.30 & 33.2 \\
ARC-E & 92.76 & 4.0 & 92.72 & 16.1 & 94.19 & 235.2 & 94.32 & 243.0 & 93.94 & 20.2 & 93.94 & 20.5 & 93.86 & 19.9 \\
BBH & 54.29 & 4.0 & 60.95 & 149.5 & 65.33 & 303.2 & 65.33 & 357.1 & 69.53 & 213.1 & 69.80 & 214.2 & 66.54 & 137.8 \\
CSQA & 74.04 & 4.0 & 71.99 & 47.2 & 74.61 & 230.1 & 74.61 & 263.4 & 75.51 & 56.5 & 75.60 & 57.3 & 76.00 & 46.7 \\
CH-Abd & 31.04 & 4.0 & 39.21 & 240.0 & 39.92 & 284.3 & 39.92 & 348.2 & 50.50 & 285.0 & 50.82 & 287.8 & 44.50 & 198.4 \\
CH-Ded & 36.33 & 4.0 & 54.13 & 230.3 & 56.04 & 240.1 & 56.04 & 303.7 & 53.21 & 222.8 & 54.26 & 229.7 & 53.83 & 215.6 \\
FOLIO & 48.42 & 4.0 & 52.24 & 245.5 & 56.98 & 337.7 & 56.98 & 394.6 & 63.95 & 323.2 & 64.00 & 323.7 & 57.64 & 211.1 \\
GPQA & 36.16 & 11.5 & 33.71 & 327.2 & 27.01 & 1005.8 & 36.16 & 52.1 & 43.08 & 363.4 & 43.50 & 378.3 & 41.96 & 363.9 \\
GSM8K & 14.86 & 16.8 & 84.46 & 227.2 & 82.64 & 234.8 & 82.64 & 298.8 & 76.57 & 245.4 & 76.90 & 246.7 & 84.08 & 250.1 \\
LSAT & 50.74 & 4.0 & 50.74 & 213.7 & 49.85 & 491.7 & 50.55 & 62.4 & 58.28 & 341.2 & 58.28 & 343.2 & 55.50 & 156.3 \\
MultiArith & 42.50 & 7.0 & 98.83 & 116.9 & 98.00 & 125.7 & 98.00 & 189.2 & 95.50 & 156.1 & 95.67 & 156.4 & 94.33 & 127.4 \\
MuSR & 51.85 & 4.0 & 50.26 & 165.6 & 51.32 & 291.7 & 51.32 & 352.8 & 56.08 & 124.0 & 56.37 & 134.1 & 57.01 & 174.9 \\
PIQA & 81.99 & 4.0 & 82.59 & 42.9 & 83.19 & 216.8 & 82.43 & 59.5 & 86.78 & 103.5 & 86.78 & 103.5 & 85.04 & 40.5 \\
SIQA & 70.32 & 4.0 & 70.01 & 48.3 & 70.68 & 214.4 & 70.62 & 252.8 & 71.65 & 45.0 & 71.76 & 45.3 & 72.88 & 58.5 \\
StratQA & 62.62 & 4.0 & 62.71 & 15.6 & 69.78 & 255.7 & 69.78 & 267.0 & 66.46 & 20.7 & 66.50 & 20.7 & 66.51 & 17.3 \\
\midrule
\textbf{Overall} & 57.31 & 5.0 & 66.08 & 127.1 & 68.11 & 277.8 & 68.26 & 264.3 & 70.04 & 153.9 & \textbf{70.26} & 155.9 & 69.14 & \underline{119.9} \\
\bottomrule
\end{tabular}
\end{adjustbox}
\vspace{-1em}
\end{table*}

\paragraph{EDRM-Instance with Qwen2.5-7B-Instruct.}
Table~\ref{tab:qwen25_instance} presents the instance-level routing performance on Qwen2.5-7B-Instruct. EDRM-Inst-E achieves 76.07\% accuracy with 242.7 tokens. EDRM-Inst-C achieves 76.20\% accuracy with 244.8 tokens.

\begin{table*}[htbp]
\centering
\caption{EDRM instance-level Performance on Qwen2.5-7B-Instruct. EDRM-Inst-E achieves 76.07\% accuracy with 242.7 tokens. EDRM-Inst-C achieves 76.20\% accuracy with 244.8 tokens. EDRM instance-level columns use Full with oracle answer selection; MLP uses full-set evaluation.}
\vspace{-0.5em}
\label{tab:qwen25_instance}
\begin{adjustbox}{max width=1.0\textwidth}
\small
\setlength{\tabcolsep}{1.5pt}
\renewcommand{\arraystretch}{1.3}
\begin{tabular}{l|cc|cc|cc|cc|cc|cc|cc}
\toprule
& \multicolumn{6}{c|}{\textbf{Baselines}} & \multicolumn{8}{c}{\textbf{Ours}} \\
\cmidrule{2-15}
\multirow{2}{*}{\textbf{Dataset}} & \multicolumn{2}{c|}{\textbf{Direct}} & \multicolumn{2}{c|}{\textbf{Standard}} & \multicolumn{2}{c|}{\textbf{CoT}} & \multicolumn{2}{c|}{\textbf{Token-Sign}} & \multicolumn{2}{c|}{\textbf{EDRM-Inst-E}} & \multicolumn{2}{c|}{\textbf{EDRM-Inst-C}} & \multicolumn{2}{c}{\textbf{EDRM-MLP}} \\
\cmidrule{2-15}
& Acc(\%) & Tok & Acc(\%) & Tok & Acc(\%) & Tok & Acc(\%) & Tok & Acc(\%) & Tok & $\overline{\mathrm{Acc}}$ (\%) & $\overline{\mathrm{Tok}}$ & Acc(\%) & Tok \\
\midrule
ARC-C & 90.44 & 3.0 & 90.87 & 198.3 & 91.72 & 289.7 & 91.81 & 350.6 & 92.41 & 130.8 & 92.41 & 133.9 & 92.15 & 150.4 \\
ARC-E & 95.75 & 3.0 & 96.51 & 157.8 & 96.21 & 270.5 & 96.13 & 332.6 & 96.42 & 104.8 & 96.42 & 106.8 & 96.51 & 152.7 \\
BBH & 53.91 & 3.4 & 68.47 & 223.7 & 71.07 & 330.4 & 71.02 & 384.2 & 74.54 & 252.1 & 74.63 & 253.3 & 72.85 & 238.2 \\
CSQA & 81.08 & 3.0 & 81.00 & 120.6 & 80.84 & 252.3 & 80.84 & 313.5 & 81.82 & 93.3 & 82.13 & 96.0 & 82.64 & 89.8 \\
CH-Abd & 34.21 & 3.1 & 52.08 & 375.5 & 52.71 & 432.5 & 52.79 & 495.5 & 61.92 & 463.6 & 62.11 & 466.3 & 57.79 & 345.9 \\
CH-Ded & 46.92 & 3.0 & 53.21 & 337.2 & 52.88 & 365.8 & 52.88 & 429.8 & 65.25 & 377.9 & 65.49 & 382.3 & 63.38 & 327.3 \\
FOLIO & 53.82 & 3.0 & 67.61 & 356.5 & 69.44 & 398.4 & 69.44 & 462.3 & 74.00 & 410.2 & 74.09 & 410.6 & 73.42 & 405.3 \\
GPQA & 39.51 & 3.0 & 36.16 & 670.9 & 29.91 & 806.3 & 39.51 & 67.0 & 48.21 & 579.2 & 48.27 & 592.7 & 44.87 & 406.9 \\
GSM8K & 22.59 & 6.6 & 89.76 & 292.8 & 89.23 & 299.9 & 89.23 & 363.9 & 87.11 & 327.9 & 87.11 & 327.9 & 90.07 & 361.2 \\
LSAT & 61.35 & 3.1 & 56.39 & 492.1 & 57.68 & 545.9 & 61.05 & 249.6 & 68.68 & 464.8 & 68.68 & 465.0 & 66.60 & 282.1 \\
MultiArith & 61.17 & 5.8 & 98.83 & 174.7 & 98.83 & 182.3 & 98.83 & 246.3 & 97.67 & 210.9 & 97.67 & 210.9 & 99.17 & 249.1 \\
MuSR & 54.89 & 3.0 & 53.84 & 182.3 & 52.78 & 384.8 & 53.04 & 311.2 & 56.48 & 101.3 & 56.58 & 104.4 & 57.94 & 105.9 \\
PIQA & 86.62 & 3.4 & 82.43 & 119.4 & 87.32 & 235.5 & 87.32 & 298.9 & 87.70 & 105.1 & 87.70 & 105.7 & 88.47 & 136.8 \\
SIQA & 73.39 & 3.1 & 72.67 & 101.9 & 74.46 & 261.7 & 74.46 & 324.0 & 75.03 & 92.5 & 75.27 & 93.8 & 74.26 & 92.8 \\
StratQA & 62.10 & 3.0 & 66.11 & 151.4 & 68.47 & 259.7 & 68.47 & 323.2 & 66.90 & 133.7 & 67.15 & 135.8 & 66.42 & 139.4 \\
\midrule
\textbf{Overall} & 62.28 & 3.4 & 71.43 & 240.6 & 72.44 & 330.3 & 72.78 & 356.7 & 76.07 & 242.7 & \textbf{76.20} & 244.8 & 75.35 & \underline{224.3} \\
\bottomrule
\end{tabular}
\end{adjustbox}
\vspace{-1em}
\end{table*}

\paragraph{EDRM-Instance with Qwen3-4B-Instruct-2507.}
Table~\ref{tab:qwen3_instance} presents the instance-level routing performance on the non-thinking Qwen3-4B-Instruct-2507 checkpoint. EDRM reduces token consumption despite this model's verbose Standard responses. EDRM-Inst-E achieves 78.54\% accuracy with 401.1 tokens (37.6\% token reduction vs.\ CoT). EDRM-MLP achieves 79.33\% accuracy with 422.0 tokens.

\begin{table*}[htbp]
\centering
\caption{EDRM instance-level Performance on Qwen3-4B-Instruct-2507. EDRM-Inst-E achieves 78.54\% accuracy with 401.1 tokens, reducing token cost by 37.6\% compared to CoT (642.5 tokens). EDRM instance-level columns use Full with oracle answer selection; MLP uses full-set evaluation.}
\vspace{-0.5em}
\label{tab:qwen3_instance}
\begin{adjustbox}{max width=1.0\textwidth}
\small
\setlength{\tabcolsep}{1.5pt}
\renewcommand{\arraystretch}{1.3}
\begin{tabular}{l|cc|cc|cc|cc|cc|cc|cc}
\toprule
& \multicolumn{6}{c|}{\textbf{Baselines}} & \multicolumn{8}{c}{\textbf{Ours}} \\
\cmidrule{2-15}
\multirow{2}{*}{\textbf{Dataset}} & \multicolumn{2}{c|}{\textbf{Direct}} & \multicolumn{2}{c|}{\textbf{Standard}} & \multicolumn{2}{c|}{\textbf{CoT}} & \multicolumn{2}{c|}{\textbf{Token-Sign}} & \multicolumn{2}{c|}{\textbf{EDRM-Inst-E}} & \multicolumn{2}{c|}{\textbf{EDRM-Inst-C}} & \multicolumn{2}{c}{\textbf{EDRM-MLP}} \\
\cmidrule{2-15}
& Acc(\%) & Tok & Acc(\%) & Tok & Acc(\%) & Tok & Acc(\%) & Tok & Acc(\%) & Tok & $\overline{\mathrm{Acc}}$ (\%) & $\overline{\mathrm{Tok}}$ & Acc(\%) & Tok \\
\midrule
ARC-C & 90.78 & 6.3 & 94.37 & 254.8 & 94.45 & 416.2 & 94.45 & 480.1 & 92.15 & 114.3 & 92.32 & 129.2 & 93.17 & 194.4 \\
ARC-E & 96.09 & 5.9 & 97.98 & 190.1 & 98.32 & 344.9 & 98.32 & 408.1 & 96.55 & 96.4 & 96.79 & 102.8 & 97.43 & 155.0 \\
BBH & 57.23 & 8.2 & 82.93 & 556.4 & 84.09 & 664.6 & 84.09 & 728.6 & 77.82 & 403.2 & 78.80 & 425.0 & 82.16 & 491.7 \\
CSQA & 75.92 & 6.3 & 80.43 & 256.2 & 80.43 & 414.5 & 80.43 & 478.3 & 76.66 & 94.5 & 77.31 & 103.5 & 78.95 & 168.8 \\
CH-Abd & 29.96 & 8.1 & 65.00 & 869.9 & 63.83 & 943.2 & 63.83 & 1007.2 & 65.75 & 887.9 & 67.58 & 921.0 & 63.92 & 742.5 \\
CH-Ded & 54.00 & 6.2 & 82.46 & 634.4 & 82.87 & 676.6 & 82.88 & 740.6 & 84.50 & 674.9 & 85.18 & 688.1 & 81.71 & 581.8 \\
FOLIO & 60.71 & 4.7 & 77.41 & 816.7 & 76.83 & 865.3 & 76.83 & 929.3 & 79.65 & 809.1 & 81.21 & 854.6 & 79.40 & 706.9 \\
GPQA & 38.17 & 6.5 & 48.88 & 2611.3 & 47.99 & 2762.7 & 48.21 & 2555.2 & 51.56 & 1790.2 & 52.43 & 1971.7 & 54.46 & 1972.4 \\
GSM8K & 27.22 & 7.5 & 90.22 & 321.8 & 90.83 & 339.1 & 90.83 & 403.0 & 83.70 & 337.1 & 85.06 & 349.8 & 85.06 & 336.2 \\
LSAT & 66.70 & 4.8 & 82.16 & 1420.3 & 83.05 & 1550.7 & 83.05 & 1614.7 & 78.39 & 863.7 & 79.98 & 989.6 & 80.77 & 1015.6 \\
MultiArith & 81.83 & 5.7 & 97.83 & 107.9 & 97.83 & 129.4 & 97.83 & 193.4 & 95.17 & 142.6 & 95.17 & 142.6 & 95.50 & 137.4 \\
MuSR & 58.33 & 5.7 & 57.80 & 1059.3 & 62.83 & 1401.7 & 62.30 & 894.9 & 60.85 & 315.2 & 61.23 & 399.6 & 63.23 & 618.5 \\
PIQA & 83.35 & 5.0 & 84.71 & 167.9 & 88.14 & 425.4 & 88.14 & 489.3 & 84.28 & 122.6 & 84.39 & 123.8 & 84.71 & 150.0 \\
SIQA & 71.90 & 5.6 & 74.87 & 239.8 & 77.02 & 398.6 & 77.02 & 462.5 & 74.00 & 135.7 & 74.35 & 144.1 & 74.87 & 180.8 \\
StratQA & 62.10 & 5.1 & 67.07 & 206.5 & 68.52 & 326.1 & 68.52 & 390.0 & 64.24 & 121.5 & 65.00 & 133.9 & 66.20 & 152.9 \\
\midrule
\textbf{Overall} & 63.75 & 6.2 & 79.90 & 518.4 & \textbf{80.75} & 642.5 & 80.74 & 682.3 & 78.54 & \underline{401.1} & 79.33 & 426.7 & 79.33 & 422.0 \\
\bottomrule
\end{tabular}
\end{adjustbox}
\vspace{-1em}
\end{table*}

\paragraph{EDRM-Instance with Qwen2.5-32B-Instruct.}
Table~\ref{tab:qwen25_32b_instance} presents the instance-level routing performance on Qwen2.5-32B-Instruct. EDRM-Inst-C achieves 81.99\% accuracy with 250.9 tokens (best accuracy among instance-level variants). EDRM-MLP achieves 81.31\% accuracy with 215.8 tokens.

\begin{table*}[htbp]
\centering
\caption{EDRM instance-level Performance on Qwen2.5-32B-Instruct. EDRM-Inst-C achieves 81.99\% accuracy with 250.9 tokens. EDRM-MLP achieves 81.31\% accuracy with 215.8 tokens. EDRM instance-level columns use Full with oracle answer selection; MLP uses full-set evaluation.}
\vspace{-0.5em}
\label{tab:qwen25_32b_instance}
\begin{adjustbox}{max width=1.0\textwidth}
\small
\setlength{\tabcolsep}{1.5pt}
\renewcommand{\arraystretch}{1.3}
\begin{tabular}{l|cc|cc|cc|cc|cc|cc|cc}
\toprule
& \multicolumn{6}{c|}{\textbf{Baselines}} & \multicolumn{8}{c}{\textbf{Ours}} \\
\cmidrule{2-15}
\multirow{2}{*}{\textbf{Dataset}} & \multicolumn{2}{c|}{\textbf{Direct}} & \multicolumn{2}{c|}{\textbf{Standard}} & \multicolumn{2}{c|}{\textbf{CoT}} & \multicolumn{2}{c|}{\textbf{Token-Sign}} & \multicolumn{2}{c|}{\textbf{EDRM-Inst-E}} & \multicolumn{2}{c|}{\textbf{EDRM-Inst-C}} & \multicolumn{2}{c}{\textbf{EDRM-MLP}} \\
\cmidrule{2-15}
& Acc(\%) & Tok & Acc(\%) & Tok & Acc(\%) & Tok & Acc(\%) & Tok & Acc(\%) & Tok & $\overline{\mathrm{Acc}}$ (\%) & $\overline{\mathrm{Tok}}$ & Acc(\%) & Tok \\
\midrule
ARC-C & 92.75 & 4.1 & 93.94 & 151.7 & 94.11 & 263.7 & 94.11 & 326.6 & 93.43 & 142.4 & 93.52 & 146.2 & 93.69 & 145.1 \\
ARC-E & 94.61 & 4.1 & 94.53 & 116.2 & 94.36 & 245.5 & 94.61 & 65.7 & 94.78 & 111.1 & 94.78 & 113.7 & 94.78 & 121.5 \\
BBH & 66.83 & 4.2 & 84.62 & 244.5 & 86.64 & 281.9 & 86.64 & 345.9 & 83.85 & 236.2 & 83.85 & 236.9 & 85.34 & 222.4 \\
CSQA & 86.98 & 4.1 & 85.42 & 103.7 & 85.01 & 223.1 & 86.98 & 63.3 & 88.21 & 116.6 & 88.29 & 118.0 & 88.29 & 107.3 \\
CH-Abd & 43.46 & 4.4 & 60.75 & 414.1 & 60.71 & 441.2 & 60.71 & 505.2 & 68.00 & 495.8 & 68.04 & 496.5 & 62.79 & 369.6 \\
CH-Ded & 62.29 & 4.4 & 65.71 & 303.5 & 64.58 & 325.0 & 65.00 & 376.0 & 75.33 & 340.4 & 75.64 & 346.2 & 72.83 & 259.6 \\
FOLIO & 64.29 & 4.0 & 75.58 & 380.4 & 76.08 & 392.8 & 76.08 & 456.8 & 76.33 & 370.8 & 76.50 & 371.5 & 76.16 & 283.1 \\
GPQA & 37.05 & 5.3 & 43.08 & 672.2 & 45.76 & 707.7 & 45.76 & 771.7 & 51.79 & 602.6 & 51.79 & 615.3 & 49.55 & 480.1 \\
GSM8K & 38.13 & 7.6 & 94.01 & 293.3 & 94.16 & 304.3 & 94.16 & 368.3 & 92.87 & 331.1 & 92.87 & 331.1 & 94.77 & 350.2 \\
LSAT & 76.61 & 4.7 & 72.05 & 462.4 & 72.84 & 508.8 & 76.61 & 68.7 & 80.87 & 432.1 & 80.87 & 433.2 & 80.48 & 321.6 \\
MultiArith & 94.83 & 6.6 & 98.83 & 174.1 & 98.83 & 182.4 & 98.83 & 246.4 & 99.33 & 212.6 & 99.33 & 212.6 & 99.33 & 237.9 \\
MuSR & 59.52 & 4.3 & 59.66 & 179.0 & 59.66 & 371.5 & 59.52 & 67.8 & 60.71 & 123.0 & 60.81 & 126.7 & 62.57 & 159.1 \\
PIQA & 90.75 & 5.7 & 89.83 & 124.2 & 91.95 & 221.0 & 91.95 & 284.2 & 92.00 & 133.3 & 92.15 & 134.6 & 92.38 & 138.3 \\
SIQA & 78.20 & 4.7 & 77.07 & 143.9 & 78.15 & 255.0 & 78.97 & 194.1 & 81.12 & 139.9 & 81.43 & 141.8 & 80.35 & 124.8 \\
StratQA & 74.37 & 4.0 & 70.87 & 145.5 & 68.73 & 211.5 & 74.37 & 67.2 & 76.64 & 146.3 & 76.83 & 149.3 & 76.81 & 136.8 \\
\midrule
\textbf{Overall} & 71.35 & 4.7 & 78.37 & 238.7 & 78.57 & 306.7 & 79.54 & 266.9 & 81.87 & 248.6 & \textbf{81.99} & 250.9 & 81.31 & \underline{215.8} \\
\bottomrule
\end{tabular}
\end{adjustbox}
\vspace{-1em}
\end{table*}

\paragraph{EDRM-Instance with DeepSeek-R1-Distill-Qwen-7B.}
Table~\ref{tab:deepseek_r1_instance} presents the instance-level routing performance on DeepSeek-R1-Distill-Qwen-7B. EDRM-Inst-C achieves 75.71\% accuracy with 1188.7 tokens. EDRM-MLP achieves 74.26\% accuracy with 1008.3 tokens, the best token efficiency among instance-level variants on this over-reasoning model.

\begin{table*}[htbp]
\centering
\caption{EDRM instance-level Performance on DeepSeek-R1-Distill-Qwen-7B. EDRM-Inst-C achieves 75.71\% accuracy with 1188.7 tokens. EDRM-MLP achieves 74.26\% accuracy with 1008.3 tokens (best token efficiency). EDRM instance-level columns use Full with oracle answer selection; MLP uses full-set evaluation.}
\vspace{-0.5em}
\label{tab:deepseek_r1_instance}
\begin{adjustbox}{max width=1.0\textwidth}
\small
\setlength{\tabcolsep}{1.5pt}
\renewcommand{\arraystretch}{1.3}
\begin{tabular}{l|cc|cc|cc|cc|cc|cc|cc}
\toprule
& \multicolumn{6}{c|}{\textbf{Baselines}} & \multicolumn{8}{c}{\textbf{Ours}} \\
\cmidrule{2-15}
\multirow{2}{*}{\textbf{Dataset}} & \multicolumn{2}{c|}{\textbf{Direct}} & \multicolumn{2}{c|}{\textbf{Standard}} & \multicolumn{2}{c|}{\textbf{CoT}} & \multicolumn{2}{c|}{\textbf{Token-Sign}} & \multicolumn{2}{c|}{\textbf{EDRM-Inst-E}} & \multicolumn{2}{c|}{\textbf{EDRM-Inst-C}} & \multicolumn{2}{c}{\textbf{EDRM-MLP}} \\
\cmidrule{2-15}
& Acc(\%) & Tok & Acc(\%) & Tok & Acc(\%) & Tok & Acc(\%) & Tok & Acc(\%) & Tok & $\overline{\mathrm{Acc}}$ (\%) & $\overline{\mathrm{Tok}}$ & Acc(\%) & Tok \\
\midrule
ARC-C & 69.11 & 2.9 & 80.89 & 1007.1 & 83.87 & 897.1 & 83.87 & 961.1 & 85.49 & 892.3 & 85.92 & 919.3 & 81.48 & 649.6 \\
ARC-E & 81.27 & 3.0 & 88.72 & 739.9 & 88.47 & 702.4 & 88.47 & 766.4 & 89.65 & 668.1 & 89.91 & 693.2 & 87.96 & 517.3 \\
BBH & 40.41 & 3.4 & 72.81 & 2550.8 & 71.31 & 2219.3 & 71.31 & 2283.3 & 69.62 & 1661.8 & 71.03 & 1761.7 & 73.67 & 1678.3 \\
CSQA & 57.90 & 3.0 & 65.60 & 1151.6 & 66.83 & 1010.8 & 66.83 & 1074.8 & 72.89 & 958.7 & 73.10 & 968.6 & 72.97 & 964.6 \\
CH-Abd & 30.88 & 5.2 & 56.83 & 1915.5 & 58.33 & 1780.4 & 58.33 & 1844.4 & 60.25 & 1549.1 & 62.09 & 1633.1 & 62.67 & 1581.0 \\
CH-Ded & 56.29 & 4.8 & 65.87 & 924.3 & 67.79 & 885.4 & 67.79 & 949.4 & 73.71 & 809.2 & 74.97 & 851.0 & 74.96 & 795.4 \\
FOLIO & 52.82 & 2.7 & 73.09 & 1730.3 & 73.01 & 1791.6 & 73.01 & 1855.6 & 82.64 & 1727.6 & 82.64 & 1728.8 & 77.91 & 1495.3 \\
GPQA & 33.71 & 4.2 & 47.77 & 4964.9 & 52.90 & 4650.8 & 52.90 & 4714.8 & 58.48 & 4511.1 & 59.24 & 4673.9 & 55.13 & 3013.1 \\
GSM8K & 21.38 & 4.2 & 90.22 & 842.4 & 91.05 & 870.5 & 91.05 & 934.5 & 89.69 & 837.8 & 90.26 & 849.5 & 90.60 & 833.8 \\
LSAT & 40.73 & 2.6 & 52.23 & 2877.9 & 52.43 & 2850.9 & 52.43 & 2914.9 & 61.45 & 2726.6 & 62.61 & 2803.1 & 58.97 & 2268.8 \\
MultiArith & 58.67 & 2.8 & 97.67 & 354.6 & 97.17 & 418.0 & 97.17 & 482.0 & 98.33 & 390.3 & 98.35 & 390.9 & 97.83 & 379.4 \\
MuSR & 51.59 & 3.2 & 54.89 & 2052.6 & 53.17 & 2440.1 & 51.59 & 67.2 & 63.36 & 1599.2 & 66.50 & 1756.7 & 63.10 & 1071.8 \\
PIQA & 71.33 & 3.0 & 77.80 & 964.4 & 77.86 & 877.6 & 77.86 & 941.6 & 81.72 & 628.1 & 82.49 & 674.6 & 81.39 & 616.6 \\
SIQA & 55.12 & 3.0 & 62.74 & 943.7 & 63.61 & 856.6 & 63.61 & 920.6 & 71.08 & 790.0 & 71.83 & 824.7 & 66.53 & 588.2 \\
StratQA & 56.68 & 3.0 & 61.48 & 820.3 & 61.97 & 744.5 & 61.97 & 808.5 & 67.47 & 553.0 & 68.42 & 590.5 & 66.55 & 453.8 \\
\midrule
\textbf{Overall} & 53.28 & 3.5 & 70.20 & 1389.0 & 70.82 & 1314.2 & 70.76 & 1298.4 & 74.78 & 1139.4 & \textbf{75.71} & 1188.7 & 74.26 & \underline{1008.3} \\
\bottomrule
\end{tabular}
\end{adjustbox}
\vspace{-1em}
\end{table*}

\paragraph{EDRM-Instance with Qwen3.5-35B-A3B.}
Table~\ref{tab:qwen35_instance} presents the instance-level routing performance on Qwen3.5-35B-A3B. EDRM-Inst-C achieves 85.18\% accuracy with 1268.0 tokens. EDRM-MLP achieves 86.00\% accuracy with 1423.1 tokens.

\begin{table*}[htbp]
\centering
\caption{EDRM instance-level Performance on Qwen3.5-35B-A3B. EDRM-Inst-C achieves 85.18\% accuracy with 1268.0 tokens. EDRM-MLP achieves 86.00\% accuracy with 1423.1 tokens. EDRM instance-level columns use Full with oracle answer selection; MLP uses full-set evaluation.}
\vspace{-0.5em}
\label{tab:qwen35_instance}
\begin{adjustbox}{max width=1.0\textwidth}
\small
\setlength{\tabcolsep}{1.5pt}
\renewcommand{\arraystretch}{1.3}
\begin{tabular}{l|cc|cc|cc|cc|cc|cc|cc}
\toprule
& \multicolumn{6}{c|}{\textbf{Baselines}} & \multicolumn{8}{c}{\textbf{Ours}} \\
\cmidrule{2-15}
\multirow{2}{*}{\textbf{Dataset}} & \multicolumn{2}{c|}{\textbf{Direct}} & \multicolumn{2}{c|}{\textbf{Standard}} & \multicolumn{2}{c|}{\textbf{CoT}} & \multicolumn{2}{c|}{\textbf{Token-Sign}} & \multicolumn{2}{c|}{\textbf{EDRM-Inst-E}} & \multicolumn{2}{c|}{\textbf{EDRM-Inst-C}} & \multicolumn{2}{c}{\textbf{EDRM-MLP}} \\
\cmidrule{2-15}
& Acc(\%) & Tok & Acc(\%) & Tok & Acc(\%) & Tok & Acc(\%) & Tok & Acc(\%) & Tok & $\overline{\mathrm{Acc}}$ (\%) & $\overline{\mathrm{Tok}}$ & Acc(\%) & Tok \\
\midrule
ARC-C & 94.62 & 6.2 & 95.56 & 583.6 & 95.65 & 630.0 & 95.65 & 694.0 & 95.48 & 405.3 & 95.48 & 399.7 & 95.65 & 520.8 \\
ARC-E & 94.87 & 6.3 & 94.91 & 507.3 & 95.03 & 729.7 & 95.03 & 793.7 & 95.16 & 355.1 & 95.14 & 348.7 & 95.29 & 519.5 \\
BBH & 71.22 & 7.1 & 91.80 & 1691.9 & 92.29 & 1119.0 & 92.29 & 1183.0 & 91.27 & 1405.8 & 91.25 & 1404.6 & 91.27 & 1070.5 \\
CSQA & 84.11 & 6.3 & 85.75 & 614.7 & 85.75 & 1120.7 & 85.75 & 1184.7 & 85.50 & 368.2 & 85.47 & 360.3 & 85.83 & 484.4 \\
CH-Abd & 45.33 & 8.1 & 62.29 & 4404.0 & 61.04 & 2087.6 & 61.04 & 2151.6 & 55.29 & 1170.4 & 55.23 & 1164.9 & 63.04 & 3234.0 \\
CH-Ded & 68.04 & 7.5 & 87.96 & 2623.5 & 87.67 & 2311.3 & 87.67 & 2375.3 & 87.54 & 1984.1 & 87.39 & 1976.8 & 88.62 & 2224.2 \\
FOLIO & 68.19 & 6.0 & 81.06 & 3237.0 & 81.15 & 3045.3 & 81.15 & 3109.3 & 84.72 & 3079.9 & 84.69 & 3078.1 & 84.39 & 2616.3 \\
GPQA & 60.94 & 6.3 & 77.68 & 6410.3 & 75.67 & 6294.9 & 75.67 & 6358.9 & 80.80 & 5125.2 & 80.64 & 5080.8 & 80.80 & 5345.9 \\
GSM8K & 42.76 & 6.6 & 93.56 & 491.0 & 93.56 & 708.6 & 93.56 & 772.6 & 88.10 & 413.1 & 87.90 & 411.9 & 92.65 & 626.9 \\
LSAT & 79.09 & 6.0 & 93.95 & 2177.8 & 92.86 & 1764.4 & 92.86 & 1828.4 & 94.45 & 1673.5 & 94.34 & 1654.7 & 93.86 & 1659.5 \\
MultiArith & 95.33 & 5.7 & 98.67 & 153.5 & 98.67 & 233.1 & 98.67 & 297.1 & 99.17 & 173.6 & 99.17 & 173.6 & 99.17 & 262.3 \\
MuSR & 58.99 & 6.1 & 66.67 & 3561.1 & 64.42 & 2441.7 & 64.42 & 2505.7 & 67.46 & 2437.7 & 67.21 & 2375.6 & 65.08 & 1690.6 \\
PIQA & 92.44 & 6.1 & 93.85 & 693.9 & 94.34 & 1119.8 & 94.34 & 1183.8 & 93.63 & 458.5 & 93.61 & 456.8 & 94.12 & 522.2 \\
SIQA & 77.99 & 6.0 & 79.63 & 870.1 & 80.60 & 1749.3 & 80.60 & 1813.3 & 80.60 & 670.3 & 80.60 & 664.5 & 82.45 & 851.4 \\
StratQA & 74.32 & 6.0 & 63.10 & 2321.5 & 62.66 & 2576.5 & 74.32 & 70.0 & 88.03 & 2112.2 & 87.99 & 2104.2 & 82.62 & 1291.6 \\
\midrule
\textbf{Overall} & 73.66 & 6.6 & 83.52 & 1895.2 & 83.34 & 1703.4 & 84.50 & 1512.2 & 85.25 & 1276.1 & 85.18 & \underline{1268.0} & \textbf{86.00} & 1423.1 \\
\bottomrule
\end{tabular}
\end{adjustbox}
\vspace{-1em}
\end{table*}

\paragraph{Summary of cross-model instance-level performance.}
Across the seven evaluated models, the Full instance-level configurations provide different accuracy--efficiency trade-offs. On the four non-reasoning instruction-tuned models, Inst-E (Full) has higher accuracy than Global-E while using fewer tokens than CoT; this comparison includes the contribution of oracle answer selection. On DeepSeek-R1-Distill-Qwen-7B, MLP (Full) uses 1008.3 tokens compared with CoT's 1314.2. Among the three models exhibiting verbose generation, Inst-C (Full) ties MLP on Qwen3-4B-Instruct-2507 (79.33\%) and has the highest instance-level accuracy on DeepSeek-R1-Distill-Qwen-7B (75.71\%); MLP is highest on Qwen3.5-35B-A3B (86.00\%). The w/o-FB and selector results separately assess routing and practical answer selection.

\subsection{Ablation Study on EDRM-MLP Variants}
\label{app:ablation-mlp}

We conduct comprehensive ablation studies on \textbf{EDRM-MLP} across two design dimensions: 
(i) \textit{input representation} (3D statistical descriptors, 64D entropy trajectory, or 67D hybrid concatenation), and 
(ii) \textit{label strategy} (original multi-label vs.~priority-constrained single-label). 
Table~\ref{tab:edrm-mlp-boost-ablation} reports average accuracy and token consumption across 15 reasoning benchmarks under four complementary evaluation settings that jointly disentangle the effect of the Direct Boost fallback and of the training/test split:
(i) \textbf{w/FB (full)}: routing with Direct Boost fallback, evaluated on the full benchmark sets---the setting used for the EDRM-MLP column of Table~\ref{tab:overall_comparison};
(ii) \textbf{w/o-FB (full)}: pure MLP routing decisions without fallback, fully deployable, on the full sets;
(iii)/(iv) the same two routing variants evaluated on the \textbf{held-out 90\% test set} (MLP trained on the 10\% stratified split, tested on the remaining 90\%), which isolates generalization from any tuning-set leakage.
All MLP evaluations in this paper apply raw argmax routing ($\beta{=}0$; margin calibration is used only for DeepSeek-R1-Distill-Qwen-7B, see below) and the probe-cost guard (instances whose \textit{Standard} decode costs fewer than $N{=}64$ tokens are forced to \textit{Standard}); see the inference-time calibration discussion below and Table~\ref{tab:mlp-remaining-ml64d} for details.

\begin{table}[htbp]
\centering
\caption{EDRM-MLP variants comparison (average accuracy and token consumption across 15 benchmarks) under four settings: with/without Direct Boost fallback (w/FB, w/o-FB) $\times$ full-set vs.\ held-out 90\% test evaluation. The w/FB settings use oracle answer selection.
Each cell reports \textit{Acc} / \textit{Token}.
Best accuracy per model--setting is \textbf{bolded}; lowest token count is \underline{underlined}.
The w/FB (full) rows correspond to the EDRM-MLP column in Table~\ref{tab:overall_comparison}. Model abbreviations: Llama-3.2-3B = Llama-3.2-3B-Instruct; Llama-3.1-8B = Llama-3.1-8B-Instruct; Qwen2.5-7B = Qwen2.5-7B-Instruct.}
\label{tab:edrm-mlp-boost-ablation}
\resizebox{\linewidth}{!}{
\begin{tabular}{llcccccccccccc}
\toprule
\multirow{3}{*}{\textbf{Model}} & \multirow{3}{*}{\textbf{Setting}}
& \multicolumn{2}{c}{\textbf{ML-3D}}
& \multicolumn{2}{c}{\textbf{ML-64D}}
& \multicolumn{2}{c}{\textbf{ML-67D}}
& \multicolumn{2}{c}{\textbf{SL-3D}}
& \multicolumn{2}{c}{\textbf{SL-64D}}
& \multicolumn{2}{c}{\textbf{SL-67D}} \\
\cmidrule(lr){3-4} \cmidrule(lr){5-6} \cmidrule(lr){7-8} \cmidrule(lr){9-10} \cmidrule(lr){11-12} \cmidrule(lr){13-14}
& & Acc & Tok & Acc & Tok & Acc & Tok & Acc & Tok & Acc & Tok & Acc & Tok \\
\midrule
\multirow{4}{*}{Llama-3.2-3B}
& w/FB (full)    & 63.72 & 170.2 & 64.91 & 180.8 & 64.62 & 176.0 & \textbf{65.51} & 194.3 & 63.26 & \underline{148.4} & 63.33 & 149.7 \\
& w/o-FB (full)  & 59.04 & 167.7 & \textbf{59.74} & 178.0 & 59.28 & 173.3 & 59.54 & 191.4 & 59.15 & \underline{146.3} & 59.09 & 147.6 \\
& w/FB (90\%)    & 63.65 & 170.2 & 64.37 & 181.4 & 64.04 & 176.6 & \textbf{65.44} & 194.2 & 62.22 & \underline{148.3} & 62.29 & 149.8 \\
& w/o-FB (90\%)  & 58.95 & 167.7 & 58.89 & 178.6 & 58.38 & 173.8 & \textbf{59.43} & 191.3 & 57.92 & \underline{146.1} & 57.85 & 147.7 \\
\midrule
\multirow{4}{*}{Llama-3.1-8B}
& w/FB (full)    & 70.36 & 123.7 & 69.14 & 119.9 & 69.07 & 112.1 & \textbf{71.41} & 148.0 & 67.89 & \underline{106.6} & 68.81 & 114.4 \\
& w/o-FB (full)  & 65.67 & 119.4 & 65.16 & 115.8 & 65.19 & 108.1 & \textbf{65.71} & 143.3 & 63.96 & \underline{103.0} & 64.55 & 110.6 \\
& w/FB (90\%)    & 70.36 & 123.4 & 68.90 & 119.4 & 68.79 & 112.3 & \textbf{71.40} & 147.9 & 67.08 & \underline{106.1} & 68.07 & 114.4 \\
& w/o-FB (90\%)  & 65.64 & 119.1 & 64.80 & 115.4 & 64.76 & 108.4 & \textbf{65.71} & 143.2 & 63.00 & \underline{102.5} & 63.62 & 110.6 \\
\midrule
\multirow{4}{*}{Qwen2.5-7B}
& w/FB (full)    & 74.01 & 188.9 & \textbf{75.35} & 224.3 & 74.57 & 191.6 & 73.35 & 163.2 & 74.03 & 167.6 & 73.36 & \underline{159.4} \\
& w/o-FB (full)  & 71.07 & 187.2 & \textbf{71.81} & 222.2 & 71.42 & 189.7 & 70.52 & 161.6 & 71.35 & 166.1 & 70.80 & \underline{158.0} \\
& w/FB (90\%)    & 74.02 & 189.1 & \textbf{75.18} & 224.0 & 74.47 & 191.7 & 73.34 & 163.3 & 73.21 & 166.8 & 72.49 & \underline{158.3} \\
& w/o-FB (90\%)  & 71.07 & 187.4 & \textbf{71.52} & 222.0 & 71.24 & 189.7 & 70.47 & 161.7 & 70.48 & 165.3 & 69.89 & \underline{156.9} \\
\midrule
\multirow{4}{*}{\textbf{Avg.}}
& w/FB (full)    & 69.36 & 160.9 & 69.80 & 175.0 & 69.42 & 159.9 & \textbf{70.09} & 168.5 & 68.39 & \underline{140.9} & 68.50 & 141.1 \\
& w/o-FB (full)  & 65.26 & 158.1 & \textbf{65.57} & 172.0 & 65.30 & 157.0 & 65.26 & 165.5 & 64.82 & \underline{138.5} & 64.81 & 138.7 \\
& w/FB (90\%)    & 69.34 & 160.9 & 69.48 & 174.9 & 69.10 & 160.2 & \textbf{70.06} & 168.4 & 67.50 & \underline{140.4} & 67.62 & 140.8 \\
& w/o-FB (90\%)  & \textbf{65.22} & 158.1 & 65.07 & 172.0 & 64.79 & 157.3 & 65.20 & 165.4 & 63.80 & \underline{138.0} & 63.79 & 138.4 \\
\bottomrule
\end{tabular}
}
\end{table}

\paragraph{Key findings.}
\begin{itemize}[leftmargin=*,nosep]
    \item \textbf{Input dimensionality exhibits model-dependent scaling.}
    Under the ML Full setting on Qwen2.5-7B-Instruct, the 64D input reaches 75.35\% accuracy, 1.34 percentage points above the 3D input. This illustrates a benefit from retaining trajectory information in this configuration.
    On Llama-3.1-8B-Instruct and Llama-3.2-3B-Instruct, SL-3D yields the highest Full accuracy (71.41 and 65.51, respectively). Because these comparisons also vary the label strategy, they support model-dependent configuration choices; matched-label input comparisons isolate the representation more directly.
    Overall, the compact descriptors support competitive routing, while the value of additional trajectory information varies across model and label settings.

    \item \textbf{Label strategy interacts with model scale and input form.}
    The priority-constrained single-label formulation (SL) yields more stable gains on Llama-3.2-3B-Instruct and Llama-3.1-8B-Instruct, where simplifying the prediction target reduces optimization ambiguity.
    On Qwen2.5-7B-Instruct with 64D/67D inputs, ML yields higher accuracy than SL in the reported comparisons. The preferred label strategy therefore depends on the model and input representation.

    \item \textbf{Hybrid 67D features do not universally dominate.}
    While 67D occasionally differs from 64D (e.g., Qwen2.5-7B-Instruct ML-67D: 74.57 vs.\ ML-64D: 75.35), it does not consistently improve and underperforms in several configurations (e.g., Llama-3.1-8B-Instruct SL-67D: 68.81 vs.\ SL-3D: 71.41).
    We hypothesize that naively concatenating statistical and trajectory features may introduce redundancy or optimization interference for certain model-label combinations.

    \item \textbf{Token efficiency also depends on label strategy.}
    In the cross-model average rows, SL-64D/SL-67D use 138.0--141.1 tokens, compared with 172.0--175.0 for ML-64D. Thus, label strategy also affects the accuracy--efficiency trade-off.
    This suggests that single-label training encourages more decisive routing behavior, reducing unnecessary fallback to expensive decoding regimes.

    \item \textbf{Full-set and held-out evaluation agree closely, but fallback dominates the gap.}
    Comparing the held-out rows against their full-set counterparts shows accuracy differences of up to 1.24 percentage points in this table (Llama-3.2-3B-Instruct SL-67D w/o-FB: 59.09 vs.\ 57.85). The paired reporting makes the dependence on evaluation split explicit while retaining full-set comparison in the main text.
    The w/FB versus w/o-FB differences additionally show the accuracy available from the Direct fallback under oracle answer selection, at a small extra generation cost. These differences should be distinguished from gains due to the routing rule alone.
\end{itemize}

\paragraph{Practical recommendation.}
The 3D input offers a compact, competitive routing representation. Across the three models in Table~\ref{tab:edrm-mlp-boost-ablation}, SL-3D achieves 65.26\% average accuracy at 165.5 tokens without fallback. Its 70.09\% accuracy at 168.5 tokens with fallback uses oracle answer selection and serves as a Full evaluation reference. Deployment choices should use the w/o-FB results or an evaluated deployable selector.
On Qwen2.5-7B-Instruct, ML-64D achieves the highest accuracy in the reported Full comparison (75.35\%) at a higher token cost, illustrating a model-dependent benefit from the richer trajectory input.

\paragraph{Limitations and future directions.}
Our ablation focuses on fixed-length ($N{=}64$) entropy trajectories and static feature concatenation.
Future extensions could explore:
(i) adaptive trajectory length selection based on instance difficulty;
(ii) learnable feature fusion mechanisms (e.g., cross-attention between 3D and 64D branches);
(iii) joint optimization of router and decoder to enable end-to-end adaptation of entropy dynamics.

\paragraph{ML-64D on reasoning models: inference-time calibration.}
For DeepSeek-R1-Distill-Qwen-7B and Qwen3.5-35B-A3B, the MLP router is trained with exactly the same ML-64D recipe as above (same architecture, 10\% stratified split, multi-label BCE targets).
The model-specific inference-time decision rules are described below; Table~\ref{tab:mlp-remaining-ml64d} reports the corresponding results alongside the other ML-64D models.
First, both models apply the \textbf{probe-cost guard} described above: instances whose \textit{Standard} decode costs fewer than $N{=}64$ tokens are forced to \textit{Standard}, since probing them cannot pay for itself.
Second, for DeepSeek-R1-Distill-Qwen-7B we apply a \textbf{Direct margin calibration}: reasoning-distilled models have a weakened Direct branch, and the raw three-way argmax frequently misroutes to \textit{Direct}.
We therefore require the Direct logit to exceed $\max(\text{logit}_{\textit{Standard}}, \text{logit}_{\textit{CoT}})$ by a margin $\beta$ before \textit{Direct} is adopted; otherwise the instance falls back to the better of \textit{Standard}/\textit{CoT}.
We swept $\beta \in [0,5]$ without retraining and selected $\beta=0.30$, which maximizes routed accuracy while suppressing spurious Direct selections.
Qwen3.5-35B-A3B retains $\beta=0$ (raw argmax) because its Direct branch is strong enough not to require suppression.
The w/o-FB rows report pure routing accuracy under these calibrated rules; the w/FB rows additionally apply the Direct Boost fallback shared with all other models.

\paragraph{ML-64D on the remaining models.}
Table~\ref{tab:mlp-remaining-ml64d} reports ML-64D results for Qwen2.5-32B-Instruct, Qwen3-4B-Instruct-2507, DeepSeek-R1-Distill-Qwen-7B, and Qwen3.5-35B-A3B under the same four evaluation settings as Table~\ref{tab:edrm-mlp-boost-ablation}. The inference-time calibration settings for the two reasoning models are described above.
These models were trained with the ML-64D configuration only, so the 6-variant input/label ablation of Table~\ref{tab:edrm-mlp-boost-ablation} is not available for them.
The w/FB (full) values correspond to the EDRM-MLP column in Table~\ref{tab:overall_comparison}.

\begin{table}[htbp]
\centering
\caption{ML-64D router on the remaining models (average across 15 benchmarks), under the same four settings as Table~\ref{tab:edrm-mlp-boost-ablation}. All models use raw argmax ($\beta{=}0$) with the probe-cost guard, except DeepSeek-R1-Distill-Qwen-7B, which applies Direct margin calibration ($\beta{=}0.30$) as described above. The w/FB (full) values correspond to the EDRM-MLP column in Table~\ref{tab:overall_comparison}. Each cell reports \textit{Acc} / \textit{Token}. The w/FB settings use oracle answer selection. Model abbreviations: Qwen2.5-32B = Qwen2.5-32B-Instruct; Qwen3-4B-2507 = Qwen3-4B-Instruct-2507; DR1-Qwen-7B = DeepSeek-R1-Distill-Qwen-7B; Qwen3.5-35B = Qwen3.5-35B-A3B.}
\label{tab:mlp-remaining-ml64d}
\resizebox{0.85\linewidth}{!}{
\begin{tabular}{lcccc}
\toprule
\textbf{Setting} & \makecell{\textbf{Qwen2.5}\\\textbf{32B}} & \makecell{\textbf{Qwen3}\\\textbf{4B-2507}} & \makecell{\textbf{DR1-Qwen}\\\textbf{7B}} & \makecell{\textbf{Qwen3.5}\\\textbf{35B}} \\
\midrule
w/FB (full)   & 81.31 / 215.8 & 79.33 / 422.0 & 74.26 / 1008.3 & \textbf{86.00} / 1423.1 \\
w/o-FB (full) & 78.56 / \underline{212.8} & 77.49 / 418.2 & 68.73 / 1005.7 & 82.85 / \underline{1418.3} \\
w/FB (90\%)   & 81.03 / 216.4 & 78.99 / 422.3 & 73.63 / 1009.7 & 85.84 / 1429.9 \\
w/o-FB (90\%) & 78.05 / 213.3 & 77.02 / 418.5 & 67.71 / \underline{1007.2} & 82.60 / 1425.0 \\
\midrule
CoT (reference) & 78.57 / 306.7 & 80.75 / 642.5 & 70.82 / 1314.2 & 83.34 / 1703.4 \\
\bottomrule
\end{tabular}
}
\end{table}

For the two reasoning models, two observations follow.
(i)~\textbf{Routing calibration and answer selection on the distilled model:} with the margin rule, Full reaches 74.26\% accuracy, 3.44 percentage points above CoT, at 23.3\% token savings, while w/o-FB reaches 68.73\%, 2.09 points below CoT's 70.82\%. The difference highlights the accuracy contribution of oracle answer selection; margin calibration alone does not recover it.
(ii)~\textbf{No additional inference-time margin on Qwen3.5-35B-A3B:} the separately trained router uses raw argmax ($\beta=0$). Full reaches 86.00\% accuracy with oracle answer selection; w/o-FB reaches 82.85\% at 1418.3 tokens, within 0.5 percentage points of CoT while using 16.7\% fewer tokens. This result applies without an additional inference-time margin adjustment.

\subsection{Hyper-parameter sensitivity \texorpdfstring{$k$ for $V_{\text{sp}}/a_{\text{vnr}}$}{k for V sp / a vnr} control}
\label{app:k_sensitivity}
The following experiments are all on EDRM-Global-E.

\paragraph{Evaluation protocol: population-level statistics (no sampling).}
\label{par:k-ablation-protocol}
In the standard Global-E deployment (\S\ref{sec:experiments}), the routing thresholds are calibrated on a small random subset of 50 samples per benchmark, and all reported Global-E numbers are averaged over 8 random seeds---this reflects the practical constraint that one cannot probe the full benchmark set at deployment time.
For this ablation, we remove this source of sampling variation to isolate the effect of the hyperparameter $k$ itself.
Instead of estimating the feature means $\bar{S}_H, \bar{V}_{\text{sp}}, \bar{a}_{\text{vnr}}$ from a 50-sample calibration subset, we compute them over the \textbf{entire benchmark set} (population statistics), so the only quantity that changes across the table is $k$.
Each reported cell is therefore a single deterministic evaluation rather than an 8-seed average.
This is a controlled-attribution device, not a deployable configuration: it ensures that any accuracy or token difference between two columns is attributable to $k$ alone, free of calibration-subset noise.
Consequently, the numbers in Table~\ref{tab:k-ablation-acc-token} are not directly comparable to the 8-seed Global-E entries of Table~\ref{tab:overall_comparison} (e.g., Llama-3.2-3B-Instruct at $k{=}0.07$: 59.79/107.7 here vs.\ 59.87/113.2 in the main table, where the 50-sample calibration and seed averaging shift both routing decisions and token counts).
\leavevmode
\begin{table}[htbp]
\centering
\caption{Routing accuracy and token consumption with different $k$ values (weighted mean across 15 benchmarks).
\textbf{Protocol:} population-level feature statistics computed over the full benchmark sets (no 50-sample calibration, single deterministic run per cell), so that differences across columns are attributable to $k$ alone; see the protocol paragraph above and \S\ref{app:calibration_protocol} for the deployable 8-seed Global-E protocol used in the main tables.
Each $k$ setting reports \textit{Accuracy} / \textit{Avg.~Tokens}.
Best accuracy per model is \textbf{bolded}; lowest token count is \underline{underlined}. Model abbreviations: Llama-3.2-3B = Llama-3.2-3B-Instruct; Llama-3.1-8B = Llama-3.1-8B-Instruct; Qwen2.5-7B = Qwen2.5-7B-Instruct; Qwen3-4B-2507 = Qwen3-4B-Instruct-2507.}
\label{tab:k-ablation-acc-token}
\resizebox{0.85\linewidth}{!}{
\begin{tabular}{lcccccccc}
\toprule
\multirow{2}{*}{\textbf{Model}}
& \multicolumn{2}{c}{\textbf{$k{=}0.10$}}
& \multicolumn{2}{c}{\textbf{$k{=}0.07$}}
& \multicolumn{2}{c}{\textbf{$k{=}0.05$}}
& \multicolumn{2}{c}{\textbf{$k{=}0.03$}} \\
\cmidrule(lr){2-3} \cmidrule(lr){4-5} \cmidrule(lr){6-7} \cmidrule(lr){8-9}
& Acc & Token & Acc & Token & Acc & Token & Acc & Token \\
\midrule
Llama-3.2-3B   & 59.99 & 110.1 & 59.79 & \underline{107.7} & 59.79 & \underline{107.7} & \textbf{60.08} & 114.6 \\
Llama-3.1-8B   & 66.58 & \underline{123.9} & \textbf{66.98} & 137.7 & \textbf{66.98} & 164.3 & 66.93 & 178.6 \\
Qwen2.5-7B     & 71.55 & \underline{183.8} & \textbf{71.75} & 194.0 & 71.63 & 196.6 & 71.63 & 196.6 \\
Qwen3-4B-2507     & 76.02 & \underline{351.4} & \textbf{76.06} & 352.4 & \textbf{76.06} & 353.0 & 76.04 & 355.9 \\
\midrule
\textbf{Avg.}  & 68.54 & \underline{192.3} & 68.65 & 197.9 & 68.62 & 205.4 & \textbf{68.67} & 211.4 \\
\bottomrule
\end{tabular}
}
\end{table}

Table~\ref{tab:k-ablation-acc-token} shows the trade-off between routing accuracy and token efficiency across four $k$ thresholds.
Three patterns emerge:

\textbf{(1) Accuracy is robust across $k$.}
Across all models, routing accuracy varies by at most 0.4\% across $k \in \{0.03, 0.05, 0.07, 0.10\}$ (avg.\ 68.54--68.67), demonstrating that EDRM's routing performance is not sensitive to this hyperparameter.
At $k{=}0.07$, average accuracy is 68.65\%, close to the highest reported average of 68.67\% at $k{=}0.03$.

\textbf{(2) Average token cost decreases with $k$ in this comparison.}
Increasing $k$ widens the trend-based Standard region of the routing rule. In this comparison, average token use falls from 211.4 at $k{=}0.03$ to 192.3 at $k{=}0.10$, although individual model costs need not change monotonically.
The $k{=}0.07$ setting improves mean accuracy by 0.11 percentage points over $k{=}0.10$, with a modest increase in token cost, motivating its use as a balanced default.

\textbf{(3) Comparison with the sample-based deployment protocol.}
The population-level protocol above removes sampling variation. We also compare its default $k{=}0.07$ setting with the 50-sample protocol used in the main table.
The main-table Global-E entries (Table~\ref{tab:overall_comparison})---computed with 50-sample calibration subsets and averaged over 8 seeds---differ from the corresponding cells here by at most 0.29\% in accuracy across the four models (e.g., Llama-3.2-3B-Instruct: 59.79 vs.\ 59.87; Qwen2.5-7B-Instruct: 71.75 vs.\ 71.66), showing similar accuracy at the default $k$ under the two protocols.
Token counts shift more visibly (e.g., Llama-3.2-3B-Instruct: 107.7 vs.\ 113.2) because different routing decisions on borderline benchmarks redistribute work between Direct and Standard/CoT, while accuracy remains close at the default $k$. The full $k$ sweep establishes sensitivity under the population-level protocol; this comparison does not separately evaluate all $k$ values under sampling.

\subsection{Change-Point Detection Results}
\label{app:changepoint}

This section provides the complete change-point detection results across all 4 models and 3 routing groups (12 model--group pairs), as referenced in \S\ref{sec:systematic_analysis}.

\paragraph{Method details.}
We apply two change-point detection algorithms to each sample's first-64-token entropy sequence:
(1) \textbf{CUSUM} (cumulative sum): a sequential analysis technique that detects shifts in the mean of a sequence by monitoring cumulative deviations from the overall mean. We use a two-sided CUSUM with a detection threshold set at the 95th percentile of the permutation null distribution.
(2) \textbf{BinSeg} (binary segmentation): a recursive algorithm that exhaustively searches for the single change point maximizing the between-segment variance reduction, then reports the optimal split location.
For each model, we partition all evaluation samples into three groups based on their Global-E routing decisions (Direct, Standard, CoT) and analyze change-point distributions within each group.
To assess statistical significance, we perform permutation tests ($B{=}1000$) comparing the between-group variance explained by detected change points against that of randomly assigned split points.

\paragraph{Evaluation protocol: population-level routing, label-free.}
The routing groups used here are assigned by the same \emph{population-level} (no-sampling) protocol as the $k$-sensitivity analysis (Appendix~\ref{app:k_sensitivity}): each sample is routed once, deterministically, from its full-set feature statistics, rather than from a 50-sample calibration subset averaged over 8 seeds as in the main-table Global-E. This choice is deliberate and has two consequences. First, the analysis is \emph{entirely label-free}: routing decisions depend only on the entropy-derived features $(S_H, V_{\text{sp}}, a_{\text{vnr}})$, and change-point detection reads only the raw 64-token entropy sequences---correctness labels are never consulted, so the result is independent of any benchmark label correction. Second, group memberships on borderline benchmarks may differ slightly from the 8-seed Global-E routing (cf.\ the token-count discussion in Appendix~\ref{app:k_sensitivity}); we use the deterministic partition here because a fixed, reproducible group assignment is a controlled-attribution device for testing \emph{where} in the entropy sequence the regime shift occurs, not a deployable routing configuration.

\paragraph{Results summary.}
Table~\ref{tab:changepoint_full} presents the complete results. Key findings:
(1) BinSeg-detected change points explain 2.27--3.97$\times$ more between-group variance than random split points (all reported $p<0.001$).
(2) Change-point location correlates strongly with routing semantics: CoT-routed samples exhibit the earliest change points (BinSeg median 9--28 tokens), Direct-routed samples the latest (median 33--38 tokens), with Standard-routed samples lying between the two extremes on three of four models.
(3) Pre/post change-point entropy means differ significantly in 11/12 model--group pairs ($t$-test, $p < 10^{-3}$; exception: Qwen2.5-7B-Instruct under Standard, $p = 0.87$).

\begin{table}[htbp]
\centering
\caption{Complete change-point detection results (12 model--group pairs). Groups are assigned by the deterministic population-level routing protocol (label-free; see protocol note above), not the 8-seed Global-E calibration of the main tables. Permutation ratio $>$ 1 indicates detected change points are more informative than random; all reported $p<0.001$. Model abbreviations: Llama-3.2-3B = Llama-3.2-3B-Instruct; Llama-3.1-8B = Llama-3.1-8B-Instruct; Qwen2.5-7B = Qwen2.5-7B-Instruct; Qwen3-4B-2507 = Qwen3-4B-Instruct-2507.}
\label{tab:changepoint_full}
\begin{adjustbox}{width=0.95\columnwidth, center}
\setlength{\tabcolsep}{2.5pt}
\renewcommand{\arraystretch}{1.2}
\begin{tabular}{l|l|ccc|cc|c}
\toprule
\textbf{Model} & \textbf{Group} & \textbf{$n$} & \textbf{CUSUM det.} & \textbf{BinSeg med.} & \textbf{$\Delta$ Entropy} & \textbf{$t$-test $p$} & \textbf{Perm.\ ratio} \\
\midrule
\multirow{3}{*}{Llama-3.2-3B}
 & Direct & 10925 & 21.8\% & 33 & $-$0.389 & $<10^{-5}$ & 2.78$\times$ \\
 & Standard & 5278 & 9.2\% & 34 & $+$0.141 & $<10^{-5}$ & 3.42$\times$ \\
 & CoT & 6857 & 42.2\% & 24 & $+$0.480 & $<10^{-5}$ & 3.04$\times$ \\
\midrule
\multirow{3}{*}{Llama-3.1-8B}
 & Direct & 4801 & 18.7\% & 35 & $-$0.264 & $<10^{-5}$ & 3.31$\times$ \\
 & Standard & 5160 & 18.5\% & 15 & $+$0.254 & $<10^{-5}$ & 3.18$\times$ \\
 & CoT & 13100 & 29.9\% & 9 & $+$0.594 & $<10^{-5}$ & 3.17$\times$ \\
\midrule
\multirow{3}{*}{Qwen2.5-7B}
 & Direct & 10015 & 23.2\% & 33 & $-$0.482 & $<10^{-5}$ & 2.53$\times$ \\
 & Standard & 4456 & 6.8\% & 36 & $+$0.001 & $0.87$ & 3.97$\times$ \\
 & CoT & 8590 & 36.5\% & 28 & $+$0.468 & $<10^{-5}$ & 2.79$\times$ \\
\midrule
\multirow{3}{*}{Qwen3-4B-2507}
 & Direct & 11869 & 30.3\% & 38 & $-$0.295 & $<10^{-5}$ & 2.27$\times$ \\
 & Standard & 3771 & 8.1\% & 38 & $-$0.012 & $<10^{-3}$ & 3.35$\times$ \\
 & CoT & 7421 & 25.1\% & 28 & $+$0.224 & $<10^{-5}$ & 3.13$\times$ \\
\bottomrule
\end{tabular}
\end{adjustbox}
\end{table}

\subsection{Probe Length Ablation}
\label{app:probe_n}

This section presents the full probe length ablation across $N \in \{16, 32, 48, 64, 80, 96\}$, as referenced in \S\ref{sec:edrm} and~\S\ref{sec:experimental_setup}.

\begin{table}[htbp]
\centering
\caption{Probe length ablation: Instance-level w/o-Fallback / Full accuracy (\%) and average tokens across 4 models. $H_{\mathrm{th}}$ values are adaptively calibrated per $N$ (e.g., Llama-3.2-3B-Instruct: 30.0 at $N{=}64$, vs.\ the bin-searched 32 used in the main-table Inst-E), so the $N{=}64$ row is an independently calibrated run and its values (58.87/63.54/175.5) differ slightly from Table~\ref{tab:ablation} (58.87/63.63/179.4, token includes amortized probe overhead). Full uses oracle answer selection. Model abbreviations: Llama-3.2-3B = Llama-3.2-3B-Instruct; Llama-3.1-8B = Llama-3.1-8B-Instruct; Qwen2.5-7B = Qwen2.5-7B-Instruct; Qwen3-4B-2507 = Qwen3-4B-Instruct-2507.}
\label{tab:probe_n_ablation}
\begin{adjustbox}{width=0.85\columnwidth, center}
\setlength{\tabcolsep}{2pt}
\renewcommand{\arraystretch}{1.2}
\begin{tabular}{c|ccc|ccc|ccc|ccc}
\toprule
\multirow{2}{*}{$N$} & \multicolumn{3}{c|}{\textbf{Llama-3.2-3B}} & \multicolumn{3}{c|}{\textbf{Llama-3.1-8B}} & \multicolumn{3}{c|}{\textbf{Qwen2.5-7B}} & \multicolumn{3}{c}{\textbf{Qwen3-4B-2507}} \\
\cmidrule{2-13}
 & w/oFB & Full & Tok & w/oFB & Full & Tok & w/oFB & Full & Tok & w/oFB & Full & Tok \\
\midrule
16 & 51.97 & 55.33 & 140.8 & 60.59 & 65.85 & 190.9 & 70.21 & 76.01 & 291.4 & 75.66 & 78.66 & 496.0 \\
32 & 56.76 & 61.56 & 173.4 & 63.14 & 68.72 & 170.9 & 69.91 & 74.60 & 240.4 & 76.86 & 79.29 & 458.4 \\
48 & 58.59 & 63.12 & 175.6 & 64.29 & 69.35 & 152.5 & 70.92 & 75.34 & 237.2 & 76.39 & 78.34 & 404.3 \\
64 & 58.87 & 63.54 & 175.5 & 65.11 & 70.01 & 149.5 & 71.60 & 75.99 & 239.6 & 76.91 & 78.65 & 400.6 \\
80 & 58.70 & 63.46 & 172.5 & 64.98 & 69.85 & 144.4 & 71.74 & 76.18 & 241.4 & 76.61 & 78.26 & 388.1 \\
96 & 58.69 & 63.50 & 170.8 & 64.88 & 69.71 & 138.3 & 71.69 & 76.18 & 241.7 & 76.64 & 78.24 & 381.8 \\
\bottomrule
\end{tabular}
\end{adjustbox}
\end{table}

\paragraph{Analysis.}
Accuracy enters a plateau region at $N \geq 48$: w/o-Fallback accuracy varies by at most 1.2\% and Full by at most 0.9 percentage points within $N \in [48, 96]$.
$N{=}64$ sits at the center of this plateau, providing a robust operating point.
Token consumption is non-monotonic in $N$: changes in routing decisions can offset the additional probing cost.
The $H_{\mathrm{th}}$ threshold adapts to $N$ via our calibration procedure (e.g., Llama-3.2-3B-Instruct: $H_{\mathrm{th}}$ = 6.0/30.0/43.0 for $N$ = 16/64/96), confirming that the cumulative entropy threshold correctly scales with probe length.

\subsection{Deployable Selector: Full Results and Empirical Oracle Recovery}
\label{app:selector}

This section provides the selector results referenced in \S\ref{sec:experiments}. The Full configuration generates a routed answer and, when routing selects Standard or CoT, a Direct fallback answer. Full uses oracle answer selection as an accuracy reference. We evaluate six selector schemes that use decoding signals or task priors to choose between disagreeing answers without consulting test labels during deployment. Table~\ref{tab:selector_all_schemes} reports the two probability-based schemes alongside w/o-FB and Full.

\paragraph{Design space.}
We explored three signal families:
(i) \textbf{Posterior signals} (reusing decoding logs): PAE (tail-entropy comparison), PSC (tail-surprisal comparison), DCC (Direct min-probability threshold via leave-one-benchmark-out);
(ii) \textbf{Task priors}: 1A (D-ratio prior, label-free), 1B (global-accuracy prior);
(iii) \textbf{Per-benchmark calibration}: the per-bench $d_{\text{mp}}$ selector (5-fold CV on 50 calibration samples per benchmark, no test labels), which is the best-performing scheme and the one reported in the main-text Table~\ref{tab:ablation} (Selected-FB column).

\begin{table}[htbp]
\centering
\caption{Selector comparison on EDRM-Inst-E (Acc\% / Tok). Token counts include amortized probe overhead. All selector variants share the same token cost as Full, since the Direct fallback branch is determined by routing, not by the answer-selection logic. Gap-recovery percentages cited in the text (e.g., $-$6.3\% on Qwen3-4B-Instruct-2507) are computed from unrounded accuracies and therefore differ slightly from ratios of the rounded values shown here (e.g., $-$5.8\%). Model abbreviations: Llama-3.2-3B = Llama-3.2-3B-Instruct; Llama-3.1-8B = Llama-3.1-8B-Instruct; Qwen2.5-7B = Qwen2.5-7B-Instruct; Qwen3-4B-2507 = Qwen3-4B-Instruct-2507.}
\label{tab:selector_all_schemes}
\begin{adjustbox}{width=0.80\columnwidth, center}
\setlength{\tabcolsep}{2.5pt}
\renewcommand{\arraystretch}{1.15}
\begin{tabular}{l|cccc}
\toprule
\textbf{Scheme} & \textbf{Llama-3.2-3B} & \textbf{Llama-3.1-8B} & \textbf{Qwen2.5-7B} & \textbf{Qwen3-4B-2507} \\
\midrule
w/o-FB (baseline) & 58.87 / 177.1 & 65.11 / 149.8 & 71.63 / 240.7 & 76.81 / 397.9 \\
Global LOO $d_{\text{mp}}$ & 59.17 / 179.4 & 65.31 / 153.9 & 70.93 / 242.7 & 76.81 / 401.1 \\
\textbf{Per-bench $d_{\text{mp}}$} & \textbf{59.59 / 179.4} & \textbf{65.44 / 153.9} & \textbf{71.77 / 242.7} & \textbf{76.71 / 401.1} \\
\midrule
Full (oracle selection) & 63.63 / 179.4 & 70.04 / 153.9 & 76.07 / 242.7 & 78.54 / 401.1 \\
\bottomrule
\end{tabular}
\end{adjustbox}
\end{table}

\paragraph{Exhaustive scheme exploration.}
Beyond the two $d_{\text{mp}}$-based selectors in Table~\ref{tab:selector_all_schemes}, we evaluated four additional signal families: (i) \textbf{posterior signals}---PAE (tail-entropy comparison), PSC (tail-surprisal comparison), both reusing decoding logs; (ii) \textbf{task priors}---1A (D-ratio prior, label-free), 1B (global-accuracy prior). None of these achieved consistent positive gap recovery: PAE/PSC degrade Qwen3-4B-Instruct-2507 by 6--10\% (tail-entropy signals fail under over-reasoning), and 1A/1B generalize poorly across scales. The $d_{\text{mp}}$ signal is the single best discriminator (AUC 0.55--0.72), but per-benchmark calibration is essential---the global LOO variant fails on Qwen2.5-7B-Instruct ($-$15.6\% gap recovery) due to cross-benchmark distribution shift.

\paragraph{Why most selectors fail.}
On the disagreement subset, cases uniquely solved by Direct are less frequent than those uniquely solved by the routed mode ($D_d{:}D_m\approx1{:}3$ to $1{:}8.6$). Switching all disagreements to Direct would therefore reduce accuracy. The evaluated signals have AUC 0.55--0.72; useful selection under this imbalance depends on performance at the chosen switching threshold, rather than on AUC alone.

\paragraph{Empirical Oracle Recovery under the Evaluated Selectors.}
We also evaluate the recovery achievable by the tested selectors when their thresholds are chosen with access to test labels. This retrospective analysis gives an empirical oracle reference for these selector families. In Table~\ref{tab:selector_infobound}, the best single signal ($d_{\text{mp}}$) recovers 0.07--0.89 percentage points of the 1.73--4.93-point Full--w/o-FB gap. Combining five signals through logistic regression raises the observed recovery to at most 1.18 points, still no more than 25\% of the gap. These results characterize the limited recovery obtained by the evaluated signals and selector forms; they do not establish an information-theoretic limit for other selectors.

\begin{table}[htbp]
\centering
\caption{Empirical Oracle Recovery under the Evaluated Selectors. Entries are recovered accuracy in percentage points under per-benchmark thresholds selected with access to test labels, recomputed on the label-corrected evaluation data. ``5-signal LR'' combines $d_{\text{mp}}$, tail entropy, mean probability, answer agreement, and token length via per-benchmark logistic regression with an oracle probability threshold. The final row gives the Full--w/o-FB accuracy gap. All values are computed from unrounded per-sample accuracies. Model abbreviations: Llama-3.2-3B = Llama-3.2-3B-Instruct; Llama-3.1-8B = Llama-3.1-8B-Instruct; Qwen2.5-7B = Qwen2.5-7B-Instruct; Qwen3-4B-2507 = Qwen3-4B-Instruct-2507.}
\label{tab:selector_infobound}
\begin{adjustbox}{width=0.80\columnwidth, center}
\setlength{\tabcolsep}{4pt}
\renewcommand{\arraystretch}{1.2}
\begin{tabular}{l|cccc}
\toprule
\textbf{Signal} & \textbf{Llama-3.2-3B} & \textbf{Llama-3.1-8B} & \textbf{Qwen2.5-7B} & \textbf{Qwen3-4B-2507} \\
\midrule
$d_{\text{mp}}$ (best single) & +0.89 & +0.50 & +0.58 & +0.07 \\
5-signal LR & +1.18 & +0.69 & +0.70 & +0.20 \\
Full--w/o-FB gap & 4.76 & 4.93 & 4.44 & 1.73 \\
\bottomrule
\end{tabular}
\end{adjustbox}
\end{table}

\clearpage
\subsection{Complete Bootstrap and Permutation Test Results}
\label{app:bootstrap}

This section reports the full per-model confidence intervals and $p$-values underlying the statistical analysis in \S\ref{sec:stat_analysis}.

\paragraph{Two complementary aggregation levels.}
Let $J$ be the number of benchmarks, $n_j$ their sample counts, and $\mathrm{Acc}_j(m)$ the accuracy of method $m$ on benchmark $j$. We distinguish the pooled accuracy difference from the unweighted benchmark-mean difference:
\begin{align}
\mathrm{Acc}_{\mathrm{pool}}(m) &= \frac{\sum_{j=1}^{J} n_j\,\mathrm{Acc}_j(m)}{\sum_{j=1}^{J} n_j}, \nonumber\\
\Delta_{\mathrm{inst}} &= \mathrm{Acc}_{\mathrm{pool}}(m_1)-\mathrm{Acc}_{\mathrm{pool}}(m_2), \nonumber\\
\Delta_{\mathrm{bench}} &= \frac{1}{J}\sum_{j=1}^{J}\bigl[\mathrm{Acc}_j(m_1)-\mathrm{Acc}_j(m_2)\bigr].
\label{eq:accuracy_aggregation}
\end{align}
The pooled difference weights benchmarks by sample count and matches the main-table Overall rows and Table~\ref{tab:bootstrap_instance_full}. The benchmark-mean difference gives each benchmark equal weight and is used in Table~\ref{tab:bootstrap_global_full} and the top block of Table~\ref{tab:ablation}. Unequal benchmark sizes therefore produce different numerical differences. For Llama-3.2-3B-Instruct w/o-FB vs.\ CoT, the unrounded values are $\Delta_{\mathrm{inst}}=-1.32$ and $\Delta_{\mathrm{bench}}=-0.88$ percentage points. Subtracting the displayed pooled accuracies, $58.87-60.20$, gives $-1.33$ because those accuracies are rounded separately.

The two analyses address different units of variation: sample-level performance on the pooled evaluation suite and benchmark-level variation across tasks. They also yield different significance assessments in some comparisons. For example, Llama-3.2-3B-Instruct w/o-FB vs.\ CoT has bootstrap $p=0.001$ but benchmark-level permutation $p=0.13$. We report both rather than treating their significance conclusions as interchangeable.

\paragraph{Resampling protocols.}
The instance-level bootstrap draws $23{,}061$ sample indices with replacement for each of $1{,}000$ resamples, applying each draw jointly to all methods to preserve pairing. Table~\ref{tab:bootstrap_instance_full} reports the resulting 95\% confidence intervals and paired two-sided $p$-values. The benchmark-level paired permutation test randomly flips the signs of the 15 per-benchmark differences over $10{,}000$ permutations. Token-saving percentages in Table~\ref{tab:bootstrap_global_full} use pooled token counts without global-probe amortization; Appendix~\ref{app:amortization} separately reports this small overhead, and the main-table token counts follow Appendix~\ref{app:eval_metrics}.

\begin{table}[htbp]
\centering
\caption{Instance-level per-sample bootstrap: full 95\% CIs and paired $p$-values ($n{=}23061$/model, 1000 resamples). $\Delta$Acc is the pooled instance-level difference $\Delta_{\mathrm{inst}}$, relative to CoT unless noted, consistent with the main-table Overall rows. Full rows use oracle answer selection. The benchmark-mean convention in Table~\ref{tab:bootstrap_global_full} is defined above. Model abbreviations: Llama-3.2-3B = Llama-3.2-3B-Instruct; Llama-3.1-8B = Llama-3.1-8B-Instruct; Qwen2.5-7B = Qwen2.5-7B-Instruct; Qwen3-4B-2507 = Qwen3-4B-Instruct-2507.}
\label{tab:bootstrap_instance_full}
\begin{adjustbox}{width=0.75\columnwidth, center}
\setlength{\tabcolsep}{2.5pt}
\renewcommand{\arraystretch}{1.15}
\begin{tabular}{l|l|ccc|cc}
\toprule
\textbf{Model} & \textbf{Method} & \textbf{Acc(\%)} & \textbf{CI lo} & \textbf{CI hi} & \textbf{$\Delta$ vs CoT} & \textbf{$p$} \\
\midrule
\multirow{5}{*}{Llama-3.2-3B}
 & CoT & 60.20 & 59.56 & 60.84 & -- & -- \\
 & Standard & 59.89 & 59.25 & 60.51 & $-$0.31 & -- \\
 & w/o-FB & 58.87 & 58.22 & 59.54 & $-$1.32 & 0.001 ($^{**}$) \\
 & Full & 63.63 & 63.00 & 64.25 & +3.43 & 0.001 ($^{**}$) \\
 & \textit{Full vs w/o-FB} & -- & -- & -- & \textit{+4.76} & \textit{0.001 ($^{**}$)} \\
\midrule
\multirow{5}{*}{Llama-3.1-8B}
 & CoT & 68.11 & 67.47 & 68.70 & -- & -- \\
 & Standard & 66.08 & 65.45 & 66.70 & $-$2.03 & -- \\
 & w/o-FB & 65.11 & 64.54 & 65.74 & $-$3.00 & 0.001 ($^{**}$) \\
 & Full & 70.04 & 69.48 & 70.61 & +1.93 & 0.001 ($^{**}$) \\
 & \textit{Full vs w/o-FB} & -- & -- & -- & \textit{+4.93} & \textit{0.001 ($^{**}$)} \\
\midrule
\multirow{5}{*}{Qwen2.5-7B}
 & CoT & 72.44 & 71.89 & 73.01 & -- & -- \\
 & Standard & 71.43 & 70.89 & 71.99 & $-$1.01 & -- \\
 & w/o-FB & 71.63 & 71.06 & 72.23 & $-$0.81 & 0.001 ($^{**}$) \\
 & Full & 76.07 & 75.55 & 76.63 & +3.63 & 0.001 ($^{**}$) \\
 & \textit{Full vs w/o-FB} & -- & -- & -- & \textit{+4.44} & \textit{0.001 ($^{**}$)} \\
\midrule
\multirow{5}{*}{Qwen3-4B-2507}
 & CoT & 80.75 & 80.26 & 81.26 & -- & -- \\
 & Standard & 79.90 & 79.39 & 80.44 & $-$0.85 & -- \\
 & w/o-FB & 76.81 & 76.27 & 77.37 & $-$3.94 & 0.001 ($^{**}$) \\
 & Full & 78.54 & 78.01 & 79.07 & $-$2.21 & 0.001 ($^{**}$) \\
 & \textit{Full vs w/o-FB} & -- & -- & -- & \textit{+1.73} & \textit{0.001 ($^{**}$)} \\
\bottomrule
\end{tabular}
\end{adjustbox}
\end{table}

\begin{table}[htbp]
\centering
\caption{Benchmark-level paired permutation tests ($n{=}15$, $B{=}10000$). $\Delta$Acc is the unweighted benchmark-mean difference $\Delta_{\mathrm{bench}}$, distinct from $\Delta_{\mathrm{inst}}$ in Table~\ref{tab:bootstrap_instance_full}. Token savings are significant on all four evaluated models ($p<0.001^{***}$). Model abbreviations: Llama-3.2-3B = Llama-3.2-3B-Instruct; Llama-3.1-8B = Llama-3.1-8B-Instruct; Qwen2.5-7B = Qwen2.5-7B-Instruct; Qwen3-4B-2507 = Qwen3-4B-Instruct-2507.}
\label{tab:bootstrap_global_full}
\begin{adjustbox}{width=0.80\columnwidth, center}
\setlength{\tabcolsep}{2.5pt}
\renewcommand{\arraystretch}{1.2}
\begin{tabular}{l|l|cc|cc}
\toprule
\textbf{Model} & \textbf{Comparison} & \textbf{$\Delta$Acc (\%)} & \textbf{$p$} & \textbf{Tok saving} & \textbf{$p$} \\
\midrule
\multirow{3}{*}{Llama-3.2-3B} & w/o-FB vs CoT & $-$0.88 & 0.13 (ns) & \multirow{3}{*}{29.7\%} & \multirow{3}{*}{$<0.001^{***}$} \\
 & w/o-FB vs Std & $-$0.73 & 0.27 (ns) & & \\
 & Full vs CoT & +3.86 & 0.005 ($^{**}$) & & \\
\midrule
\multirow{3}{*}{Llama-3.1-8B} & w/o-FB vs CoT & $-$2.42 & 0.014 ($^{*}$) & \multirow{3}{*}{46.1\%} & \multirow{3}{*}{$<0.001^{***}$} \\
 & w/o-FB vs Std & $-$1.06 & 0.15 (ns) & & \\
 & Full vs CoT & +2.63 & 0.11 (ns) & & \\
\midrule
\multirow{3}{*}{Qwen2.5-7B} & w/o-FB vs CoT & $-$0.38 & 0.54 (ns) & \multirow{3}{*}{27.1\%} & \multirow{3}{*}{$<0.001^{***}$} \\
 & w/o-FB vs Std & +0.13 & 0.77 (ns) & & \\
 & Full vs CoT & +4.04 & 0.014 ($^{*}$) & & \\
\midrule
\multirow{3}{*}{Qwen3-4B-2507} & w/o-FB vs CoT & $-$4.01 & $<0.001$ ($^{***}$) & \multirow{3}{*}{38.1\%} & \multirow{3}{*}{$<0.001^{***}$} \\
 & w/o-FB vs Std & $-$3.15 & $<0.001$ ($^{***}$) & & \\
 & Full vs CoT & $-$2.12 & 0.025 ($^{*}$) & & \\
\bottomrule
\end{tabular}
\end{adjustbox}
\end{table}

\FloatBarrier
\subsection{Token Amortization of the Global Probe}
\label{app:amortization}

The global-level routing probe requires decoding 50 short probe sequences per benchmark. We amortize this overhead across all test samples of the benchmark:
\begin{equation}
\label{eq:amortization}
\mathrm{amortized}_j = \frac{N_{\text{probe}} \times \overline{\mathrm{probe\_tok}}_j}{n_j},
\end{equation}
where $N_{\text{probe}}{=}50$, $\overline{\mathrm{probe\_tok}}_j$ is the mean decode length of the 50 probe samples on benchmark $j$ (capped at $N{=}64$), and $n_j$ is the number of test samples on benchmark $j$.

\begin{table}[htbp]
\centering
\caption{Amortized probe overhead per sample. The overhead is 1.46--2.08 tokens/sample ($<1.8\%$ of total token cost), negligible in all settings. Instance-level routing embeds the probe in the predecode ($\min(64, T_{\text{std}})$), so no additional amortization is needed. Model abbreviations: Llama-3.2-3B = Llama-3.2-3B-Instruct; Llama-3.1-8B = Llama-3.1-8B-Instruct; Qwen2.5-7B = Qwen2.5-7B-Instruct; Qwen3-4B-2507 = Qwen3-4B-Instruct-2507.}
\label{tab:amortization}
\begin{adjustbox}{width=0.50\columnwidth, center}
\setlength{\tabcolsep}{5pt}
\renewcommand{\arraystretch}{1.2}
\begin{tabular}{l|cccc}
\toprule
\textbf{Model} & \textbf{Amortized} & \textbf{Min} & \textbf{Max} & \textbf{$n$ total} \\
\midrule
Llama-3.2-3B & 1.97 & 1.25 & 6.46 & 23061 \\
Llama-3.1-8B & 1.46 & 0.31 & 5.29 & 23061 \\
Qwen2.5-7B & 2.07 & 1.33 & 7.14 & 23061 \\
Qwen3-4B-2507 & 2.08 & 1.33 & 7.14 & 23061 \\
\bottomrule
\end{tabular}
\end{adjustbox}
\end{table}

The per-benchmark amortized cost ranges from 0.31 (large benchmarks such as ARC-Easy) to 7.14 tokens (small benchmarks such as GPQA with $n{=}448$). Even the worst case ($7.14$ tokens on GPQA) is under 3\% of that benchmark's average CoT length. Instance-level routing does not incur this overhead, as its probe is the $\min(64, T_{\text{std}})$ predecode already counted in the Standard token cost.

\FloatBarrier
\subsection{Token-Signature Baseline Specification}
\label{app:ts_baseline}

We use the original Dynamic CoT routing mechanism of Token-Signature (TS)~\cite{Liu2025TokenSP}, with a 64-step probe in the main comparison instead of the original 50-step setting. Both configurations use the same routing procedure; Table~\ref{tab:ts_probe_length} reports their results.

\paragraph{Signal.}
TS and EDRM use the same Standard greedy-decoding logs. TS reads the probability assigned to each greedily generated token (the maximum softmax probability), whereas EDRM reads entropy over the full next-token distribution. TS computes its instance-level statistic as the Spearman correlation between the first $N$ observed token probabilities and their positions. A positive correlation indicates increasing token probability, which the original paper associates with CoT benefit. Correlations that are undefined, including sequences shorter than two steps, are set to zero.

\paragraph{Supervised fitting and binary routing.}
For each benchmark, training candidates are instances on which Direct and CoT differ in correctness; cases where both are correct or both are wrong are excluded from fitting. We sample up to 50 candidates with random seed 42 and fit a single-feature logistic regression on their probability-trend statistics, using CoT-correct as class 1 and Direct-correct as class 0. The implementation uses the default \texttt{LogisticRegression} settings with \texttt{random\_state=42}. It routes to CoT when $P(\mathrm{CoT})>0.5$ and to Direct otherwise. If no training candidates exist, it routes all instances to CoT; if the fitting set contains only one class, it routes all instances to that class. The router has no Standard output branch. This fitting procedure uses correctness labels and is distinct from EDRM's label-free entropy-threshold calibration.

\paragraph{Token accounting and evaluation sets.}
Each instance is charged the selected Direct or CoT generation plus the actual probe cost $\min(N,T_{\mathrm{std}})$. Table~\ref{tab:overall_comparison} uses the 64-step, whole-set configuration, including the router-fitting samples, to match the full-set basis of the other methods. The held-out (clean) evaluation uses the same fitted classifier and routing decisions but excludes its fitting samples from the reported means; no second classifier is trained. For $N=64$, whole-set and held-out accuracies differ by at most 0.16 percentage points across the seven models. This comparison describes sensitivity to the evaluation set; the fitting samples remain included in the main-table results.

\paragraph{Probe-length comparison.}
Accuracy differs by at most 0.22 percentage points between the 50- and 64-step configurations in Table~\ref{tab:ts_probe_length}. The original 50-step setting uses fewer tokens on every model. We report it alongside the 64-step configuration so that the effect of matching EDRM's probe length is explicit.

\begin{table}[htbp]
\centering
\caption{Token-Signature under the original $N{=}50$ probe setting vs.\ the $N{=}64$ setting used in Table~\ref{tab:overall_comparison} (whole-set evaluation, Acc\% / Tok). Accuracy differences are $\leq 0.22$ percentage points, and $N{=}50$ is cheaper on all models. Model abbreviations: Llama-3.2-3B = Llama-3.2-3B-Instruct; Llama-3.1-8B = Llama-3.1-8B-Instruct; Qwen2.5-7B = Qwen2.5-7B-Instruct; Qwen2.5-32B = Qwen2.5-32B-Instruct; Qwen3-4B-2507 = Qwen3-4B-Instruct-2507; DR1-Qwen-7B = DeepSeek-R1-Distill-Qwen-7B; Qwen3.5-35B = Qwen3.5-35B-A3B.}
\label{tab:ts_probe_length}
\small
\setlength{\tabcolsep}{4pt}
\renewcommand{\arraystretch}{1.15}
\begin{tabular}{l|cc}
\toprule
\textbf{Model} & \textbf{TS ($N{=}50$)} & \textbf{TS ($N{=}64$)} \\
\midrule
Llama-3.2-3B & 60.83 / 233.1 & 60.78 / 251.0 \\
Llama-3.1-8B & 68.18 / 255.3 & 68.26 / 264.3 \\
Qwen2.5-7B & 72.91 / 331.1 & 72.78 / 356.7 \\
Qwen2.5-32B & 79.34 / 249.6 & 79.54 / 266.9 \\
Qwen3-4B-2507 & 80.70 / 676.5 & 80.74 / 682.3 \\
DR1-Qwen-7B & 70.76 / 1284.4 & 70.76 / 1298.4 \\
Qwen3.5-35B & 84.50 / 1498.2 & 84.50 / 1512.2 \\
\bottomrule
\end{tabular}
\end{table}

\paragraph{Relation to EDRM's feature ablations.}
The TS comparison evaluates complete routing methods. EDRM's w/o-$a_{\mathrm{vnr}}$ ablation uses entropy and three-way thresholds; TS uses token probabilities and a learned binary selector. Table~\ref{tab:ablation} separately assesses EDRM's feature contributions.

\clearpage
\subsection{Calibration Protocol and Data Partitioning}
\label{app:calibration_protocol}

We clarify the roles of training, calibration, and evaluation data for EDRM and TS. Table~\ref{tab:data_partition} summarizes these roles; Appendix~\ref{app:ts_baseline} details the TS fitting and evaluation sets.

\begin{table}[htbp]
\centering
\caption{Data roles for training, calibration, and evaluation. MLP and TS fitting use correctness labels. Main results use full-set evaluation, including their fitting samples; held-out evaluations are reported separately. Selector calibration is distinct from the retrospective oracle-threshold analysis.}
\label{tab:data_partition}
\begin{adjustbox}{width=0.95\columnwidth, center}
\setlength{\tabcolsep}{4pt}
\renewcommand{\arraystretch}{1.2}
\begin{tabular}{p{0.18\linewidth}|p{0.22\linewidth}p{0.34\linewidth}p{0.17\linewidth}}
\toprule
\textbf{Role} & \textbf{Used for} & \textbf{Source} & \textbf{Labels?} \\
\midrule
Tuning (MLP) & Train MLP router & 10\% stratified per benchmark & Yes \\
Fitting (TS) & Train logistic router & Up to 50 disagreement samples per benchmark & Yes \\
Calibration & Calibrate $H_{\text{th}}$, $\theta_{\text{bench}}$ & 50 samples per benchmark & No test labels \\
Evaluation & Report main / held-out results & Full sets / held-out subsets & Eval; Full selection \\
\bottomrule
\end{tabular}
\end{adjustbox}
\end{table}

\paragraph{Routing calibration and answer-selection calibration.}
The heuristic entropy threshold uses the entropy-derived features of the probe samples without reading answer correctness. The per-benchmark selector threshold $\theta_{\text{bench}}$ is a separate calibration step: it uses the Direct answer's minimum token probability $d_{\text{mp}}$ and the 5-fold calibration procedure on 50 samples described in Appendix~\ref{app:selector}. Test labels are not used to choose the deployed selector threshold. The retrospective oracle-threshold analysis in that appendix is reported separately and explicitly permits test-label access.

\paragraph{MLP evaluation protocol.}
EDRM-MLP uses a 10\% stratified training split per benchmark. Main results cover the full sets, including these fitting samples. Tables~\ref{tab:edrm-mlp-boost-ablation} and~\ref{tab:mlp-remaining-ml64d} provide held-out 90\% results with and without fallback, documenting sensitivity to the evaluation split.

\end{document}